\documentclass[fleqn,10pt]{wlscirep}
\usepackage[utf8]{inputenc}
\usepackage[T1]{fontenc}

\usepackage{multirow}
\title{Predicting drug-gene relations via analogy tasks with word embeddings}
\usepackage{caption}
\usepackage{subcaption}

\usepackage{amsmath}
\DeclareMathOperator{\argmax}{argmax}

\usepackage{comment}

\author[1,$\alpha$,$\ast$]{Hiroaki Yamagiwa}
\author[2,$\beta$]{Ryoma Hashimoto}
\author[3,$\gamma$]{Kiwamu Arakane}
\author[4,$\delta$]{Ken Murakami}
\author[3,$\varepsilon$]{Shou Soeda}
\author[1,5,$\zeta$]{Momose Oyama}
\author[1,$\eta$]{Yihua Zhu}
\author[3,$\theta$]{Mariko Okada}
\author[1,5,$\iota$]{Hidetoshi Shimodaira}

\affil[1]{Kyoto University, Kyoto, Japan}
\affil[2]{Recruit Co., Ltd., Tokyo, Japan}
\affil[3]{Institute for Protein Research, Osaka University, Osaka, Japan}
\affil[4]{Present address: Research Institute of Molecular Pathology, Vienna BioCenter, Vienna, Austria}
\affil[5]{RIKEN, Tokyo, Japan}

\affil[$\alpha$]{h.yamagiwa@i.kyoto-u.ac.jp}
\affil[$\beta$]{ryoma\_hashimoto@r.recruit.co.jp}
\affil[$\gamma$]{k.arakane@protein.osaka-u.ac.jp}
\affil[$\delta$]{ken.murakami@imp.ac.at}
\affil[$\varepsilon$]{shousoeda@protein.osaka-u.ac.jp}
\affil[$\zeta$]{oyama.momose@sys.i.kyoto-u.ac.jp}
\affil[$\eta$]{zhu.yihua.22h@st.kyoto-u.ac.jp}
\affil[$\theta$]{mokada@protein.osaka-u.ac.jp}
\affil[$\iota$]{shimo@i.kyoto-u.ac.jp}

\affil[$\ast$]{Corresponding author}

\begin{abstract}
Natural language processing (NLP) is utilized in a wide range of fields, where words in text are typically transformed into feature vectors called embeddings. BioConceptVec is a specific example of embeddings tailored for biology, trained on approximately 30 million PubMed abstracts using models such as skip-gram. Generally, word embeddings are known to solve analogy tasks through simple vector arithmetic. For example, subtracting the vector for man from that of king and then adding the vector for woman yields a point that lies closer to queen in the embedding space. In this study, we demonstrate that BioConceptVec embeddings, along with our own embeddings trained on PubMed abstracts, contain information about drug-gene relations and can predict target genes from a given drug through analogy computations. We also show that categorizing drugs and genes using biological pathways improves performance. Furthermore, we illustrate that vectors derived from known relations in the past can predict unknown future relations in datasets divided by year. Despite the simplicity of implementing analogy tasks as vector additions, our approach demonstrated performance comparable to that of large language models such as GPT-4 in predicting drug-gene relations.
\end{abstract}

\begin{document}

\flushbottom
\maketitle

\thispagestyle{empty}

\section*{Introduction}
Natural language processing (NLP) is a computer technology for processing human language. NLP is used in various applications, such as machine translation~\cite{NIPS2017_3f5ee243,DBLP:journals/jmlr/RaffelSRLNMZLL20}, sentiment analysis~\cite{DBLP:conf/emnlp/SocherPWCMNP13,devlin-etal-2019-bert}, and sentence similarity computation~\cite{cer-etal-2017-semeval,DBLP:conf/iclr/MuV18}. Many of these applications use models such as skip-gram~\cite{DBLP:journals/corr/abs-1301-3781,DBLP:conf/nips/MikolovSCCD13} or BERT~\cite{devlin-etal-2019-bert} to convert words in text to embeddings or distributed representations, which are feature vectors with hundreds of dimensions. In particular, skip-gram is a method that learns high-performance embeddings by predicting the surrounding words of a word in a sentence. 
These embeddings form the foundational building blocks of recent advances in large language models, enabling models to understand and generate human language with unprecedented accuracy and coherence.

Mikolov et al.~\cite{DBLP:journals/corr/abs-1301-3781} showed that skip-gram embeddings have properties such as $\mathbf{u}_\mathrm{\emph{king}} - \mathbf{u}_\mathrm{\emph{man}} + \mathbf{u}_\mathrm{\emph{woman}}\approx\mathbf{u}_\mathrm{\emph{queen}}$, where the vector $\mathbf{u}_w$ represents the embedding of word $w$. In the embedding space, the vector differences such as $\mathbf{u}_\mathrm{\emph{king}} - \mathbf{u}_\mathrm{\emph{man}}$ and $\mathbf{u}_\mathrm{\emph{queen}} - \mathbf{u}_\mathrm{\emph{woman}}$ can be seen as vectors representing the \emph{royalty} relation. Since these relations are not explicitly taught during the training of skip-gram, the embeddings acquire these properties spontaneously. Solving questions such as ``If \emph{man} corresponds to \emph{king}, what does \emph{woman} correspond to?'' requires understanding the relations between words. This type of problem is known as an \emph{analogy task} and is used to evaluate a model's language comprehension and reasoning skills. In this example, the model must understand the \emph{royalty} relations from \emph{man} to \emph{king} and apply a similar relation to \emph{woman}. 
Allen and Hospedales~\cite{pmlr-v97-allen19a} explained why analogy computation with such simple vector arithmetic can effectively solve analogy tasks.

Recently, NLP methods have also gained attention in the field of biology~\cite{DBLP:journals/bioinformatics/LeeYKKKSK20,giorgi-etal-2022-sequence,DBLP:journals/bib/LuoSXQZPL22}. However, since traditional skip-gram models are typically trained on web corpora, they do not properly handle the specialized terms that appear in biological texts. In particular, when multiple words represent the same concept, these concepts should be normalized beforehand and represented by the same embedding. With this in mind, Chen et al.~\cite{DBLP:journals/ploscb/ChenLYKWL20} proposed a method to compute word embeddings of biological concepts using approximately 30 million PubMed abstracts in which the mentions of these concepts had been previously normalized using PubTator~\cite{DBLP:journals/nar/WeiKL13}. PubTator is an online tool that supports the automatic annotation of biomedical text and aims at extracting specific information efficiently. Particularly, it can identify the mentions of biological concepts and entities within text and classify them into appropriate categories (e.g., diseases, genes, drugs). They trained four embedding models, including skip-gram, and named their embeddings BioConceptVec, and assessed the usefulness of BioConceptVec in two ways: intrinsic and extrinsic evaluations. For intrinsic evaluations, they identified related genes based on drug-gene and gene-gene interactions using cosine similarity of the embeddings. For extrinsic evaluations, they performed protein-protein interaction prediction and drug-drug interaction extraction using neural network classifiers with the embeddings. However, the embeddings' performance on solving the analogy tasks is yet to be explored.

In this study, we consider analogy tasks in biology using word embeddings.
We trained skip-gram embeddings similarly to BioConceptVec, aiming to compare them with BioConceptVec embeddings and to develop embeddings from datasets divided by year in an extended experiment.
To evaluate the performance of analogy computation, we focus on predicting target genes from a given drug.
These target genes are associated with proteins that the drug interacts with, and these interactions are called \emph{drug-target interactions} (DTIs)~\cite{sachdev2019comprehensive,djeddi2023advancing}. In this paper, we will refer to these connections as \emph{drug-gene relations}. Our research aims to show that embeddings learned from biological text data contain information about drug-gene relations.
In the example of $\mathbf{u}_\mathrm{\emph{king}} - \mathbf{u}_\mathrm{\emph{man}}  + \mathbf{u}_\mathrm{\emph{woman}} \approx \mathbf{u}_\mathrm{\emph{queen}}$, the vector differences $\mathbf{u}_\mathrm{\emph{king}} - \mathbf{u}_\mathrm{\emph{man}}$ and $\mathbf{u}_\mathrm{\emph{queen}} - \mathbf{u}_\mathrm{\emph{woman}}$ represent the \emph{royalty} relation. However, there are multiple drug-gene pairs with drug-gene relations, and a single drug often has multiple target genes. Therefore, we calculate the vector differences between the embeddings of each drug and its target genes and average these vector differences to define a mean vector representing the relation. 

To evaluate the performance in predicting drug-gene relations, we use data derived from KEGG~\cite{10.1093/nar/28.1.27,Kanehisa2019_ProteinSci,Kanehisa2025_NAR} as the ground truth.
We first consider analogy tasks in the global setting where all drugs and genes are included and demonstrate high performance in solving these tasks. Next, to consider more detailed analogy tasks, we use information common to both drugs and genes. Specifically, we focus on biological pathways to categorize drugs and genes. 
To do this, we use a list of human pathways from the KEGG API.
We group all drugs and genes that are associated with the same pathway into a single category and define vectors representing drug-gene relations for each pathway. In this pathway-wise setting, we demonstrate that using these vectors in the analogy computation can improve its performance. Finally, as an application of analogy computation, we investigate the potential of our approach to predict unknown drug-gene relations. For this purpose, we divide the vocabulary by year to distinguish between known and unknown drug-gene relations. We then train skip-gram embeddings using PubMed abstracts published before the specified year and redefine vectors representing the known relations to predict unknown relations. The experimental results show that our approach can predict unknown relations to a certain extent.

The structure of this paper is as follows. First, we explain related work and detail our approach in this study. In the following sections, we perform experiments for each setting and evaluate the performance using metrics such as top-1 accuracy. Finally, we discuss the prediction results and then conclude.

\section*{Related work}
Analogy computation of word embeddings trained from corpora in a particular field of natural science can be used to predict the relations that exist between specialized concepts or terms. Tshitoyan et al.~\cite{DBLP:journals/nature/TshitoyanDWDRKP19} have shown that the relation between specialized concepts in the field of materials science such as ``$\text{ferromagnetism} - \text{NiFe} + \text{IrMn} \approx \text{antiferromagnetism}$'' can be correctly predicted using analogy computation of word embeddings trained with domain-specific literature. This success in the field of materials science demonstrates the potential of analogy computation of word embeddings to predict relationships between specialized concepts in scientific domains.

In the biomedical field, capturing the relations between biomedical concepts, such as drug-drug and protein-protein interactions, is an important issue. Word embeddings trained on biomedical domain corpora have been used to predict drug-drug interactions for new drugs~\cite{shtar2022predicting} and to construct networks of gene-gene interactions~\cite{alachram2021textmining}. 

While word embeddings have shown promise in various biomedical applications, their potential for predicting drug-gene relations remains relatively unexplored.
To the best of our knowledge, there is limited research specifically focused on utilizing analogy computation of word embeddings to predict drug-gene relations in the biological domain.
Our study aims to address this gap by investigating the application of analogy computation of word embeddings to drug-gene relation prediction.

In addition, neural networks have been used in a number of studies to predict these relations~\cite{liu2016drug-drug,sahu2018drug-drug,jiang2016protein,quan2020protein,zhang2018hybrid}.
However, it is important to note that these relation extraction approaches differ from our proposed task.
While they are useful for identifying explicitly stated relationships in text, our focus is on leveraging the latent semantic information captured by word embeddings to predict potential drug-gene relations, even when such relationships are not explicitly mentioned in the literature.

The technologies developed in NLP, not limited to word embeddings, have been applied to a wide variety of problems in the biomedical domain, demonstrating their utility in this field. M\"{u}ller et al.~\cite{muller2004textpresso} have created an online literature search and curation system using an ``ontology dictionary'' obtained by text mining, and Friedman et al.~\cite{friedman2001genies} have used syntactic parsing to extract and structure information about cellular pathways from biological literature. Yeganova et al.~\cite{yeganova2020bettersynonyms} normalize synonyms in the biomedical literature by word embedding similarity. Furthermore, Du et al.~\cite{du2019gene2vec} applied the word embedding algorithm to gene co-expression across the transcriptome to compute vector representations of genes. In recent years, there has also been intensive research on fine-tuning BERT~\cite{devlin-etal-2019-bert} and other pre-trained language models to the biomedical domain~\cite{DBLP:journals/bioinformatics/LeeYKKKSK20,peng-etal-2019-transfer,fang2023bioformer}.

\section*{Methods}
\begin{figure}[!t]
\centering
\begin{minipage}{0.49\linewidth}
\centering
\includegraphics[height=4cm,keepaspectratio]{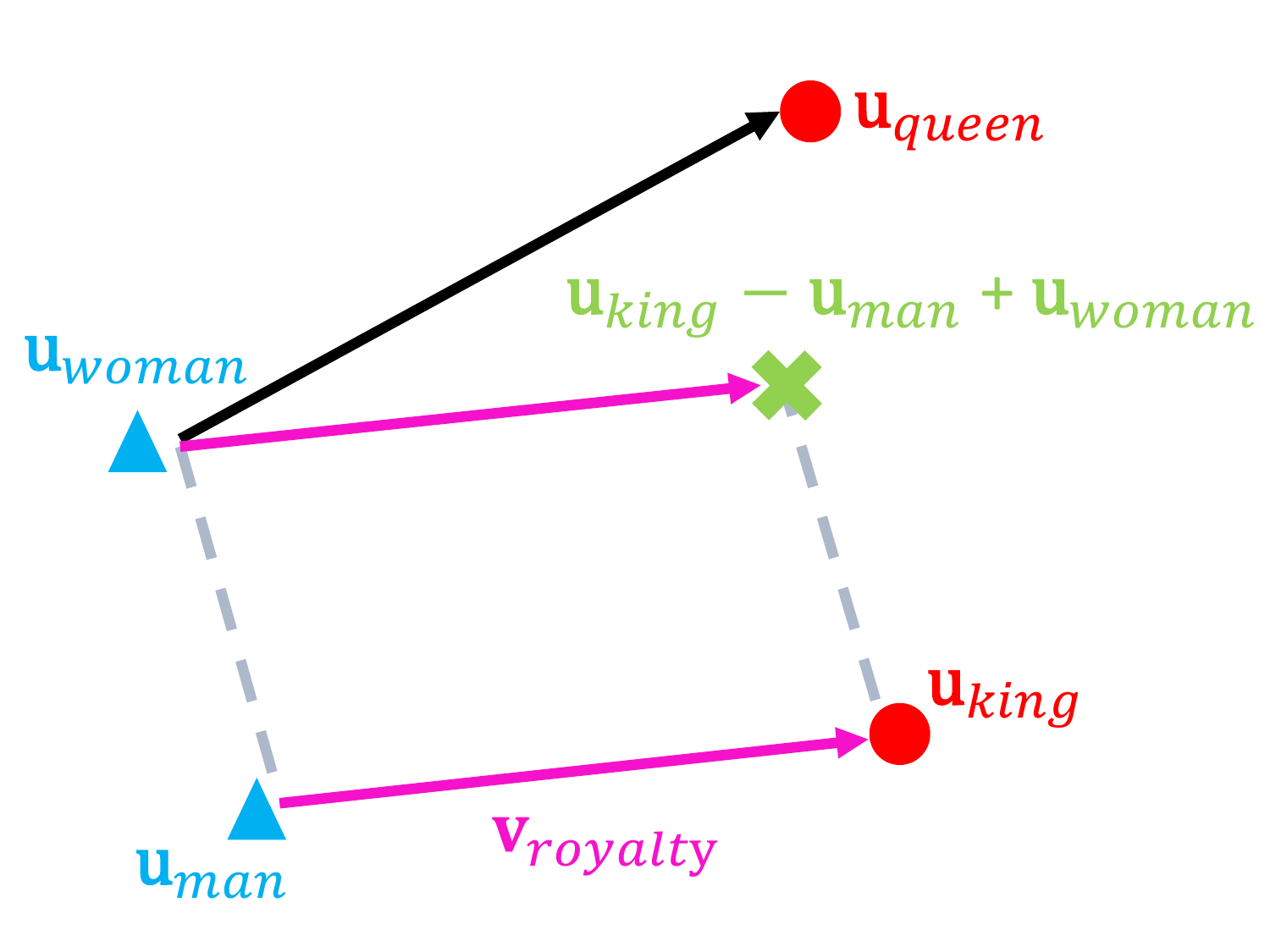}
\subcaption{Predicting \emph{queen} from \emph{woman}}
\label{fig:analogy}
\end{minipage}\hfill
\begin{minipage}{0.49\linewidth}
\centering
\includegraphics[height=4cm,keepaspectratio]{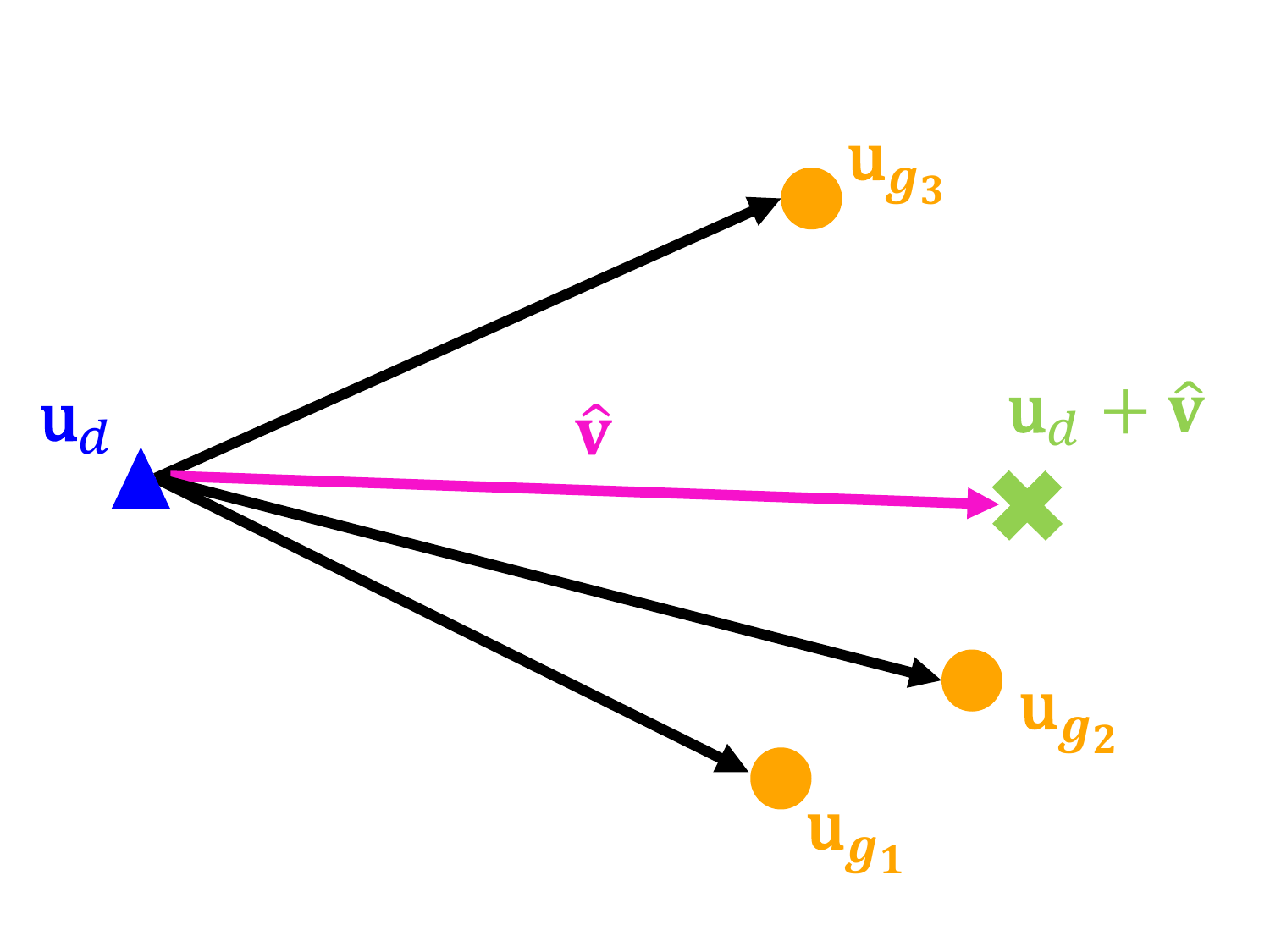}
\subcaption{Predicting genes from a drug}
\label{fig:G}
\end{minipage}
\caption{
Analogy computations by adding the relation vector. (a)~Example of a basic analogy task. The question ``If \emph{man} corresponds to \emph{king}, what does \emph{woman} correspond to?'' is solved by adding the relation vector $\mathbf{v}_\emph{royalty}$ to $\mathbf{u}_\emph{woman}$.
(b)~Example of an analogy task in setting G.
The target genes of the drug $d$ are $g_1$, $g_2$, and $g_3$, predicted by adding the relation estimator $\hat{\mathbf{v}}$ to the drug embedding $\mathbf{u}_d$.
}
\label{fig:analogy_explanation}
\end{figure}

\subsection*{Analogy tasks}
This section illustrates a basic analogy task, a problem setting commonly used in benchmark tests to evaluate the performance of word embedding models.
We define $\mathcal{V}$ as the vocabulary set, and $\mathbf{u}_w \in \mathbb{R}^K$ as the word embedding of word $w \in \mathcal{V}$. In skip-gram models, the ``closeness of word embeddings'' measured by cosine similarity correlates well with the ``closeness of word meanings''~\cite{DBLP:journals/corr/abs-1301-3781}. For example, in word2vec~\cite{DBLP:journals/corr/abs-1301-3781}, $\cos(\mathbf{u}_\mathrm{\emph{media}},\mathbf{u}_\mathrm{\emph{press}})=0.601>\cos(\mathbf{u}_\mathrm{\emph{media}},\mathbf{u}_\mathrm{\emph{car}})=0.020$. This result shows that the embedding of \emph{media} is closer to the embedding of \emph{press} than to the embedding of \emph{car}. This is consistent with the rational human interpretation that \emph{media} is semantically closer to \emph{press} than to \emph{car}. The task for predicting relations between words is known as an analogy task, and word embeddings can effectively solve analogy tasks using vector arithmetic\cite{DBLP:journals/corr/abs-1301-3781}. For example, to solve the question ``If \emph{man} corresponds to \emph{king}, what does \emph{woman} correspond to?'', using pre-trained skip-gram embeddings, we search for $w\in \mathcal{V}$ that maximizes $\cos(\mathbf{u}_\mathrm{\emph{king}} - \mathbf{u}_\mathrm{\emph{man}} + \mathbf{u}_\mathrm{\emph{woman}},\mathbf{u}_w)$, and find $w=\mathrm{\emph{queen}}$, indicating
\begin{align}
    \mathbf{u}_\mathrm{\emph{king}} - \mathbf{u}_\mathrm{\emph{man}} + \mathbf{u}_\mathrm{\emph{woman}} \approx \mathbf{u}_\mathrm{\emph{queen}}.
\end{align}
This analogy computation is illustrated in Fig.~\ref{fig:analogy}.
Here, the vector $\mathbf{v}_\mathrm{\emph{royalty}}$ represents the \emph{royalty} relation between \emph{man} and \emph{king}, and is defined as the vector difference between $\mathbf{u}_\mathrm{\emph{king}}$ and $\mathbf{u}_\mathrm{\emph{man}}$:
\begin{align}
    \mathbf{v}_\mathrm{\emph{royalty}} := \mathbf{u}_\mathrm{\emph{king}} - \mathbf{u}_\mathrm{\emph{man}}.\label{eq:v_royalty}
\end{align}
Adding the \emph{royalty} relation vector $\mathbf{v}_\mathrm{\emph{royalty}}$ to $\mathbf{u}_\mathrm{\emph{woman}}$ then yields $\mathbf{u}_\mathrm{\emph{queen}}$:
\begin{align}
    \mathbf{u}_\mathrm{\emph{woman}} + \mathbf{v}_\mathrm{\emph{royalty}} \approx \mathbf{u}_\mathrm{\emph{queen}}.\label{eq:woman_plus_royalty}
\end{align}

\subsection*{Analogy tasks for drug-gene pairs}
In this section, based on the basic analogy task from the previous section, we explain analogy tasks for predicting target genes from a drug. First, we consider the global setting, where all drugs and genes in the vocabulary set are used. Then, we consider the pathway-wise setting, where drugs and genes are categorized based on biological pathways.

\subsubsection*{Global setting}
\begin{figure}[!t]
\centering
\begin{minipage}{0.49\linewidth}
\centering
\includegraphics[height=6.5cm,keepaspectratio]{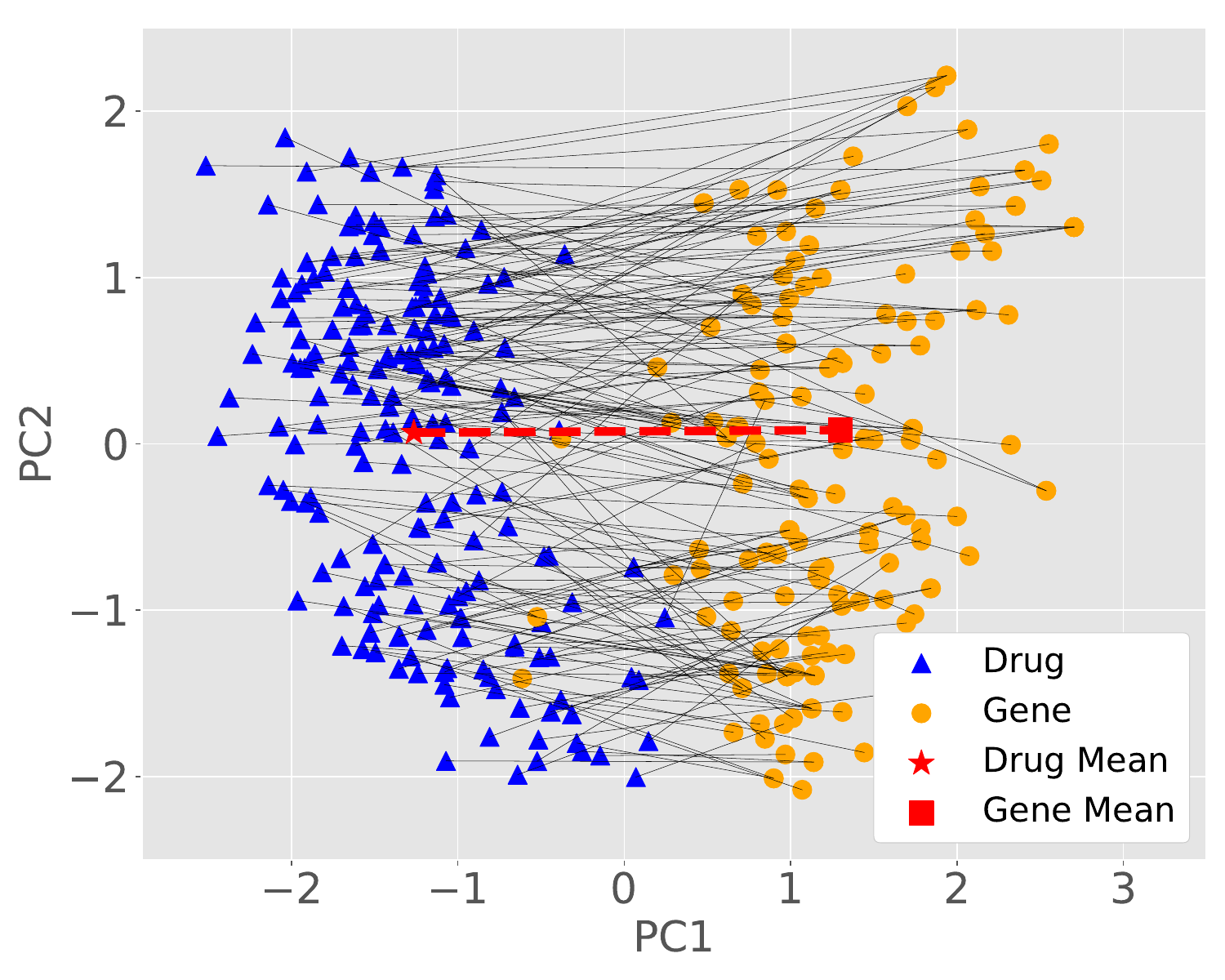}
\subcaption{Randomly sampled drug-gene pairs.}
\label{fig:relation_samples}
\end{minipage}\hfill
\begin{minipage}{0.49\linewidth}
\centering
\includegraphics[height=6.5cm,keepaspectratio]{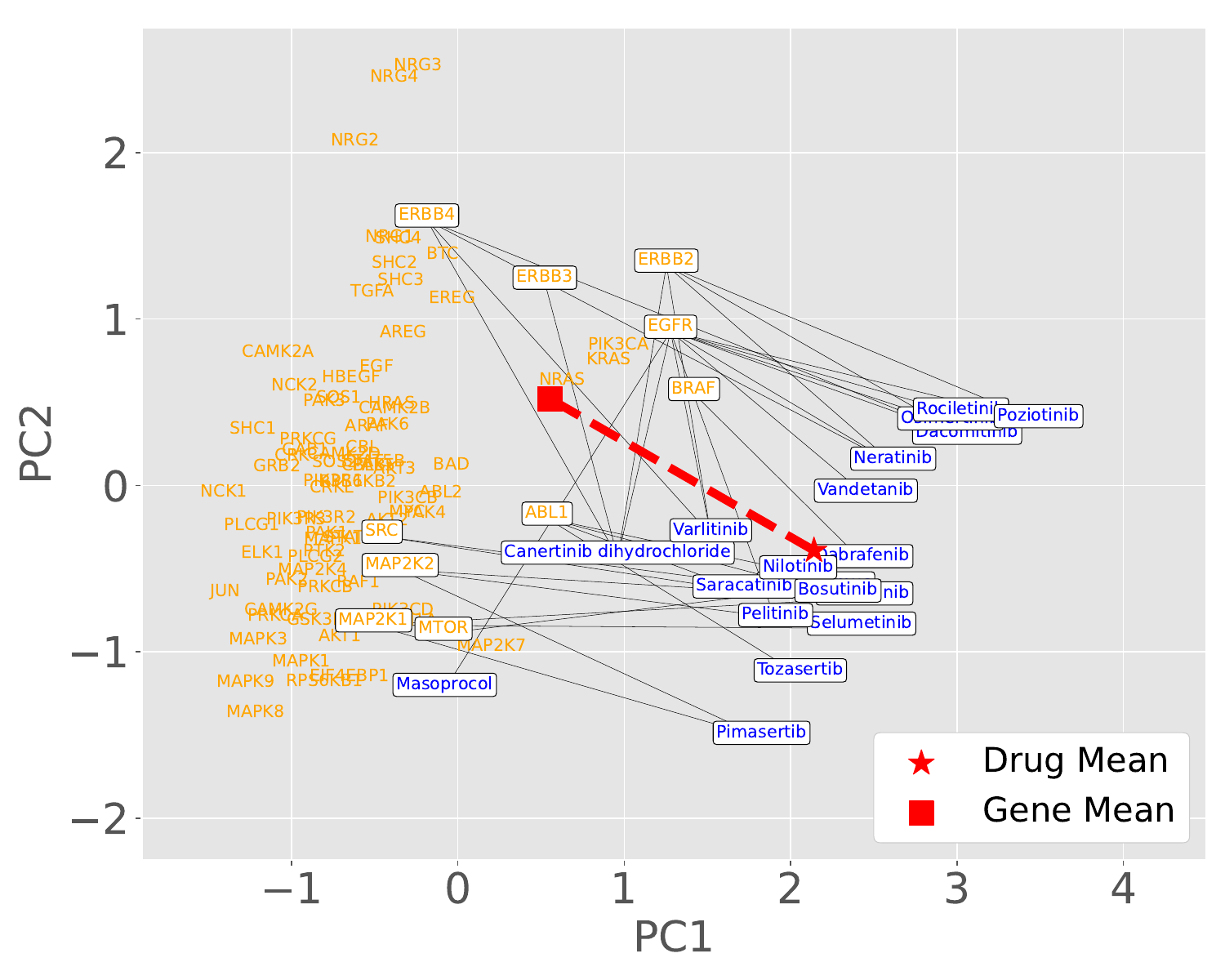}
\subcaption{Drugs and genes categorized in the \emph{ErbB signaling pathway}.}
\label{fig:hsa04012}
\end{minipage}
\caption{
2-d visualization of embeddings using PCA. Drugs and genes are shown in  {\color{blue}blue} and {\color{orange}orange}, respectively.
Solid lines represent the relations.
The symbols {\color{red}$\bigstar$} and {\color{red}$\blacksquare$} represent the mean embeddings of the drugs and genes, respectively.
The direction of the {\color{red}dashed line} connecting these two symbols can be considered as the drug-gene relation vector. 
(a)~Randomly sampled $200$ drug-gene pairs from $\mathcal{R}$.
The mean embeddings are computed by Eq.~(\ref{eq:v_def2_exp}).
(b)~Drugs $d \in \mathcal{D}_p$ and genes $g \in \mathcal{G}_p$ for $p=$\emph{ErbB signaling pathway}. Those that have drug-gene relations are shown in boxes.
The mean embeddings are computed by Eq.~(\ref{eq:vp_def_2_exp}) in Supplementary Information\ref{supp:pathway}.
}
\label{fig:PCA_explanation}
\end{figure}

As a first step, based on the basic analogy task, we consider analogy tasks in the global setting, where all drugs and genes are used.
We define $\mathcal{D}\subset \mathcal{V}$ and $\mathcal{G}\subset \mathcal{V}$ as the sets of all drugs and genes, respectively. We also define $\mathcal{R}\subset \mathcal{D}\times \mathcal{G}$ as the set of drug-gene pairs with drug-gene relations. Thus, if a drug $d\in \mathcal{D}$ and a gene $g\in \mathcal{G}$ have a drug-gene relation, the pair $(d,g)$ is in $\mathcal{R}$. 
For illustrating the positions of drugs and genes in the $100$-dimensional BioConceptVec skip-gram embeddings, we sampled $200$ drug-gene pairs in $\mathcal{R}$ and plotted them in Fig.~\ref{fig:relation_samples} using Principal Component Analysis (PCA). 
The vector difference between the mean embeddings of drugs and genes points in roughly the same direction as the vector differences between the embeddings of each sampled drug and its target gene.

To apply the analogy computation of Eq.~(\ref{eq:woman_plus_royalty}) to drug-gene pairs $(d,g)\in \mathcal{R}$, we consider the analogy tasks for predicting the gene $g$ from the drug $d$. 
Using the vector $\mathbf{v}$ representing the drug-gene relation, we predict $\mathbf{u}_g$ by adding the relation vector $\mathbf{v}$ to $\mathbf{u}_d$ as follows:
\begin{align}
    \mathbf{u}_d+\mathbf{v} \approx \mathbf{u}_g.\label{eq:d_plus_v}
\end{align}
Thus, we need methods for estimating the relation vector $\mathbf{v}$. 
In Eq.~(\ref{eq:v_royalty}), the \emph{royalty} relation vector $\mathbf{v}_\text{\emph{royalty}}$ is calculated from the embeddings of the given pair (\emph{man}, \emph{king}). Therefore, we estimate the relation vector $\mathbf{v}$ from the embeddings of genes and drugs. The following equation gives the vector difference between the mean vectors of $\mathcal{D}$ and $\mathcal{G}$ as a naive estimator of $\mathbf{v}$:
\begin{align}
    &\hat{\mathbf{v}}_\text{naive} :=\mathrm{E}_\mathcal{G}\{\mathbf{u}_g\} - \mathrm{E}_\mathcal{D}\{\mathbf{u}_d\},\label{eq:v_def1}\\
    &\mathrm{E}_\mathcal{D}\{\mathbf{u}_d\} = \frac{1}{|\mathcal{D}|}\sum_{d\in \mathcal{D}}\mathbf{u}_d,\quad\mathrm{E}_\mathcal{G}\{\mathbf{u}_g\} = \frac{1}{|\mathcal{G}|}\sum_{g\in \mathcal{G}}\mathbf{u}_g.\label{eq:v_def1_exp}
\end{align}
In Eq.~(\ref{eq:v_def1_exp}), $\mathrm{E}_\mathcal{D}\{\cdot\}$ and $\mathrm{E}_\mathcal{G}\{\cdot\}$ represent the sample means over the set of all drugs $\mathcal{D}$ and the set of all genes $\mathcal{G}$, respectively. 
However, the definition of $\hat{\mathbf{v}}_\text{naive}$ in Eq.~(\ref{eq:v_def1}) includes the embeddings of unrelated genes and drugs. 
Therefore, a better estimator may be defined by using only the pairs $(d,g)\in\mathcal{R}$.
We consider the estimator $\hat{\mathbf{v}}$ as the mean of the vector differences $\mathbf{u}_g-\mathbf{u}_d$ for $(d, g) \in \mathcal{R}$
as follows:
\begin{align}
    &\hat{\mathbf{v}} :=\mathrm{E}_\mathcal{\mathcal{R}}\{\mathbf{u}_g-\mathbf{u}_d\} =
    \frac{1}{|\mathcal{R}|}\sum_{(d,g)\in \mathcal{R}}(\mathbf{u}_g-\mathbf{u}_d),\label{eq:v_def2}
\end{align}
where $\mathrm{E}_\mathcal{R}\{\cdot\}$ is the sample mean over the set of drug-gene pairs $\mathcal{R}$. 
For easy comparison with Eq.~(\ref{eq:v_def1}), Eq.~(\ref{eq:v_def2}) is rewritten as the difference of mean vectors:
\begin{align}
    &\hat{\mathbf{v}} =
    \mathrm{E}_\mathcal{R}\{\mathbf{u}_g\}-\mathrm{E}_\mathcal{R}\{\mathbf{u}_d\},\label{eq:v_def2_dif}\\
    &\mathrm{E}_\mathcal{\mathcal{R}}\{\mathbf{u}_d\} = \frac{1}{|\mathcal{R}|}\sum_{(d,g)\in \mathcal{R}}\mathbf{u}_d,\quad\mathrm{E}_\mathcal{R}\{\mathbf{u}_g\} = \frac{1}{|\mathcal{R}|}\sum_{(d,g)\in \mathcal{R}}\mathbf{u}_g.\label{eq:v_def2_exp}
\end{align}

To measure the performance of the estimator $\hat{\mathbf{v}}$ in Eq.~(\ref{eq:v_def2}), we prepare the evaluation of the analogy tasks. We define $D$ and $G$ as the sets of drugs and genes contained in $\mathcal{R}$, respectively. 
Specifically, we define $D := \pi_\mathcal{D}(\mathcal{R})$ as the set of drugs $d$ such that $(d, g) \in \mathcal{R}$ for some genes $g$, and $G := \pi_\mathcal{G}(\mathcal{R})$ as the set of genes $g$ such that $(d, g) \in \mathcal{R}$ for some drugs $d$, where the projection operations $\pi_\mathcal{D}$ and $\pi_\mathcal{G}$ are defined as
\begin{align}
    \pi_\mathcal{D}: \mathcal{R} \mapsto \{d \mid (d,g)\in \mathcal{R}\}\subset \mathcal{D},
    \quad \pi_\mathcal{G}: \mathcal{R} \mapsto \{g\mid (d,g)\in \mathcal{R}\}\subset\mathcal{G}.\label{eq:pi}
\end{align}
We also define $[d]\subset \mathcal{G}$ as the set of genes that have drug-gene relations with a drug $d\in D$, and $[g]\subset \mathcal{D}$ as the set of drugs that have drug-gene relations with a gene $g\in G$. These are formally defined as
\begin{align}
    [d] := \{g\mid (d,g)\in \mathcal{R}\}\subset \mathcal{G},\quad[g] := \{d \mid (d,g)\in \mathcal{R}\}\subset \mathcal{D}.\label{eq:quotient}
\end{align}
Given the above, we perform the analogy computation in the following setting. 

\paragraph{Setting G.} 
In the analogy tasks, the set of answer genes for a query drug $d \in D$ is $[d]$. 
The predicted gene is $\hat{g}_d = \argmax_{g\in \mathcal{G}}\cos(\mathbf{u}_d+\hat{\mathbf{v}}, \mathbf{u}_g)$ and if $\hat{g}_d \in [d]$, then the prediction is considered correct.
We define $\hat{g}_d^{(k)}$ as the $k$-th ranked $g \in \mathcal{G}$ based on $\cos(\mathbf{u}_d+\hat{\mathbf{v}}, \mathbf{u}_g)$. 
For the top-$k$ accuracy, if any of the top $k$ predictions $\hat{g}_d^{(1)}, \ldots, \hat{g}_d^{(k)} \in [d]$, then the prediction is considered correct.

Unlike in the basic analogy task, there may be multiple target genes for a single drug. 
While the basic analogy task is one-to-one, we address one-to-many analogy tasks in this study, where a single source may correspond to multiple targets~\cite{kutuzov-etal-2019-one}.
Analogy computation for setting G is illustrated in Fig.~\ref{fig:G}, where the set of answer genes $[d]$ consists of $g_1$, $g_2$, and $g_3$. If the predicted gene $\hat{g}_d$ is one of these three genes, then the prediction is considered correct. 
Note that the analogy tasks for predicting drugs from a query gene can also be defined similarly. See Supplementary Information~\ref{supp:all-setting} for details.

\subsubsection*{Pathway-wise setting}
\begin{table}[t]
\centering
\scriptsize
\begin{tabular}{l|l|l | p{14cm}}
\toprule
Set & Size & Symbol & Elements of the set \\
\midrule
$\mathcal{D}_p$ & 19 & $d$ & Bosutinib, Canertinib dihydrochloride, Dabrafenib, Dacomitinib, Masoprocol, Neratinib, Nilotinib, Osimertinib, Pelitinib, Pimasertib, Poziotinib, Rociletinib, Saracatinib, Selumetinib, Temsirolimus, Tozasertib, Trametinib, Vandetanib, Varlitinib \\
\midrule
$\mathcal{G}_p$ & 84 & $g$ & ABL1, ABL2, AKT1, AKT2, AKT3, ARAF, AREG, BAD, BRAF, BTC, CAMK2A, CAMK2B, CAMK2D, CAMK2G, CBL, CBLB, CDKN1A, CDKN1B, CRK, CRKL, EGF, EGFR, EIF4EBP1, ELK1, ERBB2, ERBB3, ERBB4, EREG, GAB1, GRB2, GSK3B, HBEGF, HRAS, JUN, KRAS, MAP2K1, MAP2K2, MAP2K4, MAP2K7, MAPK1, MAPK10, MAPK3, MAPK8, MAPK9, MTOR, MYC, NCK1, NCK2, NRAS, NRG1, NRG2, NRG3, NRG4, PAK1, PAK2, PAK3, PAK4, PAK5, PAK6, PIK3CA, PIK3CB, PIK3CD, PIK3R1, PIK3R2, PIK3R3, PLCG1, PLCG2, PRKCA, PRKCB, PRKCG, PTK2, RAF1, RPS6KB1, RPS6KB2, SHC1, SHC2, SHC3, SHC4, SOS1, SOS2, SRC, STAT5A, STAT5B, TGFA \\
\midrule
$\mathcal{R}_p$ & 34 & $(d,g)$ & (Bosutinib, ABL1), (Bosutinib, SRC), (Canertinib dihydrochloride, EGFR), (Canertinib dihydrochloride, ERBB2), (Canertinib dihydrochloride, ERBB3), (Canertinib dihydrochloride, ERBB4), (Dabrafenib, BRAF), (Dacomitinib, EGFR), (Dacomitinib, ERBB2), (Dacomitinib, ERBB4), (Masoprocol, EGFR), (Neratinib, EGFR), (Neratinib, ERBB2), (Neratinib, ERBB4), (Nilotinib, ABL1), (Osimertinib, EGFR), (Pelitinib, EGFR), (Pimasertib, MAP2K1), (Pimasertib, MAP2K2), (Poziotinib, EGFR), (Poziotinib, ERBB2), (Rociletinib, EGFR), (Saracatinib, ABL1), (Saracatinib, SRC), (Selumetinib, MAP2K1), (Selumetinib, MAP2K2), (Temsirolimus, MTOR), (Tozasertib, ABL1), (Trametinib, MAP2K1), (Trametinib, MAP2K2), (Vandetanib, EGFR), (Varlitinib, EGFR), (Varlitinib, ERBB2), (Varlitinib, ERBB4) \\
\bottomrule
\end{tabular}
\caption{
$\mathcal{D}_p$, $\mathcal{G}_p$, and $\mathcal{R}_p$ for the pathway $p$, where $p$ is the \emph{ErbB signaling pathway}. 
Note that we only use concepts that exist in BioConceptVec and can be converted to names, so they do not exactly match the actual drugs and genes categorized in the biological pathway. 
In addition, we have changed the notation of some drugs from compounds to more readable names.
}
\label{tab:ErbB}
\end{table}

As a next step, based on the analogy tasks in setting G, we consider more detailed analogy tasks.
To do this, we consider the analogy tasks in the pathway-wise setting, where drugs and genes are categorized using biological pathways.
We define $\mathcal{P}$ as the set of pathways $p$, and $\mathcal{D}_p \subset \mathcal{D}$ and $\mathcal{G}_p \subset \mathcal{G}$ as the sets of drugs and genes categorized in each pathway $p \in \mathcal{P}$, respectively. We then restrict the set $\mathcal{R}$ to each pathway $p$ and define the subset of $\mathcal{R}$ as:
\begin{align}
    \mathcal{R}_p := \{(d,g)\in \mathcal{R} \mid d\in \mathcal{D}_p, g\in \mathcal{G}_p\}\subset\mathcal{D}_p\times\mathcal{G}_p.
\end{align}
A specific example of these sets for the \emph{ErbB signaling pathway} is shown in Table~\ref{tab:ErbB}, and their BioConceptVec skip-gram embeddings are illustrated in Fig.~\ref{fig:hsa04012}.
The vector difference between the mean embeddings of drugs and genes roughly points in the same direction as the vector differences between the embeddings of each drug and its target gene in $\mathcal{R}_p$, although such a two-dimensional illustration should be interpreted with caution.

For drug-gene pairs $(d,g) \in \mathcal{R}_p$, we consider the analogy tasks for predicting the target genes $g$ from a drug $d$. To solve these analogy tasks, we use the relation vector $\mathbf{v}_p$, which represents the relation between drugs and target genes categorized in the same pathway $p$. 
We predict $\mathbf{u}_g$ by adding the relation vector $\mathbf{v}_p$ to $\mathbf{u}_d$, expecting that
\begin{align}
    \mathbf{u}_d+\mathbf{v}_p \approx \mathbf{u}_g.\label{eq:d_plus_vp}
\end{align}
Equation (\ref{eq:d_plus_vp}) corresponds to Eq.~(\ref{eq:d_plus_v}). 
Therefore, similar to the estimator $\hat{\mathbf{v}}$ in Eq.~(\ref{eq:v_def2}),
we define an estimator $\hat{\mathbf{v}}_p$ for the relation vector $\mathbf{v}_p$ 
as the mean of the vector differences $\mathbf{u}_g-\mathbf{u}_d$ for $(d, g) \in \mathcal{R}_p$:
\begin{align}
    &\hat{\mathbf{v}}_p := 
\mathrm{E}_{\mathcal{R}_p}\{\mathbf{u}_g - \mathbf{u}_d\}
= \frac{1}{|\mathcal{R}_p|}\sum_{(d,g)\in \mathcal{R}_p}(\mathbf{u}_g - \mathbf{u}_d),\label{eq:vp_def}
\end{align}
where $\mathrm{E}_{\mathcal{R}_p}\{\cdot\}$ is the sample mean over the set of drug-gene pairs $\mathcal{R}_p$.

To measure the performance of the estimator $\hat{\mathbf{v}}_p$, we prepare the evaluation of the analogy tasks. 
Similar to $D$ and $G$, we define $D_p$ and $G_p$ as the sets of drugs and genes contained in $\mathcal{R}_p$, respectively. 
Using the operations of Eq.~(\ref{eq:pi}), we define $D_p:=\pi_{\mathcal{D}_p}(\mathcal{R}_p)\subset\mathcal{D}_p$ as the set of drugs $d$ such that $(d, g) \in \mathcal{R}_p$ for some genes $g$, and $G_p:=\pi_{\mathcal{G}_p}(\mathcal{R}_p)\subset\mathcal{G}_p$ as the set of genes $g$ such that $(d, g) \in \mathcal{R}_p$ for some drugs $d$.
We also define $[d]_p\subset \mathcal{G}_p$ as the set of genes that have drug-gene relations with a drug $d\in D_p$, and $[g]_p\subset \mathcal{D}_p$ as the set of drugs that have drug-gene relations with a gene $g\in G_p$. 
Similar to Eq.~(\ref{eq:quotient}), these sets are formally defined as
\begin{align}
    [d]_p := \{g\mid (d,g)\in \mathcal{R}_p\}\subset \mathcal{G}_p,\quad
    [g]_p := \{d \mid (d,g)\in \mathcal{R}_p\}\subset \mathcal{D}_p.\label{eq:quotient_p}
\end{align}
Given the above, we perform the analogy computation in the following two settings. 

\paragraph{Setting P1.} 
For the target genes that have drug-gene relations with a drug $d$, only genes categorized in the same pathway $p$ as the drug $d$ are considered correct.
In other words, for a query drug $d \in D_p$, the set of answer genes is $[d]_p$. 
The search space is the set of all genes $\mathcal{G}$, not limited to $\mathcal{G}_p$, the set of genes categorized in the pathway $p$. 
The predicted gene is $\hat{g}_d = \argmax_{g \in \mathcal{G}} \cos(\mathbf{u}_d + \hat{\mathbf{v}}_p, \mathbf{u}_g)$, and if $\hat{g}_d \in [d]_p$, then the prediction is considered correct. 
We define $\hat{g}_d^{(k)}$ as the $k$-th ranked $g\in \mathcal{G}$ based on $\cos(\mathbf{u}_d + \hat{\mathbf{v}}_p, \mathbf{u}_g)$. 
For the top-$k$ accuracy, if $\hat{g}_d^{(k)} \in [d]_p$, then the prediction is considered correct.

\paragraph{Setting P2.} 
The gene predictions $\hat{g}_d$ and  $\hat{g}_d^{(k)}$ are defined exactly the same as those in setting P1, but the answer genes are defined the same as in setting G.
That is, for the target genes that have drug-gene relations with a drug $d$, genes are considered correct regardless of whether they are categorized in the same pathway $p$ as the drug $d$ or not. In other words, for a query drug $d \in D$, the set of answer genes is $[d]$, and the prediction is considered correct if $\hat{g}_d \in [d]$. For the top-$k$ accuracy, if $\hat{g}_d^{(k)} \in [d]$, then the prediction is considered correct.
Note that the experiment is performed for $d \in \mathcal{D}_p \cap D$ for each $p$.

Fig.~\ref{fig:P1-P2} in Supplementary Information~\ref{supp:pathway} shows the differences between settings P1 and P2 using specific examples. 
Table~\ref{tab:drug2gene_setting} in Supplementary Information~\ref{supp:all-setting} summarizes the queries, answer sets, and search spaces in settings G, P1, and P2.

\subsection*{Analogy tasks for drug-gene pairs by year}
In this section, based on analogy tasks for drug-gene pairs, we explain analogy tasks in the setting where datasets are divided by year.
To do so, we first use embeddings trained on PubMed abstracts up to year $y$ and consider analogy tasks in a global setting such as setting G.
Next, we separate the drug-gene relations into ``known'' or ``unknown'' based on their chronological appearance in the PubMed abstracts. 
Using the embeddings from the datasets divided by year, we test whether embeddings of ``known'' relations have the ability to predict ``unknwon'' relations. 
We then consider the pathway-wise settings by year, where drugs and genes are categorized based on pathways in the datasets divided by year.
In these settings, we use the year-specific embeddings and consider analogy tasks in settings such as P1 and P2.
Finally, we evaluate whether ``unknown'' relations can be predicted by ``known'' relations.

\subsubsection*{Global setting by year}
First, using embeddings trained on PubMed abstracts up to year $y$, we consider analogy tasks in a global setting.
In preparation, we define $y_d$ as the year when a drug $d$ first appeared in a PubMed abstract and $y_g$ as the year when a gene $g$ first appeared in a PubMed abstract. 

Consider a fixed year $y$. 
When learning embeddings using PubMed abstracts up to year $y$ as training data, we define $\mathcal{D}^y := \{ d \in \mathcal{D} \mid y_d \le y\}$ and $\mathcal{G}^y := \{ g \in \mathcal{G} \mid y_g \le y\}$ as the sets of drugs and genes that appeared up to year $y$, respectively. 
The set of drug-gene pairs that have drug-gene relations and whose drugs and genes appeared up to year $y$ is expressed as
\begin{align}
    \mathcal{R}^y:=\{(d,g)\in \mathcal{R} \mid d\in \mathcal{D}^y, g\in \mathcal{G}^y\}\subset\mathcal{D}^y\times\mathcal{G}^y.\label{eq:Ry}
\end{align}

Similar to the global setting, for drug-gene pairs $(d,g) \in \mathcal{R}^y$, we consider the analogy tasks for predicting the target genes $g$ from a drug $d$. To solve these analogy tasks, we use the relation vector $\mathbf{v}^y$.
We predict $\mathbf{u}_g$ by adding the relation vector $\mathbf{v}^y$ to $\mathbf{u}_d$:
\begin{align}
    \mathbf{u}_d+\mathbf{v}^{y} \approx \mathbf{u}_g.\label{eq:d_plus_vy}
\end{align}
Equation (\ref{eq:d_plus_vy}) corresponds to Eq.~(\ref{eq:d_plus_v}). 
Therefore, we define the estimator $\hat{\mathbf{v}}^{y}$ for the relation vector $\mathbf{v}^{y}$ by using $\mathcal{R}^{y}$ instead of $\mathcal{R}$ in the estimator $\hat{\mathbf{v}}$ in Eq. (\ref{eq:v_def2}).
Given the above, similar to setting G, we perform the analogy tasks in the following setting. 

\paragraph{Setting Y1.}
Using embeddings trained on PubMed abstracts up to year $y$, we consider analogy tasks in setting G.
Thus, if $y$ is the most recent, it simply corresponds to setting G.

See Supplementary Information~\ref{supp:setting_Y1} for more details on the setting. 

\subsubsection*{Global setting to predict unknown relations by year}
Based on analogy tasks in the global setting by year, we consider analogy tasks to predict unknown relations using known relations.
Specifically, drug-gene relations that appeared up to year $y$ are considered known, while those that appeared after year $y$ are considered unknown. 
We then use embeddings trained on PubMed abstracts up to year $y$, redefine vectors representing known relations, and use these vectors to predict unknown relations.
In preparation, we define $y_{(d,g)}$ as the year when both drug $d$ and gene $g$ first appeared together in a PubMed abstract.
We consider $y_{(d,g)}$ as a substitute for the year when the relation $(d,g)$ was first identified.
By definition, $\max\{y_d, y_g\} \leq y_{(d,g)}$ holds.
The relation $(d,g)$ is interpreted as either known by year $y$ if $y_{(d,g)} \leq y$ or unknown by year $y$ if $y < y_{(d,g)}$.

We define two subsets of $\mathcal{R}^y$ based on whether $y_{(d,g)} \leq y$ or $y < y_{(d,g)}$. 
To do this, we define two intervals, $L_y:=(-\infty, y]$ and $U_y:=(y, \infty)$. Using $L_y$ and $U_y$, we define the subsets $\mathcal{R}^{y\mid L_y}$ and $\mathcal{R}^{y\mid U_y}$ of $\mathcal{R}^y$ as follows:
\begin{align}
    \mathcal{R}^{y\mid L_y}:=\{(d,g) \in \mathcal{R}^y \mid y_{(d,g)} \in L_y \},\,
    \mathcal{R}^{y\mid U_y}:=\{(d,g) \in \mathcal{R}^y \mid y_{(d,g)} \in U_y \}\subset\mathcal{R}^y.\label{eq:RyI}
\end{align}
The set of ``known'' relations is expressed as $\mathcal{R}^{y \mid L_y}$,
and the set of ``unknown'' relations is expressed as $\mathcal{R}^{y \mid U_y}$.
By definition, $\mathcal{R}^{y \mid L_y} \cap \mathcal{R}^{y \mid U_y} = \emptyset$ and $\mathcal{R}^{y \mid L_y} \cup \mathcal{R}^{y \mid U_y}\subset\mathcal{R}^y$.

In analogy tasks, we use ``known'' $\mathcal{R}^{y \mid L_y}$ and then predict the target genes $g$ from a drug $d$ for $(d,g)$ in ``unknown'' $\mathcal{R}^{y \mid U_y}$. 
Using the relation vector $\mathbf{v}^{y\mid L_y}$, which represents the drug-gene relations in $\mathcal{R}^{y \mid L_y}$, we predict $\mathbf{u}_g$ by adding the relation vector $\mathbf{v}^{y\mid L_y}$ to $\mathbf{u}_d$:
\begin{align}
    \mathbf{u}_d+\mathbf{v}^{y\mid L_y} \approx \mathbf{u}_g.\label{eq:d_plus_vy_Ly}
\end{align}
Equation (\ref{eq:d_plus_vy_Ly}) corresponds to Eq.~(\ref{eq:d_plus_vy}).
Therefore, we define the estimator $\hat{\mathbf{v}}^{y\mid L_y}$ for the relation vector $\mathbf{v}^{y\mid L_y}$ by using $\mathcal{R}^{y\mid L_y}$ instead of $\mathcal{R}$ in the estimator $\hat{\mathbf{v}}$ in Eq. (\ref{eq:v_def2}).
Given the above, we perform the analogy tasks in the following setting. 

\paragraph{Setting Y2.}
Using embeddings trained on PubMed abstracts up to year $y$, we consider analogy tasks with a reduced answer set.
For the target genes that have drug-gene relations with a drug $d$, only genes whose relations appeared after year $y$ are considered correct.
In other words, only new discoveries are counted as correct.

See Supplementary Information~\ref{supp:setting_Y2} for more details on the setting. 
Note that $y_d$, $y_g$, and $y_{(d,g)}$ are defined based on the year the drugs, genes, and drug-gene relations appeared in PubMed abstracts. 
Thus, they do not fully correspond to their actual years of discovery, and we only consider the analogy tasks in these hypothetical settings.

\subsubsection*{Pathway-wise setting by year}
As a next step, based on the analogy tasks in settings P1, P2, and Y1, we consider the analogy tasks in the pathway-wise setting by year, where drugs and genes are categorized based on pathways in datasets divided by year. To do this, we perform the analogy tasks in the following two settings. 

\paragraph{Setting P1Y1.}
Using embeddings trained on PubMed abstracts up to year $y$, we consider analogy tasks in setting P1.
Thus, if $y$ is the most recent, it corresponds to setting P1.

\paragraph{Setting P2Y1.}
Using embeddings trained on PubMed abstracts up to year $y$, we consider analogy tasks in setting P2.
Thus, if $y$ is the most recent, it corresponds to setting P2.

See Supplementary Information~\ref{supp:setting_P1Y1_P2Y1} for details on the analogy tasks and these settings.

\subsubsection*{Pathway-wise setting to predict unknown relations by year}
Furthermore, based on the analogy tasks in settings P1, P2, and Y2, we consider analogy tasks to predict unknown relations using known relations in the pathway-wise setting by year.
To do this, we perform the analogy tasks in the following two settings.

\paragraph{Setting P1Y2.}
Using embeddings trained on PubMed abstracts up to year $y$, we consider analogy tasks in setting P1 with a reduced answer set.
For the target genes that have drug-gene relations with a drug $d$, only genes categorized in the same pathway $p$ as $d$, and whose relations appeared after year $y$, are considered correct.

\paragraph{Setting P2Y2.}
Using embeddings trained on PubMed abstracts up to year $y$, we consider analogy tasks in setting P2 with a reduced answer set.
For the target genes that have drug-gene relations with a drug $d$, only genes whose relations appeared after year $y$ are considered correct, regardless of whether they are categorized in the same pathway $p$ as the drug $d$ or not.

See Supplementary Information~\ref{supp:setting_P1Y2_P2Y2} for details on the analogy tasks and these settings.

\subsection*{Embeddings}
\begin{table}[t]
\small
\centering
\begin{tabular}{lrr}
\toprule
 & BioConceptVec & Our embeddings \\
\midrule
$|\mathcal{D}|$ & 117282 & 28284\\
$|\mathcal{G}|$ & 144584 & 51057\\
$|\mathcal{R}|$ & 6645 & 5968\\
\midrule
$|D|$ & 2262 & 1980\\
$|G|$ & 664 & 634\\
\midrule
$\mathrm{E}_{d\in D}\{|[d]|\}$ & 2.938 & 3.014\\
$\mathrm{E}_{g\in G}\{|[g]|\}$ & 10.008 & 9.413\\
\bottomrule
\end{tabular}
\caption{Statistics for setting G.}
\label{tab:stats_analogy}
\end{table}
BioConceptVec~\cite{DBLP:journals/ploscb/ChenLYKWL20} provides four pre-trained 100-dimensional word embeddings: CBOW~\cite{DBLP:journals/corr/abs-1301-3781}, skip-gram~\cite{DBLP:journals/corr/abs-1301-3781,DBLP:conf/nips/MikolovSCCD13}, GloVe~\cite{DBLP:conf/emnlp/PenningtonSM14}, and fastText~\cite{DBLP:journals/tacl/BojanowskiGJM17}. Since CBOW is a simpler model than skip-gram and skip-gram performs better in analogy tasks compared to GloVe~\cite{pmlr-v97-allen19a}, we used BioConceptVec skip-gram embeddings for our experiments. 
Note that fastText can essentially be considered as a skip-gram using $n$-grams. 
To complement the pre-trained BioConceptVec embeddings, we further trained 300-dimensional skip-gram embeddings on the publicly available PubMed abstracts.
As with the original BioConceptVec, to train our embeddings, we used PubTator~\cite{DBLP:journals/nar/WeiKL13} to convert six major biological concepts (genes, mutations, diseases, chemicals, cell lines, and species) in PubMed abstracts into their respective IDs, followed by tokenization using NLTK~\cite{bird-loper-2004-nltk}.
Note that since widely used embeddings such as word2vec and GloVe typically have 300 dimensions, we set the dimensions of our embeddings to 300 instead of the original 100. 
The hyperparameters used to train our embeddings are shown in Table~\ref{tab:param} in Supplementary Information~\ref{supp:embeddings}.

For BioConceptVec and our skip-gram embeddings, Table~\ref{tab:stats_analogy} shows some basic statistics for setting G. 
Due to the difference in training data size and minimal word occurrence, the sizes of certain sets such as $|\mathcal{D}|$ and $|\mathcal{G}|$ differ significantly between BioConceptVec and our skip-gram embeddings, but the sizes of other sets, such as $|\mathcal{R}|$, which represents the size of overall relations, show somewhat similar trends.
Figure~\ref{fig:ansdist_drug2gene} in Supplementary Information~\ref{supp:embeddings} shows the distribution of the sizes of the answer sets for each drug $d$ under settings G, P1, and P2 for both BioConceptVec and our skip-gram embeddings.
Table~\ref{tab:stats_pathway} in Supplementary Information~\ref{supp:pathway} shows the statistics for settings P1 and P2. The hyperparameters used to train our skip-gram are shown in Table~\ref{tab:param} in Supplementary Information~\ref{supp:embeddings}.

\subsection*{Datasets}
\paragraph{Corpus.}
Following Chen et al.~\cite{DBLP:journals/ploscb/ChenLYKWL20}, we used PubMed (\url{https://pubmed.ncbi.nlm.nih.gov/}) abstracts to train our skip-gram embeddings. We used about 35 million abstracts up to the year 2023, while they used about 30 million abstracts.

\paragraph{Drug-gene relations.}
For drug-gene relations, we used publicly available data from AsuratDB~\cite{10.1093/bioinformatics/btac541}, which collects information from various databases including the KEGG~\cite{10.1093/nar/28.1.27,Kanehisa2019_ProteinSci,Kanehisa2025_NAR} database.
KEGG (Kyoto Encyclopedia of Genes and Genomes) is a comprehensive database system that integrates a wide range of bioinformatics information such as genomics, chemical reactions, and biological pathways.

\paragraph{Biological Pathways.}
We obtained a list of human pathways from the KEGG API (\url{https://rest.kegg.jp/list/pathway/hsa}) for use in our experiments. 
For each pathway, we again used the KEGG API to define sets of drugs and genes. 
To avoid oversimplification of analogy tasks, we excluded the pathways where only one type of drug or gene had drug-gene relations. 

For more details, see Data availability section and Supplementary Information~\ref{supp:datasets}.

\subsection*{Baselines}
To evaluate the performance of predicting target genes by adding the relation vector to a drug embedding, we compare it to baseline methods. 

As a simple yet reasonable baseline, we randomly sampled genes from the set $G$ containing the genes that have drug-gene relations with at least one drug. 
To increase the probability of sampling genes with more drug-gene relations, a gene $g \in G$ was sampled with a probability proportional to $|[g]|$, the size of the set of drugs that have \emph{correct drug-gene relations} with $g$. 
In other words, sampling was performed with probabilities proportional to the number of queries that have correct relations.
For simplicity, this sampling method was consistently used across settings G, P1, and P2. We refer to this baseline as the \emph{random baseline}. 
In this baseline, we repeated the experiments 10 times and used the average score as the final result. 
Similar random baselines can also be applied to settings by year and to settings for predicting drugs from genes. 
See Supplementary Information~\ref{supp:random_baseline} for details.

The analogy task we consider can be regarded as a task of predicting relationships between entities. 
Therefore, we adopted Knowledge Graph Embedding (KGE)~\cite{DBLP:journals/pami/AliBHVGSFTL22} as a general baseline method for predicting drug-gene relations. 
KGE learns embeddings for head entities, tail entities, and relations from triplets by leveraging the explicit structure of the knowledge graph.
Using the learned embeddings, it predicts a specific tail corresponding to a given head and relation. 
Among various KGE methods, we used TransE~\cite{DBLP:conf/nips/BordesUGWY13}, one of the most representative approaches, with the embedding dimensions set to the commonly used size of 500.
Similar to the analogy computation, TransE predicts the tail embedding by adding the relation vector to the head embedding. However, while the analogy computation relies solely on document data and pathway information, TransE leverages explicit relations in the knowledge graph to directly learn embeddings and relation vectors.
Word embeddings such as skip-gram are learned from large corpora (e.g., PubMed abstracts) without being tied to any particular research field or specific target task. 
Consequently, they capture broad linguistic and semantic information (not restricted to drug-gene relations), enabling them to be applied to various tasks (e.g., analogy tasks~\cite{analogy-msr}, sentence similarity tasks~\cite{cer-etal-2017-semeval}).
In contrast, KGE is learned based on knowledge and relational information tailored to a specific task, resulting in embeddings specialized for that task.
For details on the dataset splits and hyperparameters used for training TransE, see Supplementary Information~\ref{supp:KGE-setting}.

In addition to this, as strong generative model baselines, we used the Generative Pre-trained Transformer (GPT)~\cite{DBLP:conf/nips/BrownMRSKDNSSAA20} series to predict the top 10 target genes from a query drug by zero-shot. We used the GPT-3.5~\cite{DBLP:conf/nips/Ouyang0JAWMZASR22}, GPT-4~\cite{DBLP:journals/corr/abs-2303-08774}, and GPT-4o (\url{https://platform.openai.com/docs/models/gpt-4o}) models with the temperature hyperparameter set to $0$. 
See Supplementary Information~\ref{supp:GPT-setting} for more details on the models and prompt template for the predictions.

\subsection*{Evaluation metrics}
As evaluation metrics, we use the top-$k$ accuracy explained in each setting, especially the top-1 and top-10 accuracies. We also use Mean Reciprocal Rank (MRR) as another evaluation metric. MRR is a statistical measure to evaluate search performance, expressed as the sum of the reciprocals of the ranks at which the predicted results first appear in the answer set.
Note that, in the random baseline, all genes $g \in G$ are ranked without duplication through sampling with probabilities proportional to $|[g]|$. 
By treating these rankings as prediction results, we can calculate not only top-1 and top-10 accuracies but also MRR.
Since MRR cannot be calculated for the GPT models due to the limited number of prediction candidates output by GPT, only top-1 and top-10 accuracies are calculated.

When evaluating performance using top-1 accuracy, top-10 accuracy, and MRR, the embeddings of all genes $\mathcal{G}$ (or $\mathcal{G}^y$ in settings by year) are centered using the mean embedding $\mathrm{E}_\mathcal{G}\{\mathbf{u}_g\}$ in Eq.~(\ref{eq:v_def1_exp}) (or $\mathrm{E}_{\mathcal{G}^y}\{\mathbf{u}_g\}$). 
Since cosine similarity is affected by the origin, centering mitigates this effect.
In the case of setting G, for example, 
we actually calculate the value of $\cos(\mathbf{u}_d-\mathrm{E}_\mathcal{G}\{\mathbf{u}_g\}+\hat{\mathbf{v}}, \mathbf{u}_g-\mathrm{E}_\mathcal{G}\{\mathbf{u}_g\})$ instead of $\cos(\mathbf{u}_d+\hat{\mathbf{v}}, \mathbf{u}_g)$.

\section*{Results}
\begin{table}[t]
\small
\centering
\begin{tabular}{lllrrr}
\toprule
& & & \multicolumn{3}{c}{Metric}\\
\cmidrule{4-6}
Model &  Setting & Method & Top1 & Top10 & MRR \\
\midrule
\multirow{9}{*}{BioConceptVec} & \multirow{3}{*}{G} & Random & 0.019  & 0.137  & 0.063\\
& & TransE &  0.576 & 0.682 & 0.615\\
 &  & analogy & 0.304 & 0.645 & 0.419\\
 \cmidrule{2-6}
 & \multirow{3}{*}{P1} & Random & 0.018  & 0.140  & 0.063\\
 & & TransE & 0.745 & 0.871 & 0.792\\
 &  & analogy & 0.499 & 0.790 & 0.602\\
 \cmidrule{2-6}
 & \multirow{3}{*}{P2} & Random & 0.022 & 0.162  & 0.072\\
 & & TransE & 0.813 & 0.896 & 0.842\\
 &  & analogy & 0.522 & 0.807 & 0.624\\
\midrule
\multirow{9}{*}{Our embeddings} & \multirow{3}{*}{G} & Random & 0.020  & 0.140  & 0.065\\
 & & TransE & 0.571 & 0.689 & 0.614\\
 &  & analogy & 0.300 & 0.686 & 0.426\\
 \cmidrule{2-6}
 & \multirow{3}{*}{P1} & Random & 0.020  & 0.142  & 0.066\\
 & & TransE & 0.752 & 0.873 & 0.798\\
 &  & analogy & 0.589 & 0.862 & 0.685\\
 \cmidrule{2-6}
 & \multirow{3}{*}{P2} & Random & 0.023 & 0.164  & 0.074\\
 & & TransE & 0.819 & 0.907 & 0.854\\
 &  & analogy & 0.600 & 0.880 & 0.700\\
\midrule
GPT-3.5 & Best of all & zero-shot & 0.653 & 0.821 & -- \\
GPT-4  & Best of all  &zero-shot & 0.710 & 0.832 & -- \\
GPT-4o  & Best of all  &zero-shot & 0.760 & 0.887 & -- \\
\bottomrule
\end{tabular}
\caption{Gene prediction performance in settings G, P1, and P2.
In analogy tasks, the relation vector for prediction is $\hat{\mathbf{v}}$ in setting G and $\hat{\mathbf{v}}_p$ in settings P1 and P2. 
In the random baseline, a gene $g \in G$ is sampled with probabilities proportional to the size of the set of drugs related to $g$. 
TransE is evaluated using a different data split, while for the GPT series, only the best scores across settings are shown in the table.
Since TransE's performance depends on the size of the training data, Fig.~\ref{fig:analogy_vs_transe} compares analogy computation with TransE under varying proportions of the training data.
}
\label{tab:drug2gene_result}
\end{table}

\begin{figure}[t]
\centering
\begin{subfigure}{0.49\linewidth}
\centering
\includegraphics[height=4.5cm,keepaspectratio]{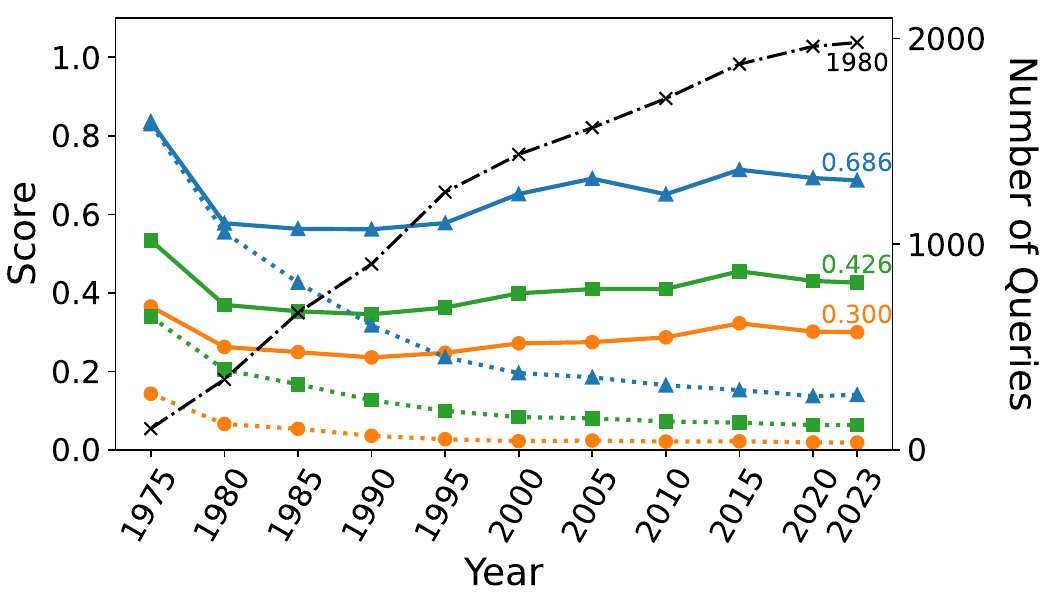}
\caption{Y1}
\label{fig:Y1}
\end{subfigure}\hfill
\begin{subfigure}{0.49\linewidth}
\centering
\includegraphics[height=4.5cm,keepaspectratio]{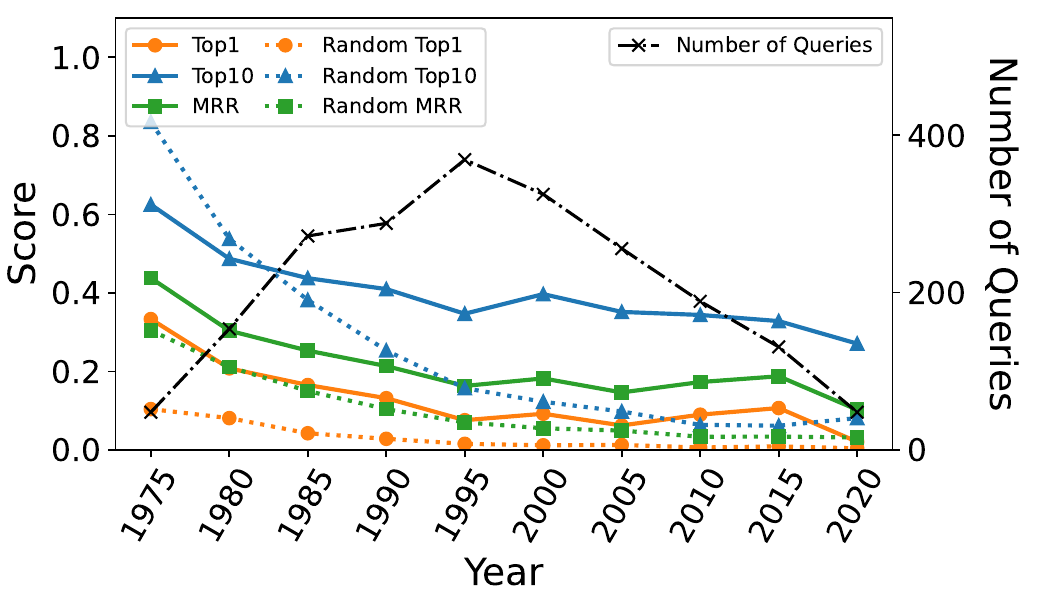}
\caption{Y2}
\label{fig:Y2}
\end{subfigure}
\begin{subfigure}{0.49\linewidth}
\centering
\includegraphics[height=4.5cm,keepaspectratio]{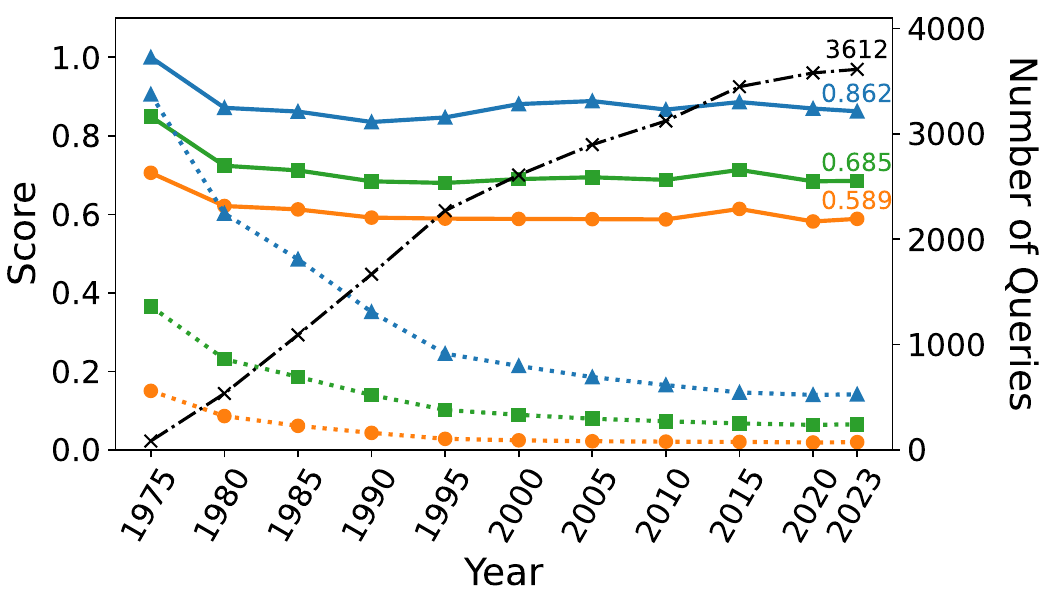}
\caption{P1Y1}
\label{fig:P1Y1}
\end{subfigure}\hfill
\begin{subfigure}{0.49\linewidth}
\centering
\includegraphics[height=4.5cm,keepaspectratio]{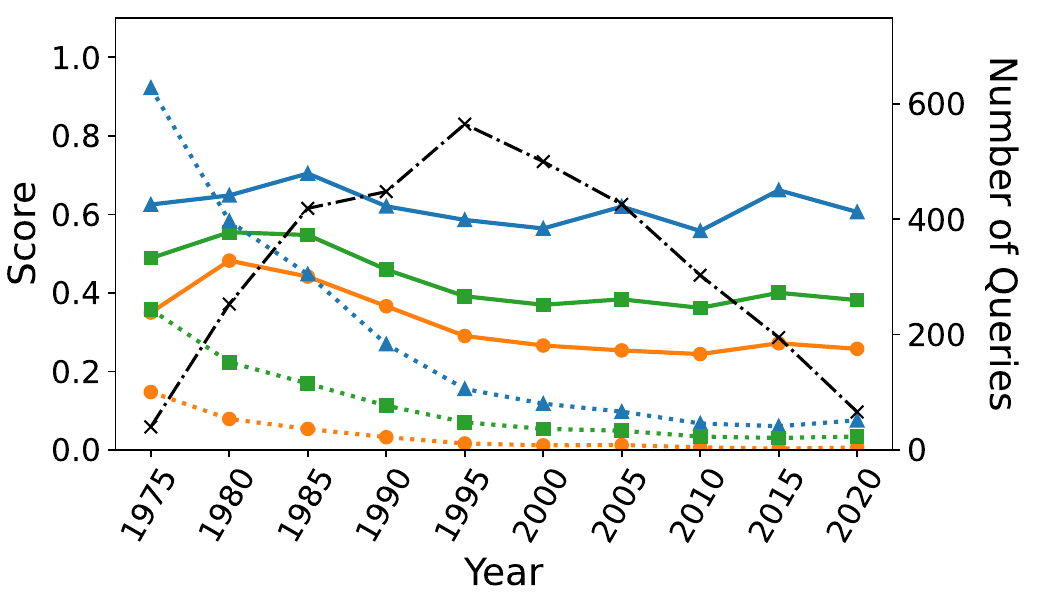}
\caption{P1Y2}
\label{fig:P1Y2}
\end{subfigure}
\caption{Gene prediction performance and the number of queries for each year in settings Y1, Y2, P1Y1, and P1Y2.}
\label{fig:Y1-Y2-P1Y1-P1Y2}
\end{figure}

\subsection*{Gene prediction performance in settings G, P1, and P2}
Table~\ref{tab:drug2gene_result} shows the results of experiments with BioConceptVec and our skip-gram embeddings in settings G, P1, and P2.
For GPT-3.5, GPT-4, and GPT-4o, only the best results are shown; see Supplementary Information~\ref{supp:GPT-result} for detailed results.
In setting G, the prediction by simply adding the global relation vector $\hat{\mathbf{v}}$ resulted in a top-1 accuracy of about 0.3, a top-10 accuracy of over 0.6, and an MRR also over 0.4. 
These results show that the embeddings have the ability to interpret the drug-gene relations.
As with the basic analogy task, these relations were not explicitly provided during the training of the embeddings.

In settings P1 and P2, prediction by adding the pathway-wise relation vector $\hat{\mathbf{v}}_p$ showed better performance than in setting G. 
For example, our skip-gram embeddings achieved a top-1 accuracy of over 0.5 in both settings P1 and P2. 
This is probably because the queries are specific drugs categorized in some pathways, and we use the pathway information to calculate the relation vectors. Since the analogy tasks are considered for each pathway, these settings are also likely to make the tasks easier than those in setting G.

The performance of the random baseline in settings G, P1, and P2 was consistently low across all evaluation metrics. 
Although random sampling was restricted to genes in the set $G$, these results demonstrate that random prediction is challenging in these analogy tasks.
This also confirms the superior performance of prediction by adding relation vectors.
TransE, as a KGE method, naturally outperforms our proposed approach across all three evaluation metrics. Interestingly, however, the top-10 accuracy of the prediction by vector addition is comparable to that of TransE. 
This suggests that our method has the latent potential to achieve performance close to TransE without directly exploiting explicit relational information.
We then hypothesize that the strong performance of TransE observed in Table~\ref{tab:drug2gene_result} depends on the size of the training data. In the Discussion section, we compare analogy computation and TransE under varying proportions of the training data (Fig.~\ref{fig:analogy_vs_transe}).
Among the GPT series, GPT-4o performed the best, followed by GPT-4, which outperformed GPT-3.5. In top-1 accuracy, the GPT series performed better than predictions made by adding relation vectors. However, in top-10 accuracy, the difference becomes smaller. Notably, predictions using $\mathbf{v}_p$ with our skip-gram embeddings in settings P1 and P2 performed comparably to GPT-4o and outperformed both GPT-3.5 and GPT-4.

In addition, our skip-gram outperformed BioConceptVec except for the top-1 accuracy in setting G.
This is probably because BioConceptVec has a dimensionality of 100 and the size of the search space is $|\mathcal{G}|=117282$, while our skip-gram has a dimensionality of 300 and the size of the search space is $|\mathcal{G}|=28284$, as shown in Table~\ref{tab:stats_analogy}.

For the results in Table~\ref{tab:drug2gene_result}, we used the better estimators in equations (\ref{eq:v_def2}) and (\ref{eq:vp_def}), not the naive estimators in equations (\ref{eq:v_def1}) and (\ref{eq:vp_naive_def1}), and calculated the cosine similarities after centering the embeddings of all genes $\mathcal{G}$ (or $\mathcal{G}^y$). 
The results of the naive estimators and the centering ablation study are shown in Table~\ref{tab:drug2gene_naive_centering_diff} in Supplementary Information~\ref{supp:naive-centering}.
The results of the analogy tasks for predicting drugs from a query gene are shown in Table~\ref{tab:gene2drug_result} in Supplementary Information~\ref{supp:gene2drug}.

\subsection*{Gene prediction performance in settings Y1 and Y2}
Figures~\ref{fig:Y1} and~\ref{fig:Y2} show the results of experiments in settings Y1 and Y2 with our skip-gram embeddings trained on the year-specific divisions up to year $y$.
BioConceptVec embeddings are not appropriate for the settings by year, since they are pre-trained on the entire dataset without year-specific divisions.

Setting Y1 is easier than setting Y2, and results in high evaluation metric scores.
For example, from $y=1980$ to $y=2023$, the top-1 accuracy reaches about 0.3, and the top-10 accuracy is about 0.6.
In setting Y1, as in setting G, all relations contained in $\cal{R}$ are considered correct, so all the scores for $y=2023$ are identical to those of setting G. For example, a top-10 accuracy of 0.686 is confirmed in both Fig.~\ref{fig:Y1} and Table~\ref{tab:drug2gene_result}.

In setting Y2, since only genes whose relations appeared after year $y$ are included in the answer set, the tasks are more challenging.
Therefore, the evaluation metric scores are lower compared to those in setting Y1, where genes whose relations appeared up to year $y$ are also included in the answer set.
Yet, in setting Y2, the top-1 accuracy from $y=1985$ to $y=2015$ is about 0.1, the top-10 accuracy is over 0.3, and the MRR is about 0.2. 
Although in setting Y2 the relations to be predicted appeared after year $y$ and are not used to calculate $\hat{\mathbf{v}}^{y\mid L_y}$, these results show that adding $\hat{\mathbf{v}}^{y\mid L_y}$ to $\mathbf{u}_d$ can predict these relations.

In setting Y1, the number of queries increases steadily from older to more recent years, while in setting Y2, the number of queries increases initially and then decreases over time.
In addition, the size of the search space is small for older years (e.g., $|\mathcal{G}^{1975}|=725$) and increases for later years (e.g., $|\mathcal{G}^{2020}|=47509$). As a result, in both settings Y1 and Y2, the datasets for older years such as $y=1975$, and $1980$ have fewer queries and smaller search spaces, resulting in unusually high evaluation metric scores. 
In setting Y2, the datasets for later years have fewer queries and larger search spaces, which tends to produce lower evaluation metric scores.
In both settings, the performance of the random baseline is extremely high for older years such as $y=1975$ and $y=1980$, where the number of genes related to drugs is small. 
However, after $y=1985$, our method consistently outperforms the random baseline. 
For more detailed results for settings Y1 and Y2, see Table~\ref{tab:drug2gene_year_result} in Supplementary Information~\ref{supp:year-results}.

\subsection*{Gene prediction performance in settings P1Y1, P2Y1, P1Y2, and P2Y2}
Figures~\ref{fig:P1Y1} and~\ref{fig:P1Y2} show the results of experiments with our skip-gram embeddings in settings P1Y1 and P1Y2. 
The results in settings P1Y1 and P1Y2 not only follow similar trends to those in settings Y1 and Y2 but also consistently show higher evaluation metric scores.
From the definition of setting P1Y1, the scores for $y=2023$ are identical to those of setting P1.
For example, a top-10 accuracy of 0.862 is confirmed in both Fig.~\ref{fig:P1Y1} and Table~\ref{tab:drug2gene_result}.

The analogy tasks in settings P1Y1 and P1Y2 use detailed pathway information. Therefore, predicting the relations that appeared after year $y$ is easier in settings P1Y1 and P1Y2 than in settings Y1 and Y2.
In addition, the results for the older years show high evaluation metric scores in settings P1Y1 and P1Y2, similar to those in settings Y1 and Y2.
The performance of the random baseline also shows similar trends to those observed in settings Y1 and Y2.

For the results in settings P2Y1 and P2Y2, see Fig.~\ref{fig:P2Y1-P2Y2} in Supplementary Information~\ref{supp:year-pathway-results}. They are very similar to those in settings P1Y1 and P1Y2. For detailed results for settings P1Y1, P2Y1, P1Y2, and P2Y2, see Table~\ref{tab:drug2gene_year_pathway_result} in Supplementary Information~\ref{supp:year-pathway-results}.

\subsection*{Biological Insights from Predicting Drug-Gene Relations}

\begin{table}[t]
\centering
\tiny
\begin{tabular}{@{\hspace{0.5em}}c@{\hspace{0.5em}}c@{\hspace{0.5em}}c@{\hspace{0.5em}}c@{\hspace{0.5em}}c@{\hspace{0.5em}}c@{\hspace{0.5em}}c@{\hspace{0.5em}}c@{\hspace{0.5em}}c@{\hspace{0.5em}}c@{\hspace{0.5em}}c@{\hspace{0.5em}}c@{\hspace{0.5em}}c@{\hspace{0.5em}}}
\toprule
& & \multicolumn{10}{c}{Gene Prediction} \\
Drug & Setting  & Top1 & Top2 & Top3 & Top4 &  Top5 & Top6 & Top7 & Top8 &  Top9 & Top10 &  Answer Set\\
\midrule
\multirow{3}{*}{Bosutinib} & G & \underline{ABL1} & TXK & BCR & BTK & PDGFRB & FYN & FLT3 & HCK & KIT & CSK & $\{\text{ABL1, SRC}\}$ \\
 & P1 & \underline{ABL1} & TXK & EGFR & BCR & ALK & RUNX1 & FLT3 & KIT & JAK2 & ERBB2 & $\{\text{ABL1, SRC}\}$ \\
 & P2 & \underline{ABL1} & TXK & EGFR & BCR & ALK & RUNX1 & FLT3 & KIT & JAK2 & ERBB2 & $\{\text{ABL1, SRC}\}$ \\
\midrule
\multirow{3}{*}{Canertinib dihydrochloride} & G & TTK & CHRM5 & CEP72 & TOP2A & AURKB & PTTG1 & -- & AKAP6 & CENPE & FPGS & $\{\text{EGFR, ERBB2, ERBB3, ERBB4}\}$ \\
 & P1 & \underline{ERBB2} & \underline{EGFR} & BRAF & ALK & MET & CTLA4 & AR & MDM2 & TOP2A & PGR & $\{\text{EGFR, ERBB2, ERBB3, ERBB4}\}$ \\
 & P2 & \underline{ERBB2} & \underline{EGFR} & BRAF & ALK & MET & CTLA4 & AR & MDM2 & TOP2A & PGR & $\{\text{EGFR, ERBB2, ERBB3, ERBB4}\}$ \\
\midrule
\multirow{3}{*}{Dacomitinib} & G & ERBB3 & ROS1 & \underline{EGFR} & FGFR2 & ALK & EML4 & MET & BTK & TXK & \underline{ERBB4} & $\{\text{EGFR, ERBB2, ERBB4}\}$ \\
 & P1 & \underline{EGFR} & ALK & ROS1 & \underline{ERBB2} & ERBB3 & KRAS & MET & BRAF & PIK3CA & TXK & $\{\text{EGFR, ERBB2, ERBB4}\}$ \\
 & P2 & \underline{EGFR} & ALK & ROS1 & \underline{ERBB2} & ERBB3 & KRAS & MET & BRAF & PIK3CA & TXK & $\{\text{EGFR, ERBB2, ERBB4}\}$ \\
\midrule
\multirow{3}{*}{Masoprocol} & G & ALOX5 & ALOX12 & PTGS2 & -- & PTGS1 & ALOX15B & -- & PTGES & COMT & HDAC6 & $\{\text{EGFR, IGF1}\}$ \\
 & P1 & ALOX5 & ERBB2 & TP53 & PTGS2 & \underline{EGFR} & HSP90AA1 & TERT & COX2 & EREG & GSK3A & $\{\text{EGFR}\}$ \\
 & P2 & ALOX5 & ERBB2 & TP53 & PTGS2 & \underline{EGFR} & HSP90AA1 & TERT & COX2 & EREG & GSK3A & $\{\text{EGFR, IGF1}\}$ \\
\midrule
\multirow{3}{*}{Neratinib} & G & ERBB3 & CDK6 & ESR1 & PDGFRA & ROS1 & CDK4 & EREG & \underline{ERBB2} & PIK3CA & \underline{ERBB4} & $\{\text{EGFR, ERBB2, ERBB4}\}$ \\
 & P1 & \underline{ERBB2} & \underline{EGFR} & ESR1 & EREG & ERBB3 & NR4A1 & PGR & ROS1 & BRAF & PIK3CA & $\{\text{EGFR, ERBB2, ERBB4}\}$ \\
 & P2 & \underline{ERBB2} & \underline{EGFR} & ESR1 & EREG & ERBB3 & NR4A1 & PGR & ROS1 & BRAF & PIK3CA & $\{\text{EGFR, ERBB2, ERBB4}\}$ \\
\midrule
\multirow{3}{*}{Nilotinib} & G & \underline{ABL1} & BCR & FYN & \underline{KIT} & \underline{PDGFRA} & FLT3 & TXK & BTK & PDE10A & RUNX1 & $\{\text{ABL1, PDGFRA, PDGFRB, KIT}\}$ \\
 & P1 & \underline{ABL1} & EGFR & ERBB2 & BRAF & ALK & RUNX1 & KIT & KRAS & JAK2 & BCR & $\{\text{ABL1}\}$ \\
 & P2 & \underline{ABL1} & EGFR & ERBB2 & BRAF & ALK & RUNX1 & \underline{KIT} & KRAS & JAK2 & BCR & $\{\text{ABL1, PDGFRA, PDGFRB, KIT}\}$ \\
\midrule
\multirow{3}{*}{Pimasertib} & G & \underline{MAP2K2} & PIK3CB & CDK4 & PIK3R1 & -- & CDK6 & PIK3CD & \underline{MAP2K1} & GRM3 & MAP2K7 & $\{\text{MAP2K1, MAP2K2}\}$ \\
 & P1 & EGFR & BRAF & \underline{MAP2K2} & MAP2K7 & \underline{MAP2K1} & CDK4 & MET & ERBB2 & NRAS & KRAS & $\{\text{MAP2K1, MAP2K2}\}$ \\
 & P2 & EGFR & BRAF & \underline{MAP2K2} & MAP2K7 & \underline{MAP2K1} & CDK4 & MET & ERBB2 & NRAS & KRAS & $\{\text{MAP2K1, MAP2K2}\}$ \\
\midrule
\multirow{3}{*}{Poziotinib} & G & ERBB3 & EML4 & FGFR2 & ROS1 & PIK3CA & ALK & MET & ERBB4 & FGFR4 & NRAS & $\{\text{EGFR, ERBB2}\}$ \\
 & P1 & \underline{EGFR} & \underline{ERBB2} & ERBB3 & ROS1 & ALK & PIK3CA & MET & NRAS & FGFR2 & MAP2K7 & $\{\text{EGFR, ERBB2}\}$ \\
 & P2 & \underline{EGFR} & \underline{ERBB2} & ERBB3 & ROS1 & ALK & PIK3CA & MET & NRAS & FGFR2 & MAP2K7 & $\{\text{EGFR, ERBB2}\}$ \\
\midrule
\multirow{3}{*}{Saracatinib} & G & FYN & DDR1 & ERBB4 & DRD4 & GRM3 & GRIN2B & CHRNA7 & GRIN2A & EPHA3 & SLC6A9 & $\{\text{ABL1, SRC}\}$ \\
 & P1 & EGFR & ERBB4 & ERBB2 & \underline{SRC} & ESR1 & MAPK1 & STAT3 & MET & KDR & PTK2 & $\{\text{ABL1, SRC}\}$ \\
 & P2 & EGFR & ERBB4 & ERBB2 & \underline{SRC} & ESR1 & MAPK1 & STAT3 & MET & KDR & PTK2 & $\{\text{ABL1, SRC}\}$ \\
\midrule
\multirow{3}{*}{Selumetinib} & G & MAP2K7 & \underline{MAP2K2} & CDK4 & RAF1 & PIK3CA & PIK3R1 & \underline{MAP2K1} & CDK6 & NRAS & BRAF & $\{\text{MAP2K1, MAP2K2}\}$ \\
 & P1 & MAP2K7 & BRAF & PIK3CA & KRAS & \underline{MAP2K1} & EGFR & NRAS & CDK4 & MAPK1 & PTEN & $\{\text{MAP2K1, MAP2K2}\}$ \\
 & P2 & MAP2K7 & BRAF & PIK3CA & KRAS & \underline{MAP2K1} & EGFR & NRAS & CDK4 & MAPK1 & PTEN & $\{\text{MAP2K1, MAP2K2}\}$ \\
\midrule
\multirow{3}{*}{Tozasertib} & G & SCN9A & INCENP & CDC25C & \underline{AURKC} & \underline{AURKB} & PLXNA1 & NEK6 & KDM5A & KIF20A & MYT1 & $\{\text{AURKA, AURKB, AURKC, ABL1, FLT3}\}$ \\
 & P1 & AURKA & SRC & CHEK1 & \underline{ABL1} & PLK1 & WEE1 & CDK2 & CASP3 & BRD4 & MAP2K1 & $\{\text{ABL1}\}$ \\
 & P2 & \underline{AURKA} & SRC & CHEK1 & \underline{ABL1} & PLK1 & WEE1 & CDK2 & CASP3 & BRD4 & MAP2K1 & $\{\text{AURKA, AURKB, AURKC, ABL1, FLT3}\}$ \\
\midrule
\multirow{3}{*}{Trametinib} & G & MAP2K7 & \underline{MAP2K2} & CDK6 & CDK4 & \underline{MAP2K1} & RAF1 & MAPK1 & BRAF & MAPK3 & PIK3CA & $\{\text{MAP2K1, MAP2K2}\}$ \\
 & P1 & MAP2K7 & EGFR & BRAF & MAPK1 & KRAS & \underline{MAP2K1} & MAPK3 & CDK4 & CD274 & CDK6 & $\{\text{MAP2K1, MAP2K2}\}$ \\
 & P2 & MAP2K7 & EGFR & BRAF & MAPK1 & KRAS & \underline{MAP2K1} & MAPK3 & CDK4 & CD274 & CDK6 & $\{\text{MAP2K1, MAP2K2}\}$ \\
\midrule
\multirow{3}{*}{Vandetanib} & G & \underline{KDR} & TSHR & \underline{RET} & FLT4 & TXK & VEGFA & IGF1R & FLT1 & AXL & PIK3CA & $\{\text{EGFR, KDR, RET}\}$ \\
 & P1 & \underline{EGFR} & KDR & BRAF & RET & VEGFA & IGF1R & MAPK3 & ERBB2 & MET & MAPK1 & $\{\text{EGFR}\}$ \\
 & P2 & \underline{EGFR} & \underline{KDR} & BRAF & \underline{RET} & VEGFA & IGF1R & MAPK3 & ERBB2 & MET & MAPK1 & $\{\text{EGFR, KDR, RET}\}$ \\
\midrule
\multirow{3}{*}{Varlitinib} & G & ERBB3 & -- & PPFIBP2 & MIR4656 & MIR4323 & MIR1181 & EFNA4 & LY6G6D & CDH10 & SLC39A11 & $\{\text{EGFR, ERBB2, ERBB4}\}$ \\
 & P1 & ERBB3 & \underline{EGFR} & \underline{ERBB2} & \underline{ERBB4} & ESR1 & KRAS & CD274 & MET & EREG & ROS1 & $\{\text{EGFR, ERBB2, ERBB4}\}$ \\
 & P2 & ERBB3 & \underline{EGFR} & \underline{ERBB2} & \underline{ERBB4} & ESR1 & KRAS & CD274 & MET & EREG & ROS1 & $\{\text{EGFR, ERBB2, ERBB4}\}$ \\
\bottomrule
\end{tabular}
\caption{
Gene prediction results for the drugs categorized in the \emph{ErbB signaling pathway}. Shown only for drugs where the size of  $[d]$ is two or more.
For each drug, the top ten predicted genes and their answer sets are shown in settings G, P1, and P2.
Predicted genes included in the answer sets are underlined. Genes whose IDs could not be converted to gene names are indicated by ``--''.
}
\label{tab:biology}
\end{table}

In this section, we investigate the biological insights obtained through analogy tasks by adding the relation vector in settings G, P1, and P2.
For this purpose, we focus on genes and drugs categorized in the \emph{ErbB signaling pathway} as shown in Table~\ref{tab:ErbB} and Fig.~\ref{fig:hsa04012}. 
Furthermore, since our skip-gram embeddings showed better performance compared to the BioConceptVec skip-gram embeddings (Table~\ref{tab:drug2gene_result}), we adopted our embeddings for this analysis. $\mathcal{D}_p$, $\mathcal{G}_p$, and $\mathcal{R}_p$ of our embeddings are identical to those of the BioConceptVec embeddings in Table~\ref{tab:ErbB}.

In Table~\ref{tab:biology}, we have listed the predicted target genes for drugs that are categorized in the \emph{ErbB signaling pathway} and  whose answer set size in setting G is two or more.
For all drugs meeting these criteria, at least one answer gene was included in the top 10 predicted target genes for any of the settings. 
Therefore, it can be said that the prediction of drug-gene relations through analogy tasks functioned appropriately.
Additionally, we provide several examples demonstrating that some genes ranked high in the predictions, even though they were not in the answer set, can still be interpreted biologically. 

First, the target genes of Bosutinib are ABL1 and SRC, which are both known to be non-receptor tyrosine kinases (non-RTKs)~\cite{siveen2018Role}. Among the top 10 predicted target genes, TXK~\cite{maruyama2007Txk} and JAK2~\cite{hu2021JAK} can also be categorized as non-RTKs, sharing several properties with ABL1 and SRC. This implies that the structural and biochemical similarities of these genes may have been reflected in their high-dimensional representations.

For Masoprocol, we were able to predict the correct target gene EGFR as the fifth prediction in settings P1 and P2. However, although EGFR is the sole target gene of Masoprocol according to the information deposited in KEGG, it has also been reported to inhibit lipoxygenase activity~\cite{tappel1953Effect}. Since ALOX5 codes for a lipoxygenase, this prediction should not be considered inaccurate.
The diseases in which Masoprocol is used for its treatment include actinic keratosis~\cite{callen1997Actinic}, and it has been previously known that the mutation of TP53 is involved in the onset of this disease~\cite{park1996P53,brash2006Roles}. Also, the gene that was ranked the highest in this setting, ALOX5, is reported to be one of the transcription targets of TP53~\cite{gilbert20155Lipoxygenase}. 
Together, this information suggests that TP53 is deeply related to the target gene and symptoms of Masoprocol. Therefore, although it may not be a direct target, its high rank in the prediction is justified.

When we set Poziotinib as the query, EGFR and ERBB2 were correctly ranked the highest in settings P1 and P2. However, in setting G, ROS1, EML4, and ALK were also found among the highly ranked targets. Since Poziotinib was initially developed as an effective drug to treat lung cancer with HER2 mutation~\cite{elamin2022Poziotinib}, 
this context appears to be reflected in the embeddings,
as ROS1 and EML4-ALK fusion genes are also characteristic mutations and therapeutic targets in lung cancer~\cite{davies2012Identifying,sasaki2010Biology}.

Selumetinib and Trametinib both target MAP2K1 and MAP2K2, and in both cases these genes were found within the top 10 predictions. Aside from these target genes, we observed BRAF and PI3KCA placed in higher rank among the predicted genes. 
Given the history that these drugs were both developed to treat cancers with BRAF V600 mutation and that combination treatment with inhibition of the PI3K-AKT pathway has been explored~\cite{us2018fda,patel2012Selumetinib,tolcher2011Phase}, we assumed that such background was reflected in these results.

\section*{Discussion}
\subsection*{Connections to trends in existing studies}
Mikolov et al.~\cite{analogy-google} has shown that skip-gram achieves a top-1 accuracy of 0.56 on the Microsoft Research Syntactic Analogies Dataset~\cite{analogy-msr}. Given this, the performance of BioConceptVec and our skip-gram embeddings in the analogy tasks for drug-gene pairs, as presented in Table~\ref{tab:drug2gene_result}, is comparable to that of skip-gram in the basic analogy tasks.
Also, Tshitoyan et al.~\cite{DBLP:journals/nature/TshitoyanDWDRKP19} reported an overall accuracy of 60.1\% when using analogy computation of word embedding to predict 29,046 relationships across various concepts in materials science, including element names, crystal symmetries, and magnetic properties.
Our results for drug-gene relation prediction show a similar level of accuracy, suggesting that the approach of using analogy computation of word embedding can be effective in the biomedical domain as well.
This demonstrates the potential of this method across different scientific disciplines.
However, it's important to note that the accuracy varies significantly across different types of relationships in the materials science study. For instance, predictions for chemical element names showed high accuracy (71.4\%), while crystal structure names had lower accuracy (18.7\%).
This variability suggests that the effectiveness of word embedding analogies may depend on the specific type of relationship being predicted.

\subsection*{Strengths and insights from prediction by adding the relation vector}
\begin{figure}[t!]
    \centering
    \includegraphics[width=\textwidth]{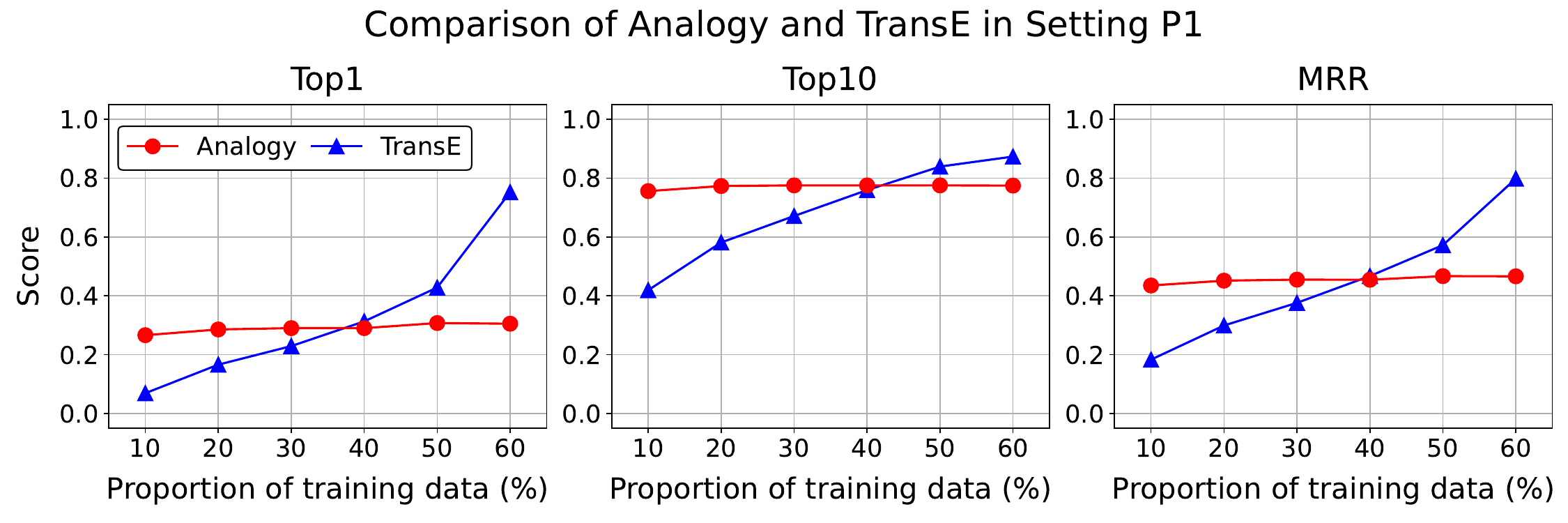}
    \caption{
Changes in evaluation metrics on the test data for analogy computation and TransE under setting P1 as the proportion of training data varies.
Analogy computation maintains consistent performance regardless of the proportion of training data, and outperforms TransE when the proportion of training data is small.
See Supplementary Information~\ref{supp:analogy_vs_transe} for experimental details.}
\label{fig:analogy_vs_transe}
\end{figure}
To test the hypothesis that the superior performance of TransE depends on the size of the training data, we compared the performance of our method with that of TransE in setting P1 by varying the proportion of training data used to compute relation vectors (see Supplementary Information~\ref{supp:analogy_vs_transe} for experimental details). 
As Fig.~\ref{fig:analogy_vs_transe} shows, TransE's performance degrades as the proportion of training data decreases, whereas our method maintains nearly constant performance. 
In particular, our method outperforms TransE when the available training data is small. 
This is because our approach derives relation vectors from already-trained word embeddings, while TransE learns all embeddings from scratch using the training data. 
These findings suggest that our method can function efficiently even with limited datasets.

In addition, in Table~\ref{tab:drug2gene_result}, predictions using $\mathbf{v}_p$ with our skip-gram embeddings in settings P1 and P2 performed on par with GPT-4o and outperformed both GPT-3.5 and GPT-4. 
It is rather surprising that simple vector addition can achieve performance comparable to large language models.
Given that the GPT series is trained on vast amounts of text data,  using relation vector addition is an efficient method. Furthermore, the results showing that target genes can be predicted with performance comparable to or better than large language models imply that high-quality information about drug-gene relations is encoded in the embeddings. This also suggests that the computation of analogies may be a fundamental inference principle within large language models.

Furthermore, we divided the datasets by year and redefined the vectors representing known relations to predict unknown relations.
The experimental results presented in Fig.~\ref{fig:Y1-Y2-P1Y1-P1Y2} demonstrate that our approach can predict unknown future
relations to some extent. The experiments conducted with year-split datasets are a novel attempt not found in existing research, and confirmed the effectiveness of predicting unknown future events.

\subsection*{Biological information intrinsic to the embedding space}
In Table~\ref{tab:biology}, predicted genes in the \emph{ErbB signaling pathway} are closely related to the genes or symptoms that have direct molecular interactions with the target genes, or alternative therapeutic targets for the target disease. This may suggest that the high-dimensional representations of genes and drugs calculated from biomedical texts not only capture the simple drug-gene relations we aimed to predict in this study but also integrate broader, higher-order information.

\subsection*{Limitations}
This study has several limitations.

First, we focus exclusively on drug-gene relations. 
Consequently, the applicability of this method to other biological relationships remains untested.

Additionally, as shown in Table~\ref{tab:drug2gene_result}, our approach for predicting genes from drugs by simply adding relation vectors performs worse than KGE models such as TransE in terms of accuracy. 
While the goal of this study is not to improve performance but to explore the information intrinsic to the embedding space, efforts to improve performance 
remain an important future direction.

Furthermore, the embeddings used in this study are static embeddings learned with skip-gram, which cannot handle out-of-vocabulary concepts. 
Assigning a single embedding to a concept that serves different roles depending on the context (e.g., concepts used differently in the past and present) may also be suboptimal.

\subsection*{Future Directions}
Building on the limitations of this study, several future directions can be considered.

First, since BioConceptVec includes concepts such as diseases and mutations, it would be useful to investigate whether analogy computations using relation vectors can also be applied to relationships beyond drugs and genes. This could serve as a valuable follow-up study.

Additionally, using dynamic embeddings instead of static embeddings represents another promising direction. Since models like BERT~\cite{devlin-etal-2019-bert} compute embeddings based on context, taking advantage of this feature may allow more accurate modeling of drug-gene relations. For example, this approach could help capture relations that vary with context, as well as drugs whose applications evolve over time.

Furthermore, it would be interesting to investigate how the properties of drugs or genes affect performance in analogy tasks.
For example, if the drug-gene relations of a drug have been extensively studied, can related genes for that drug be more easily predicted?
To address this question, we conducted a simple experiment to examine the correlation between the size of the answer set for each drug and the rank of the search results predicted by adding relation vectors. 
Scatter plots of these values for settings G, P1, and P2, using BioConceptVec and our embeddings, are shown in Fig.~\ref{fig:ansfreq_rank} in Supplementary Information~\ref{supp:ansfreq_rank}. 
Mathematically, a larger answer set is expected to make prediction easier, resulting in search result ranks closer to 1.
The results showed a slight negative correlation in setting G, as expected (BioConceptVec: $-0.201$, our embeddings: $-0.196$). However, in settings P1 and P2, little to no correlation was observed (BioConceptVec: $0.054$ and $-0.102$, our embeddings: $0.079$ and $-0.041$). This suggests that in settings P1 and P2, incorporating pathway information into relation vectors helps maintain good search results even when the answer sets are small.
By performing such detailed analyses, we can further deepen our understanding of the biological knowledge intrinsic to the embedding space.

\section*{Conclusions}
In this study, we used embeddings learned from biological texts and performed analogy tasks to predict drug-gene relations. We defined vectors representing these relations and showed that these vectors can accurately predict the target genes for given drugs. Additionally, we categorized drugs and genes based on biological pathways and defined vectors representing drug-gene relations for each pathway. Analogy computations with these vectors showed performance improvement. Our analogy computations demonstrated performance comparable to analogy tasks in other fields and predictions by state-of-the-art large language models, reinforcing the effectiveness of relationship prediction through simple vector addition. 
Moreover, the experiments with year-split datasets demonstrated that it is possible to predict unknown future relations.
Not only were the predictions highly accurate, but the top genes predicted from drugs by our analogy computations were also confirmed to be reasonable from the perspective of biological expertise.

\section*{Data availability}
The datasets generated and analysed during the current study available in the GitHub repository, \url{https://github.com/shimo-lab/Drug-Gene-Analogy}. External resources analysed in this study are publicly available from the following locations: 
AsuratDB \url{https://github.com/keita-iida/ASURATDB/blob/main/genes2bioterm/20221102_human_KEGG_drug.rda};
BioConceptVec \url{https://github.com/ncbi/BioConceptVec}; 
Drug and gene sets for each pathway (e.g. ErbB signalling pathway, hsa04012) \url{https://rest.kegg.jp/get/hsa04012}; 
KEGG pathway list \url{https://rest.kegg.jp/list/pathway/hsa}; 
PubMed abstracts corpus \url{https://ftp.ncbi.nlm.nih.gov/pubmed}; 
word2vec \url{https://code.google.com/archive/p/word2vec/}.
\section*{Code availability}
Our code is available at \url{https://github.com/shimo-lab/Drug-Gene-Analogy}.

\bibliography{article}

\section*{Acknowledgements}
This study was partially supported by JSPS KAKENHI 22H05106, 23H03355, JST CREST JPMJCR21N3.

\section*{Author contributions statement}
H.Y., R.H., K.A., Ma.O. and H.S. conceived the experiments, H.Y., R.H., K.A., Y.Z. and K.M. conducted the experiments, H.Y., R.H., K.A., K.M., S.S., Y.Z., Ma.O., and H.S. analysed the results, R.H. and Mo.O. surveyed existing research. All authors reviewed the manuscript.

\section*{Additional information}
The authors declare that they have no competing interests.

\clearpage
\section*{Supplementary Information}

\setcounter{table}{0}
\renewcommand{\thetable}{S\arabic{table}}

\setcounter{figure}{0}
\renewcommand{\thefigure}{S\arabic{figure}}

\setcounter{equation}{0}
\renewcommand{\theequation}{S\arabic{equation}}

\section{
Details of analogy tasks
}
\subsection{Analogy tasks for drug-gene pairs.}
\subsubsection{Pathway-wise setting}~\label{supp:pathway}
For easy comparison with Eq.~(\ref{eq:v_def1}), Eq.~(\ref{eq:vp_def}) is rewritten as the difference of mean vectors:
\begin{align}
    &\hat{\mathbf{v}}_p = 
    \mathrm{E}_{\mathcal{R}_p}\{\mathbf{u}_g\} - \mathrm{E}_{\mathcal{R}_p}\{\mathbf{u}_d\},\label{eq:vp_def_2}\\
    &\mathrm{E}_{\mathcal{R}_p}\{\mathbf{u}_d\}=\frac{1}{|\mathcal{R}_p|}\sum_{(d,g)\in \mathcal{R}_p}\mathbf{u}_d,
    \quad\mathrm{E}_{\mathcal{R}_p}\{\mathbf{u}_g\}=\frac{1}{|\mathcal{R}_p|}\sum_{(d,g)\in \mathcal{R}_p}\mathbf{u}_g.\label{eq:vp_def_2_exp}
\end{align}

In Eq.~(\ref{eq:vp_def}), the estimator $\hat{\mathbf{v}}_p$ for the relation vector $\mathbf{v}_p$ is calculated from the embeddings of the drug-gene pairs $(d,g) \in \mathcal{R}_p$, while, similar to Eq.~(\ref{eq:v_def1}), the estimator for $\mathbf{v}_p$ can also be calculated from the embeddings of $d \in \mathcal{D}_p$ and $g \in \mathcal{G}_p$. 
The following equation gives a naive estimator for $\mathbf{v}_p$:
\begin{align}
    &\hat{\mathbf{v}}_{p,\text{naive}}:=\mathrm{E}_{\mathcal{G}_p}\{\mathbf{u}_g\} - \mathrm{E}_{\mathcal{D}_p}\{\mathbf{u}_d\},\label{eq:vp_naive_def1}\\
    &\mathrm{E}_{\mathcal{D}_p}\{\mathbf{u}_d\} = \frac{1}{|\mathcal{D}_p|}\sum_{d\in \mathcal{D}_p}\mathbf{u}_d,\quad\mathrm{E}_{\mathcal{G}_p}\{\mathbf{u}_g\} = \frac{1}{|\mathcal{G}_p|}\sum_{g\in \mathcal{G}_p}\mathbf{u}_g.\label{eq:vp_naive_def2}
\end{align}
In Eq.~(\ref{eq:vp_naive_def2}), $\mathrm{E}_{\mathcal{D}_p}\{\cdot\}$ and $\mathrm{E}_{\mathcal{G}_p}\{\cdot\}$ are the means over the set of all drugs $\mathcal{D}$ and the set of all genes $\mathcal{G}$, respectively. 
Equation (\ref{eq:vp_naive_def1}) represents the vector difference between the mean vectors of $\mathcal{D}_p$ and $\mathcal{G}_p$. 
However, similar to $\hat{\mathbf{v}}_\text{naive}$ in Eq.~(\ref{eq:v_def1}), the definition of $\hat{\mathbf{v}}_{p,\text{naive}}$ in Eq.~(\ref{eq:vp_naive_def1}) includes the embeddings of unrelated genes and drugs. 
Therefore, in Supplementary Information~\ref{supp:naive-centering}, we compare $\hat{\mathbf{v}}$ and $\hat{\mathbf{v}}_{\text{naive}}$, as well as $\hat{\mathbf{v}}_p$ and $\hat{\mathbf{v}}_{p,\text{naive}}$.

Note the following in setting P2:
\begin{itemize}
    \item Since $D_p\subset \mathcal{D}_p$ and $ D_p=\pi_{\mathcal{D}_p}(\mathcal{R}_p)=\pi_{\mathcal{D}}(\mathcal{R}_p)\subset\pi_{\mathcal{D}}(\mathcal{R})= D$, it follows that $D_p\subset \mathcal{D}_p\cap D$. For $d \in \mathcal{D}_p \cap D$, there may be $g \in [d] \setminus [d]_p$. Therefore, in Eq.~(\ref{eq:d_plus_vp}), we consider $d \in \mathcal{D}_p \cap D$ and $g \in [d]$, instead of $(d, g) \in \mathcal{R}_p$.
\end{itemize}

\begin{figure*}[t]
  \begin{subfigure}[b]{0.48\textwidth}
    \includegraphics[keepaspectratio, width=\textwidth]{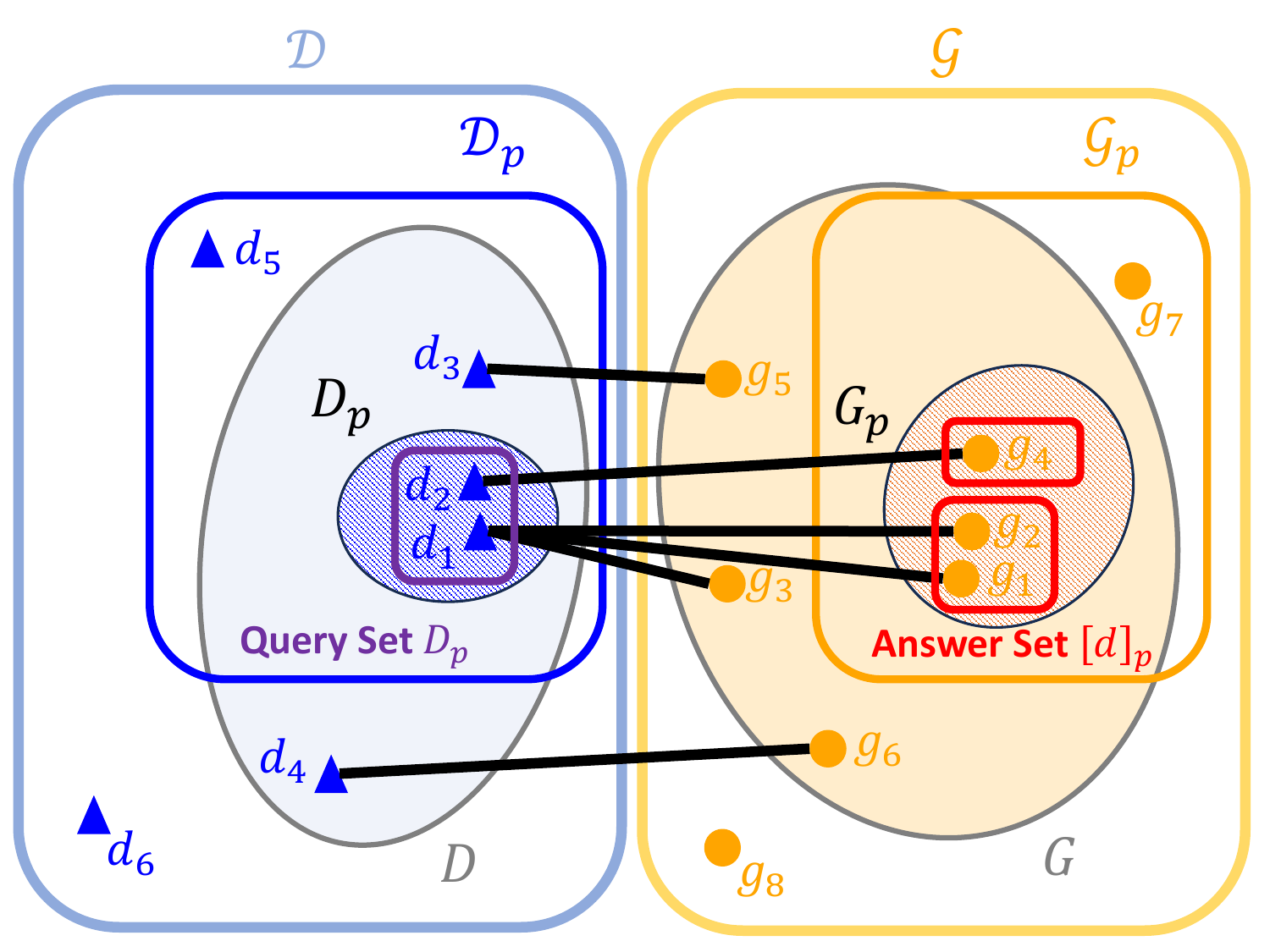}
    \caption{P1}
    \label{fig:P1}
  \end{subfigure}
  \hspace{2mm}
  \begin{subfigure}[b]{0.48\textwidth}
    \includegraphics[keepaspectratio, width=\textwidth]{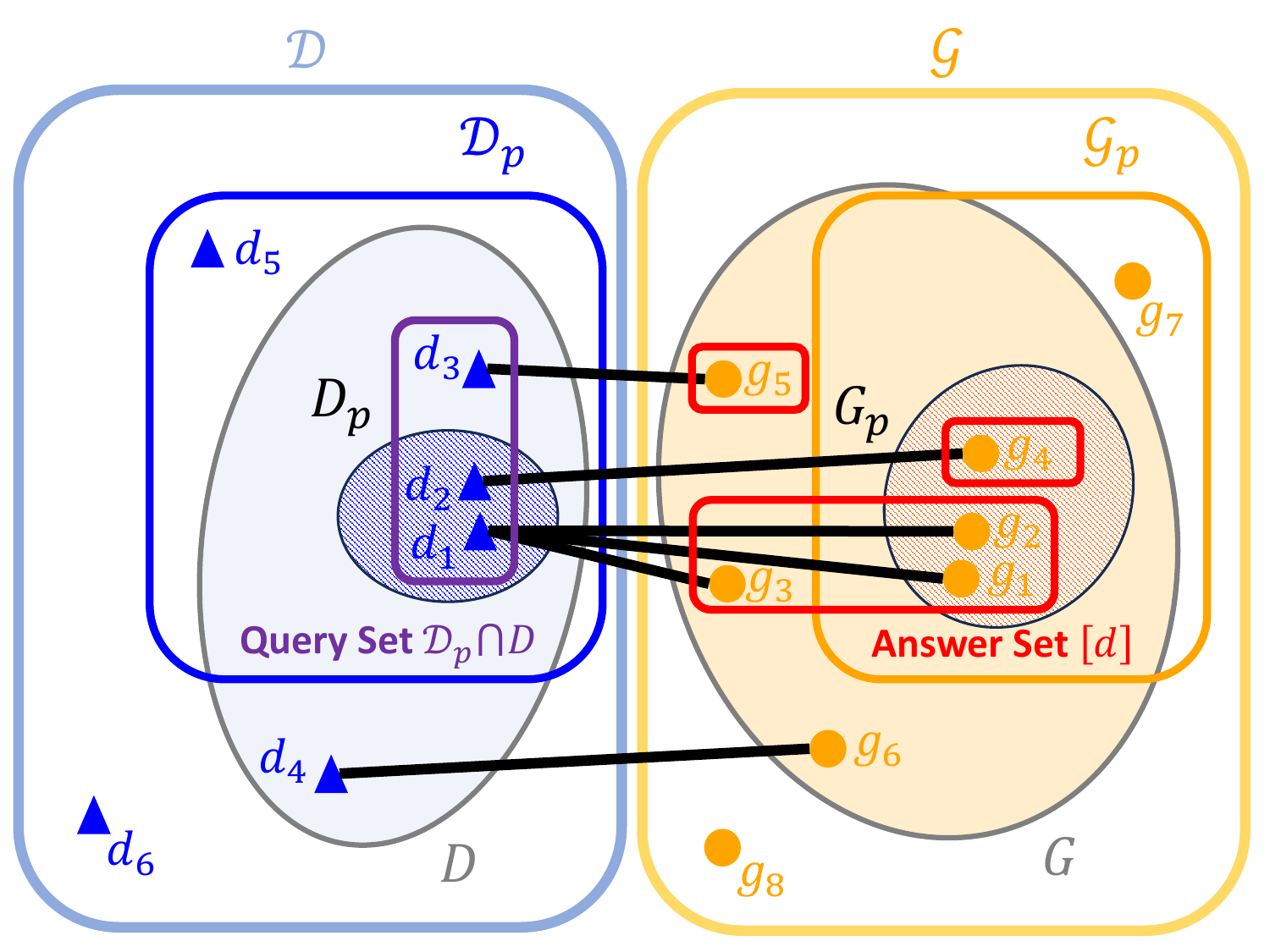}
    \caption{P2}
    \label{fig:P2}
  \end{subfigure}
  \caption{
Differences between settings (a) P1 and (b) P2 using a specific example. 
We set $\mathcal{D}=\{d_1,d_2,d_3,d_4,d_5,d_6\}$, $\mathcal{G}=\{g_1,g_2,g_3,g_4,g_5,g_6,g_7,g_8\}$, and $\mathcal{R}=\{(d_1,g_1),(d_1,g_2),(d_1,g_3),(d_2,g_4),(d_3,g_5),(d_4,g_6)\}$. 
According to the definitions of $D$ and $G$, $D=\{d_1,d_2,d_3,d_4\}$ and $G=\{g_1,g_2,g_3,g_4,g_5,g_6\}$. 
For the pathway $p$, we set $\mathcal{D}_p=\{d_1,d_2,d_3,d_5\}$ and $\mathcal{G}_p=\{g_1,g_2,g_4,g_7\}$. 
According to the definition of $\mathcal{R}_p$, $\mathcal{R}_p=\{(d_1,g_1),(d_1,g_2),(d_2,g_4)\}$. 
The definitions of $D_p$ and $G_p$ then give $D_p=\{d_1,d_2\}$ and $G_p=\{g_1,g_2,g_4\}$, based on $\mathcal{R}_p$. 
In setting P1, the query set is $D_p=\{d_1,d_2\}$, where the set of answer genes for $d_1$ is $[d_1]_p=\{g_1,g_2\}$ and for $d_2$ is $[d_2]_p=\{g_4\}$. 
In setting P2, the query set is $\mathcal{D}_p \cap D=\{d_1,d_2,d_3\}$, where the set of answer genes for $d_1$ is $[d_1]=\{g_1,g_2,g_3\}$, for $d_2$ is $[d_2]=\{g_4\}$, and for $d_3$ is $[d_3]=\{g_5\}$.
}
  \label{fig:P1-P2}
\end{figure*}

Fig.~\ref{fig:P1-P2} shows a specific example to illustrate the differences between settings P1 and P2. 
For drug-gene pairs with drug-gene relations, setting P1 focuses on drugs and genes categorized in the same pathway, while setting P2 considers those genes as well as genes not categorized in the pathway.

\subsection{Analogy tasks for drug-gene pairs by year}
\subsubsection{Global setting by year}\label{supp:setting_Y1}
Similar to the estimator $\hat{\mathbf{v}}$ in Eq.~(\ref{eq:v_def2}), we define an estimator $\hat{\mathbf{v}}^y$ for the relation vector $\mathbf{v}^{y}$ as the mean of the vector differences $\mathbf{u}_g-\mathbf{u}_d$ for $(d, g) \in \mathcal{R}^y$:
\begin{align}
    &\hat{\mathbf{v}}^y :=\mathrm{E}_{\mathcal{R}^y}\{\mathbf{u}_g-\mathbf{u}_d\} =
    \frac{1}{\left|\mathcal{R}^y\right|}\sum_{(d,g)\in \mathcal{R}^y}(\mathbf{u}_g-\mathbf{u}_d),\label{eq:vy_def}
\end{align}
where $\mathrm{E}_{\mathcal{R}^y}\{\cdot\}$ is the sample mean over the set of drug-gene pairs $\mathcal{R}^y$.
For easy comparison with Eq.~(\ref{eq:v_def1}), Eq.~(\ref{eq:vy_def}) is rewritten as the difference of mean vectors:
\begin{align}
    &\hat{\mathbf{v}}^y = 
    \mathrm{E}_{\mathcal{R}^y}\{\mathbf{u}_g\} - \mathrm{E}_{\mathcal{R}^y}\{\mathbf{u}_d\},\label{eq:vy_def_2}\\
    &\mathrm{E}_{\mathcal{R}^y}\{\mathbf{u}_d\}=\quad\frac{1}{|\mathcal{R}^y|}\sum_{(d,g)\in \mathcal{R}^y}\mathbf{u}_d,
    \quad\mathrm{E}_{\mathcal{R}^y}\{\mathbf{u}_g\}=\quad\frac{1}{|\mathcal{R}^y|}\sum_{(d,g)\in \mathcal{R}^y}\mathbf{u}_g.
\end{align}

Similar to $\hat{\mathbf{v}}$ in Eq.~(\ref{eq:v_def2}), to measure the performance of the estimator $\hat{\mathbf{v}}^y$ in Eq.~(\ref{eq:vy_def}), we prepare the evaluation of the analogy tasks.
Using the projection operations in Eq.~(\ref{eq:pi}), we define $D^y:=\pi_{\mathcal{D}^y}(\mathcal{R}^y)\subset\mathcal{D}^y$ and $G^y:=\pi_{\mathcal{G}^y}(\mathcal{R}^y)\subset\mathcal{G}^y$.

Similar to Eq.~(\ref{eq:quotient}), we define $[d]^{y}\subset\mathcal{G}^{y}$ as the set of genes that have drug-gene relations with a drug $d\in D^y$, and $[g]^y\subset\mathcal{D}^{y}$ as the set of drugs that have drug-gene relations with a gene $g\in G^y$. These are formally defined as follows:
\begin{align}
    [d]^y := \{g\mid (d,g)\in \mathcal{R}^{y}\}\subset\mathcal{G}^{y},\quad 
    [g]^y := \{d\mid (d,g)\in \mathcal{R}^{y}\}\subset\mathcal{D}^{y}.\label{eq:quotient_y}
\end{align}
Given the above, we perform the analogy tasks in the following setting. 

\paragraph{Setting Y1.}
In the analogy tasks, the set of answer genes for a query drug $d \in D^y$ is $[d]^y$. 
The predicted gene is $\hat{g}_d = \argmax_{g\in \mathcal{G}^y}\cos(\mathbf{u}_d+\hat{\mathbf{v}}^y, \mathbf{u}_g)$ and if $\hat{g}_d \in [d]^y$, then the prediction is considered correct.
We define $\hat{g}_d^{(k)}$ as the $k$-th ranked $g \in \mathcal{G}^y$ based on $\cos(\mathbf{u}_d+\hat{\mathbf{v}}^y, \mathbf{u}_g)$. 
For the top-$k$ accuracy, if any of the top $k$ predictions $\hat{g}_d^{(1)}, \ldots, \hat{g}_d^{(k)} \in [d]^y$, then the prediction is considered correct.

\subsubsection{Global setting to predict unknown relations by year}\label{supp:setting_Y2}

Similar to the estimator $\hat{\mathbf{v}}$ in Eq.~(\ref{eq:v_def2}), we define an estimator $\hat{\mathbf{v}}^{y\mid L_y}$ for the relation vector $\mathbf{v}^{y\mid L_y}$ as the mean of the vector differences $\mathbf{u}_g-\mathbf{u}_d$ for $(d, g) \in \mathcal{R}^{y \mid L_y}$:
\begin{align}
    &\hat{\mathbf{v}}^{y\mid L_y} :=\mathrm{E}_{\mathcal{R}^{y\mid L_y}}\{\mathbf{u}_g-\mathbf{u}_d\} =
    \frac{1}{\left|\mathcal{R}^{y\mid L_y}\right|}\sum_{(d,g)\in \mathcal{R}^{y\mid L_y}}(\mathbf{u}_g-\mathbf{u}_d),\label{eq:vy_Ly_def}
\end{align}
where $\mathrm{E}_{\mathcal{R}^{y\mid L_y}}\{\cdot\}$ is the sample mean over the set of drug-gene pairs $\mathcal{R}^{y\mid L_y}$.
We define the estimator $\hat{\mathbf{v}}^{y\mid L_y}$ in Eq. (\ref{eq:vy_Ly_def}) by using $\mathcal{R}^{y\mid L_y}$ instead of $\mathcal{R}^y$ in the estimator $\hat{\mathbf{v}}^y$ in Eq. (\ref{eq:vy_def}).
For easy comparison with Eq.~(\ref{eq:v_def1}), Eq.~(\ref{eq:vy_Ly_def}) is rewritten as the difference of mean vectors:
\begin{align}
    &\hat{\mathbf{v}}^{y\mid L_y} = 
    \mathrm{E}_{\mathcal{R}^{y\mid L_y}}\{\mathbf{u}_g\} - \mathrm{E}_{\mathcal{R}^{y\mid L_y}}\{\mathbf{u}_d\},\label{eq:vy_Ly_def_2}\\
    &\mathrm{E}_{\mathcal{R}^{y\mid L_y}}\{\mathbf{u}_d\}=\quad\frac{1}{|\mathcal{R}^{y\mid L_y}|}\sum_{(d,g)\in \mathcal{R}^{y\mid L_y}}\mathbf{u}_d,\quad\mathrm{E}_{\mathcal{R}^{y\mid L_y}}\{\mathbf{u}_g\}=\quad\frac{1}{|\mathcal{R}^{y\mid L_y}|}\sum_{(d,g)\in \mathcal{R}^{y\mid L_y}}\mathbf{u}_g.
\end{align}

Similar to $\hat{\mathbf{v}}$ in Eq.~(\ref{eq:v_def2}), to measure the performance of the estimator $\hat{\mathbf{v}}^{y\mid L_y}$ in Eq.~(\ref{eq:vy_Ly_def}), we prepare the evaluation of the analogy tasks.
For $I\in\{L_{y},U_{y}\}$, using the projection operations in Eq.~(\ref{eq:pi}), we define $D^{y\mid I}:=\pi_{\mathcal{D}^y}(\mathcal{R}^{y\mid I})\subset\mathcal{D}^y$ and $G^{y\mid I}:=\pi_{\mathcal{G}^y}(\mathcal{R}^{y\mid I})\subset\mathcal{G}^y$.

Similar to Eq.~(\ref{eq:quotient}), for $I\in\{L_{y},U_{y}\}$, we define $[d]^{y\mid I}\subset\mathcal{G}^{y}$ as the set of genes that have drug-gene relations with a drug $d\in D^{y\mid I}$, and $[g]^{y\mid I}\subset\mathcal{D}^{y}$ as the set of drugs that have drug-gene relations with a gene $g\in G^{y\mid I}$. These are formally defined as follows:
\begin{align}
    [d]^{y\mid I} := \{g\mid (d,g)\in \mathcal{R}^{y|I}\}\subset\mathcal{G}^{y},\quad 
    [g]^{y\mid I} := \{d\mid (d,g)\in \mathcal{R}^{y|I}\}\subset\mathcal{D}^{y}.\label{eq:quotient_yI}
\end{align}
Given the above, we perform the analogy tasks in the following setting. 

\paragraph{Setting Y2.}
For the target genes that have drug-gene relations with a drug $d$, only genes whose relations appeared after year $y$ are considered correct.
In other words, for a query drug $d\in D^{y\mid U_{y}}$, the set of answer genes is $[d]^{y\mid U_{y}}$. 
The search space is not the set of all genes $\mathcal{G}^y$, but the gene set $\mathcal{G}^y\setminus [d]^{y\mid L_{y}}$. 
The predicted gene is $\hat{g}_d=\argmax_{g\in \mathcal{G}^y\setminus [d]^{y\mid L_y}}\cos(\mathbf{u}_d+\hat{\mathbf{v}}^{y\mid L_y}, \mathbf{u}_g)$, and if $\hat{g}_d\in[d]^{y\mid U_{y}}$, then the prediction is considered correct. 
We define $\hat{g}_d^{(k)}$ as the $k$-th predicted gene, based on $\cos(\mathbf{u}_d+\hat{\mathbf{v}}^{y\mid L_y}, \mathbf{u}_g)$ for $g\in \mathcal{G}^y\setminus [d]^{y\mid L_{y}}$. 
For the top-$k$ accuracy, if $\hat{g}_d^{(k)} \in [d]^{y\mid U_{y}}$, then the prediction is considered correct.

Note the following in setting Y2:
\begin{itemize}
    \item  In setting Y2, the search space is $\mathcal{G}^y \setminus [d]^{y \mid L_{y}}$. This ensures that predicted target genes have the relations that appeared only after year $y$.
\end{itemize}

\begin{table}[t]
\centering
\begin{tabular}{lrr}
\toprule
 & BioConceptVec & Our embeddings\\
\midrule
$|\mathcal{P}|$ & 136 & 129\\
\midrule
$\sum_{p\in\mathcal{P}}|D_p|$ & 4087 & 3612\\
$\sum_{p\in\mathcal{P}}|G_p|$ & 1251 & 1178\\
$\sum_{p\in\mathcal{P}}|\mathcal{D}_p\cap D|$ & 4091 & 3614\\
$\sum_{p\in\mathcal{P}}|\mathcal{G}_p\cap G|$ & 3463 & 3186\\
\midrule
$\mathrm{E}_{p\in\mathcal{P}}\{\mathrm{E}_{d\in D_p}\{|[d]_p|\}\}$ & 1.965 & 1.990\\
$\mathrm{E}_{p\in\mathcal{P}}\{\mathrm{E}_{g\in G_p}\{|[g]_p|\}\}$ & 5.050 & 4.847\\
$\mathrm{E}_{p\in\mathcal{P}}\{\mathrm{E}_{d\in \mathcal{D}_p\cap D}\{|[d]|\}\}$ & 2.738 & 2.776\\
$\mathrm{E}_{p\in\mathcal{P}}\{\mathrm{E}_{g\in \mathcal{G}_p\cap G}\{|[g]|\}\}$ & 7.131 & 6.714\\
\bottomrule
\end{tabular}
\caption{Statistics for settings P1 and P2.}
\label{tab:stats_pathway}
\end{table}

\subsubsection{Pathway-wise setting by year}\label{supp:setting_P1Y1_P2Y1}
Based on the analogy tasks in settings P1, P2, Y1, and Y2, we consider the analogy tasks in the pathway-wise setting by year, where drugs and genes are categorized based on pathways in datasets divided by year.

Consider a fixed year $y$. 
When learning embeddings using PubMed abstracts up to year $y$ as training data, we define $\mathcal{D}_p^y \subset \mathcal{D}^y$ and $\mathcal{G}_p^y \subset \mathcal{G}^y$ as the sets of drugs and genes that are categorized in each pathway $p \in \mathcal{P}$ and appeared up to year $y$, respectively. 
We then restrict the set $\mathcal{R}^y$ in Eq.~(\ref{eq:Ry}) to each pathway $p$ and define the subset of the set $\mathcal{R}_p^y$ as follows:
\begin{align}
    \mathcal{R}_p^y:=\{(d,g)\in \mathcal{R}^y \mid d\in \mathcal{D}_p^y,\,g\in \mathcal{G}_p^y \}\subset \mathcal{D}_p^y\times\mathcal{G}_p^y.\label{eq:Rpy}
\end{align}

For drug-gene pairs $(d,g) \in \mathcal{R}_p^y$, we consider the analogy tasks for predicting the target genes $g$ from a drug $d$. To solve these analogy tasks, we use the relation vector $\mathbf{v}_p^y$, which represents the relation between drugs and target genes categorized in the same pathway $p$. 
We predict $\mathbf{u}_g$ by adding the relation vector $\mathbf{v}_p^y$ to $\mathbf{u}_d$:
\begin{align}
    \mathbf{u}_d+\mathbf{v}_p^{y} \approx \mathbf{u}_g.\label{eq:d_plus_vpy}
\end{align}
Similar to equations (\ref{eq:d_plus_vp}) and (\ref{eq:d_plus_vy}), Eq.~(\ref{eq:d_plus_vpy}) corresponds to Eq.~(\ref{eq:d_plus_v}). 
Therefore, an estimator $\hat{\mathbf{v}}_p^{y}$ for the relation vector $\mathbf{v}_p^{y}$ in Eq.~(\ref{eq:d_plus_vpy}) is defined as follows:
\begin{align}
    &\hat{\mathbf{v}}_p^y :=\mathrm{E}_{\mathcal{R}_p^y}\{\mathbf{u}_g-\mathbf{u}_d\} =
    \frac{1}{\left|\mathcal{R}_p^y\right|}\sum_{(d,g)\in \mathcal{R}_p^y}(\mathbf{u}_g-\mathbf{u}_d).\label{eq:vpy_def}
\end{align}
where $\mathrm{E}_{\mathcal{R}_p^y}\{\cdot\}$ is the sample mean over the set of drug-gene pairs $\mathcal{R}_p^y$. 
Similar to Eq.~(\ref{eq:v_def2}), Eq.~(\ref{eq:vpy_def}) defines the estimator $\hat{\mathbf{v}}_p^y$ as the mean of the vector differences $\mathbf{u}_g-\mathbf{u}_d$ for $(d, g) \in \mathcal{R}_p^y$. 
For easy comparison with Eq.~(\ref{eq:v_def1}), Eq.~(\ref{eq:vpy_def}) is rewritten as the difference of mean vectors:
\begin{align}
    &\hat{\mathbf{v}}_p^y = 
    \mathrm{E}_{\mathcal{R}_p^y}\{\mathbf{u}_g\} - \mathrm{E}_{\mathcal{R}_p^y}\{\mathbf{u}_d\},\label{eq:vpy_def_2}\\
    &\mathrm{E}_{\mathcal{R}_p^y}\{\mathbf{u}_d\}=\quad\frac{1}{|\mathcal{R}_p^y|}\sum_{(d,g)\in \mathcal{R}_p^y}\mathbf{u}_d,\quad\mathrm{E}_{\mathcal{R}_p^y}\{\mathbf{u}_g\}=\quad\frac{1}{|\mathcal{R}_p^y|}\sum_{(d,g)\in \mathcal{R}_p^y}\mathbf{u}_g.
\end{align}

Similar to $\hat{\mathbf{v}}$ in Eq.~(\ref{eq:v_def2}), to measure the performance of the estimator $\hat{\mathbf{v}}_p^y$ in Eq.~(\ref{eq:vpy_def}), we prepare the evaluation of the analogy tasks.
Using the projection operations in Eq.~(\ref{eq:pi}), we define $D_p^y:=\pi_{\mathcal{D}_p^y}(\mathcal{R}_p^y)$ and $G_p^y:=\pi_{\mathcal{G}_p^y}(\mathcal{R}_p^y)$.

Similar to Eq.~(\ref{eq:quotient}), we define $[d]_p^y\subset\mathcal{G}_p^{y}$ as the set of genes that have drug-gene relations with a drug $d\in D_p^y$, and $[g]_p^y\subset\mathcal{D}_p^y$ as the set of drugs that have drug-gene relations with a gene $g\in G_p^y$. 
These are formally defined as follows:
\begin{align}
    [d]_p^y := \{g\mid (d,g)\in \mathcal{R}_p^y\}\subset\mathcal{G}_p^{y},\quad 
    [g]_p^y := \{d\mid (d,g)\in \mathcal{R}_p^y\}\subset\mathcal{D}_p^{y}.\label{eq:quotient_py}
\end{align}
Similar to settings P1, P2, Y1, we perform the analogy tasks in the following four settings. 

\paragraph{Setting P1Y1.}
For the target genes that have drug-gene relations with a drug $d$, only genes categorized in the same pathway $p$ as the drug $d$ are considered correct.
In other words, for a query drug $d \in D_p^y$, the set of answer genes is $[d]_p^y$. 
The search space is the set of all genes $\mathcal{G}^y$, not limited to $\mathcal{G}_p^y$, the set of genes categorized in the pathway $p$. 
The predicted gene is $\hat{g}_d = \argmax_{g \in \mathcal{G}^y} \cos(\mathbf{u}_d + \hat{\mathbf{v}}_p^y, \mathbf{u}_g)$, and if $\hat{g}_d \in [d]_p^y$, then the prediction is considered correct. 
We define $\hat{g}_d^{(k)}$ as the $k$-th ranked $g\in \mathcal{G}^y$ based on $\cos(\mathbf{u}_d + \hat{\mathbf{v}}_p^y, \mathbf{u}_g)$. 
For the top-$k$ accuracy, if $\hat{g}_d^{(k)} \in [d]_p^y$, then the prediction is considered correct.

\paragraph{Setting P2Y1.}
The gene predictions $\hat{g}_d$ and  $\hat{g}_d^{(k)}$ are defined exactly the same as those in setting P1Y1, but the answer genes are defined the same as in setting Y1.
That is, for the target genes that have drug-gene relations with a drug $d$, genes are considered correct regardless of whether they are categorized in the same pathway $p$ as the drug $d$ or not. In other words, for a query drug $d \in D^y$, the set of answer genes is $[d]^y$, and the prediction is considered correct if $\hat{g}_d \in [d]^y$. For the top-$k$ accuracy, if $\hat{g}_d^{(k)} \in [d]^y$, then the prediction is considered correct.
Note that the experiment is performed for $d \in \mathcal{D}_p^y \cap D^y$ for each $p$.

\subsubsection{Pathway-wise setting to predict unknown relations by year}\label{supp:setting_P1Y2_P2Y2}

For $(d,g) \in \mathcal{R}_p^y$, we define two subsets of $\mathcal{R}_p^y$ based on whether $y_{(d,g)} \leq y$ or $y < y_{(d,g)}$.
Similar to $\mathcal{R}^{y\mid L_y}$ and $\mathcal{R}^{y\mid U_y}$ in Eq.~(\ref{eq:RyI}), using $L_y=(-\infty, y]$ and $U_y=(y, \infty)$, we define the set $\mathcal{R}_p^{y\mid L_y}$ and $\mathcal{R}_p^{y\mid U_y}$ as follows:
\begin{align}
    \mathcal{R}^{y|L_y}_p:=\{(d,g) \in \mathcal{R}_p^y \mid y_{(d,g)} \in L_y \}, \mathcal{R}^{y|U_y}_p:=\{(d,g) \in \mathcal{R}_p^y \mid y_{(d,g)} \in U_y \}\subset\mathcal{R}_p^y.\label{eq:RpyI}
\end{align}
The set of drug-gene pairs that satisfies $y_{(d,g)} \leq y$ is $\mathcal{R}_p^{y \mid L_y}$, and the set of drug-gene pairs that satisfies $y < y_{(d,g)}$ is $\mathcal{R}_p^{y \mid U_y}$. 
By definition, $\mathcal{R}_p^{y\mid L_{y}}\cap\mathcal{R}_p^{y\mid U_{y}}=\emptyset$ and $\mathcal{R}^{y|L_y}_p\cup\mathcal{R}^{y|U_y}_p\subset\mathcal{R}_p^{y}$.

In analogy tasks, we use ``known'' $\mathcal{R}_p^{y\mid L_{y}}$ and then predict the target genes $g$ from a drug $d$ for $(d,g)$ in ``unknown'' $\mathcal{R}_p^{y \mid U_y}$. 
Using the vector $\mathbf{v}_p^{y\mid L_y}$, which represents the drug-gene relations and is derived from $\mathcal{R}_p^{y \mid L_y}$, we predict $\mathbf{u}_g$ by adding the relation vector $\mathbf{v}_p^{y\mid L_y}$ to $\mathbf{u}_d$:
\begin{align}
    \mathbf{u}_d+\mathbf{v}_p^{y \mid L_y} \approx \mathbf{u}_g.\label{eq:d_plus_vpy_Ly}
\end{align}
Similar to equations (\ref{eq:d_plus_vp}) and (\ref{eq:d_plus_vy_Ly}), Eq.~(\ref{eq:d_plus_vpy_Ly}) corresponds to Eq.~(\ref{eq:d_plus_v}). 
Therefore, an estimator $\hat{\mathbf{v}}_p^{y\mid L_y}$ for the relation vector $\mathbf{v}_p^{y \mid L_y}$ in Eq.~(\ref{eq:d_plus_vpy_Ly}) is defined as follows:
\begin{align}
    &\hat{\mathbf{v}}_p^{y\mid L_y} :=\mathrm{E}_{\mathcal{R}_p^{y\mid L_y}}\{\mathbf{u}_g-\mathbf{u}_d\} =
    \frac{1}{\left|\mathcal{R}_p^{y\mid L_y}\right|}\sum_{(d,g)\in \mathcal{R}_p^{y\mid L_y}}(\mathbf{u}_g-\mathbf{u}_d).\label{eq:vpy_Ly_def}
\end{align}
where $\mathrm{E}_{\mathcal{R}_p^{y\mid L_y}}\{\cdot\}$ is the sample mean over the set of drug-gene pairs $\mathcal{R}_p^{y\mid L_y}$. 
Similar to Eq.~(\ref{eq:v_def2}), Eq.~(\ref{eq:vpy_Ly_def}) defines the estimator $\hat{\mathbf{v}}_p^{y\mid L_y}$ as the mean of the vector differences $\mathbf{u}_g-\mathbf{u}_d$ for $(d, g) \in \mathcal{R}_p^{y\mid L_y}$. 
For easy comparison with Eq.~(\ref{eq:v_def1}), Eq.~(\ref{eq:vpy_Ly_def}) is rewritten as the difference of mean vectors:
\begin{align}
    &\hat{\mathbf{v}}_p^{y\mid L_y} = 
    \mathrm{E}_{\mathcal{R}_p^{y\mid L_y}}\{\mathbf{u}_g\} - \mathrm{E}_{\mathcal{R}_p^{y\mid L_y}}\{\mathbf{u}_d\},\label{eq:vpy_Ly_def_2}\\
    &\mathrm{E}_{\mathcal{R}_p^{y\mid L_y}}\{\mathbf{u}_d\}=\quad\frac{1}{|\mathcal{R}_p^{y\mid L_y}|}\sum_{(d,g)\in \mathcal{R}_p^{y\mid L_y}}\mathbf{u}_d,\quad\mathrm{E}_{\mathcal{R}_p^{y\mid L_y}}\{\mathbf{u}_g\}=\quad\frac{1}{|\mathcal{R}_p^{y\mid L_y}|}\sum_{(d,g)\in \mathcal{R}_p^{y\mid L_y}}\mathbf{u}_g.
\end{align}

Similar to $\hat{\mathbf{v}}$ in Eq.~(\ref{eq:v_def2}), to measure the performance of the estimator $\hat{\mathbf{v}}_p^{y\mid L_y}$ in Eq.~(\ref{eq:vpy_Ly_def}), we prepare the evaluation of the analogy tasks.
For $I\in\{L_{y},U_{y}\}$, using the projection operations in Eq.~(\ref{eq:pi}), we define $D_p^{y\mid I}:=\pi_{\mathcal{D}_p^y}(\mathcal{R}_p^{y\mid I})$ and $G_p^{y\mid I}:=\pi_{\mathcal{G}_p^y}(\mathcal{R}_p^{y\mid I})$.

Similar to Eq.~(\ref{eq:quotient}), for $I\in\{L_{y},U_{y}\}$, we define $[d]_p^{y\mid I}\subset\mathcal{G}_p^{y}$ as the set of genes that have drug-gene relations with a drug $d\in D_p^{y\mid I}$, and $[g]_p^{y\mid I}\subset\mathcal{D}_p^{y}$ as the set of drugs that have drug-gene relations with a gene $g\in G_p^{y\mid I}$. 
These are formally defined as follows:
\begin{align}
    [d]_p^{y\mid I} := \{g\mid (d,g)\in \mathcal{R}_p^{y\mid I}\}\subset\mathcal{G}_p^{y},\quad 
    [g]_p^{y\mid I} := \{d\mid (d,g)\in \mathcal{R}_p^{y\mid I}\}\subset\mathcal{D}_p^{y}.\label{eq:quotient_pyI}
\end{align}
Similar to settings P1, P2, Y2, we perform the analogy tasks in the following four settings. 

\paragraph{Setting P1Y2.} 
For the target genes that have drug-gene relations with a drug $d$, only genes that are categorized in the same pathway $p$ as $d$ and whose relations appeared after year $y$ are considered correct.
In other words, for a query drug $d\in D_p^{y\mid U_{y}}$, the set of answer genes is $[d]_p^{y\mid U_{y}}$. 
The search space is not the set of all genes $\mathcal{G}^y$, but the gene set $\mathcal{G}^y\setminus [d]_p^{y\mid L_{y}}$. 
The predicted gene is $\hat{g}_d=\argmax_{g\in \mathcal{G}^y\setminus [d]_p^{y\mid L_y}}\cos(\mathbf{u}_d+\hat{\mathbf{v}}_p^{y\mid L_y}, \mathbf{u}_g)$, and if $\hat{g}_d\in[d]_p^{y\mid U_{y}}$, then the prediction is considered correct. 
We define $\hat{g}_d^{(k)}$ as the $k$-th predicted gene, based on $\cos(\mathbf{u}_d+\hat{\mathbf{v}}_p^{y\mid L_y}, \mathbf{u}_g)$ for $g\in \mathcal{G}^y\setminus [d]_p^{y\mid L_{y}}$. 
For the top-$k$ accuracy, if $\hat{g}_d^{(k)} \in [d]_p^{y\mid U_{y}}$, then the prediction is considered correct.

\paragraph{Setting P2Y2.} 
For the target genes that have drug-gene relations with a drug $d$, only genes whose relations appeared after year $y$ are considered correct, regardless of whether they are categorized in the same pathway $p$ as the drug $d$ or not.
In other words, for a query drug $d\in \mathcal{D}_p^y\cap D^{y\mid U_{y}}$, the set of answer genes is $[d]^{y\mid U_{y}}$.
The search space is not the set of all genes $\mathcal{G}^y$, but the gene set $\mathcal{G}^y\setminus [d]^{y\mid L_{y}}$. 
The predicted gene is $\hat{g}_d=\argmax_{g\in \mathcal{G}^y\setminus [d]^{y\mid L_y}}\cos(\mathbf{u}_d+\hat{\mathbf{v}}_p^{y \mid L_y}, \mathbf{u}_g)$, and if $\hat{g}_d\in[d]^{y\mid U_{y}}$, then the prediction is considered correct. 
We define $\hat{g}_d^{(k)}$ as the $k$-th predicted gene, based on $\cos(\mathbf{u}_d+\hat{\mathbf{v}}_p^{y \mid L_y}, \mathbf{u}_g)$ for $g\in \mathcal{G}^y\setminus [d]^{y\mid L_y}$. 
For the top-$k$ accuracy, if $\hat{g}_d^{(k)} \in [d]^{y\mid U_{y}}$, then the prediction is considered correct.

Note the following in settings P1Y2 and P2Y2:
\begin{itemize}
\item In setting P1Y2, the search space is $\mathcal{G}^y\setminus [d]_p^{y\mid L_{y}}$. 
This ensures that predicted target genes, which are categorized in the same pathway $p$ as the drug $d$, have the relations that appeared only after year $y$.
\item In setting P2Y2, since $D_p^{y\mid U_{y}}\subset\mathcal{D}_p^y$ and $ D_p^{y\mid U_{y}}=\pi_{\mathcal{D}_p^y}\left(\mathcal{R}_p^{y\mid U_{y}}\right)=\pi_{\mathcal{D}^y}\left(\mathcal{R}_p^{y\mid U_{y}}\right)\subset\pi_{\mathcal{D}^y}\left(\mathcal{R}^{y\mid U_{y}}\right)=D^{y\mid U_{y}}$, it follows that $D_p^{y\mid U_{y}}\subset\mathcal{D}_p^y\cap D^{y\mid U_{y}}$. 
For $d\in \mathcal{D}_p^y\cap D^{y\mid U_{y}}$, there may be $g\in [d]^{y\mid U_{y}}\setminus [d]_p^{y\mid U_{y}}$. 
Therefore, in Eq.~(\ref{eq:d_plus_vpy}), we consider $d\in \mathcal{D}_p^y\cap D^{y\mid U_{y}}$ and $g\in [d]^{y\mid U_{y}}$, instead of $(d,g)\in \mathcal{R}_p^{y\mid U_{y}}$.
The search space is $\mathcal{G}^y\setminus [d]^{y\mid L_{y}}$.
This ensure that predicted target genes have the relations that appeared only after year $y$.
\end{itemize}

\subsection{Comparison of experimental settings}\label{supp:all-setting}

\begin{table}[t]
\centering
\begin{tabular}{llll}
\toprule
Setting &  Query & Answer Set & Search Space\\
\midrule
G & $d\in D$ & $[d]$ & $\mathcal{G}$ \\
\midrule
P1 & $d\in D_p$ & $[d]_p$ & $\mathcal{G}$\\
P2 &  $d\in \mathcal{D}_p \cap D$ & $[d]$ & $\mathcal{G}$\\
\midrule
Y1 & $d\in D^{y}$ & $[d]^{y}$ & $\mathcal{G}^y$ \\
Y2 & $d\in D^{y\mid U_{y}}$ & $[d]^{y\mid U_{y}}$ & $\mathcal{G}^y\setminus [d]^{y\mid L_{y}}$ \\
\midrule
P1Y1 & $d\in D_p^{y}$ & $[d]_p^y$ & $\mathcal{G}^y$ \\
P2Y1 & $d\in \mathcal{D}_p^y\cap D^{y}$ & $[d]^{y}$ & $\mathcal{G}^y$ \\
P1Y2 & $d\in D_p^{y\mid U_{y}}$ & $[d]_p^{y\mid U_{y}}$ & $\mathcal{G}^y\setminus [d]_p^{y\mid L_{y}}$ \\
P2Y2 & $d\in \mathcal{D}_p^y\cap D^{y\mid U_{y}}$ & $[d]^{y\mid U_{y}}$ & $\mathcal{G}^y\setminus [d]^{y\mid L_{y}}$ \\
\bottomrule
\end{tabular}
\caption{
Query, answer set, and search space for each setting for predicting genes from drugs.
}
\label{tab:drug2gene_setting}
\end{table}

\begin{table}[t]
\centering
\begin{tabular}{llll}
\toprule
Setting &  Query & Answer Set & Search Space\\
\midrule
G$^{'}$ & $g\in G$ & $[g]$ & $\mathcal{D}$ \\
\midrule
P1$^{'}$ & $g\in G_p$ & $[g]_p$ & $\mathcal{D}$\\
P2$^{'}$ &  $g\in \mathcal{G}_p \cap G$ & $[g]$ & $\mathcal{D}$\\
\bottomrule
\end{tabular}
\caption{
Query, answer set, and search space for each setting for predicting drugs from genes.
}
\label{tab:gene2drug_setting}
\end{table}

\begin{table}[t]
\centering
\begin{minipage}[b]{0.48\linewidth}
\centering
\begin{tabular}{lr}
\toprule
Hyperparameter & Values \\
\midrule
Training epochs & $10$ \\
Down-sampling threshold & $10^{-5}$ \\
Learning rate & $0.025$ \\
Window size & $5$ \\
Negative samples & $5$ \\
Minimal word occurrence &$30$ \\
Dimension & $300$ \\
\bottomrule
\end{tabular}
\caption{Hyperparameter for our skip-gram.}
\label{tab:param}
\end{minipage}
\hspace{2mm}
\begin{minipage}[b]{0.48\linewidth}
\centering
\end{minipage}
\end{table}

Table~\ref{tab:drug2gene_setting} shows the settings for predicting genes from drugs, as used in the main text.
Similarly, Table~\ref{tab:gene2drug_setting} shows the settings for predicting drugs from genes. 
In Table~\ref{tab:gene2drug_setting}, we adapted the notations used for predicting genes from drugs in Table~\ref{tab:drug2gene_setting} to those used for predicting drugs from genes, denoting them as G$^{'}$, P1$^{'}$, and P2$^{'}$.

\subsection{Embeddings}\label{supp:embeddings}
\begin{figure}[!t]
\centering
\begin{subfigure}{\textwidth}
    \includegraphics[width=\textwidth]{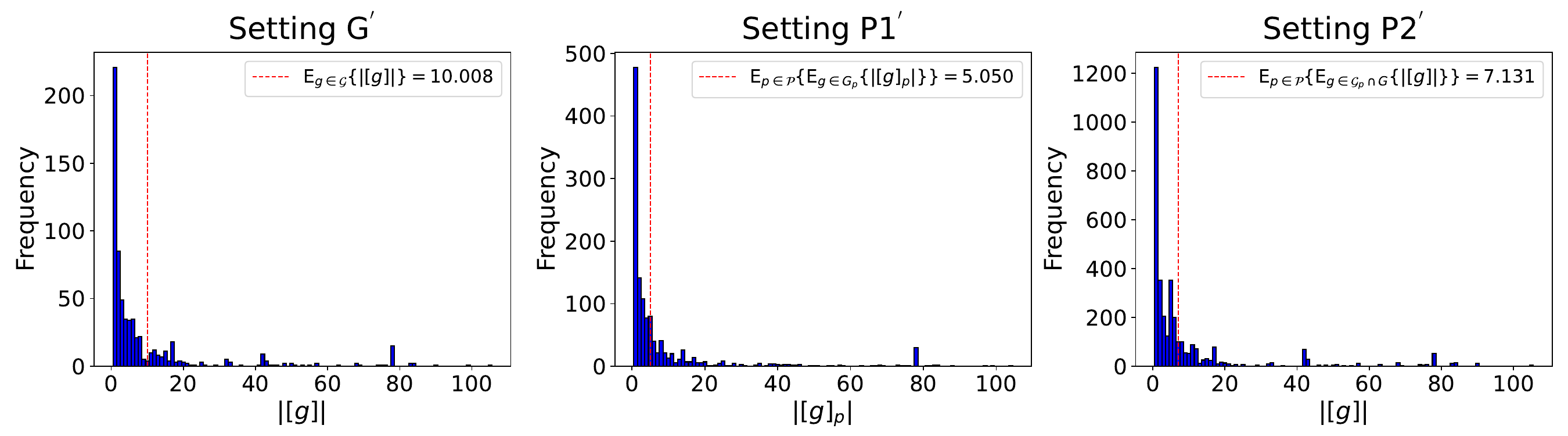}
    \caption{BioConceptVec}
    \label{fig:ansdist_gene2drug_bcv}
\end{subfigure}
\begin{subfigure}{\textwidth}
    \includegraphics[width=\textwidth]{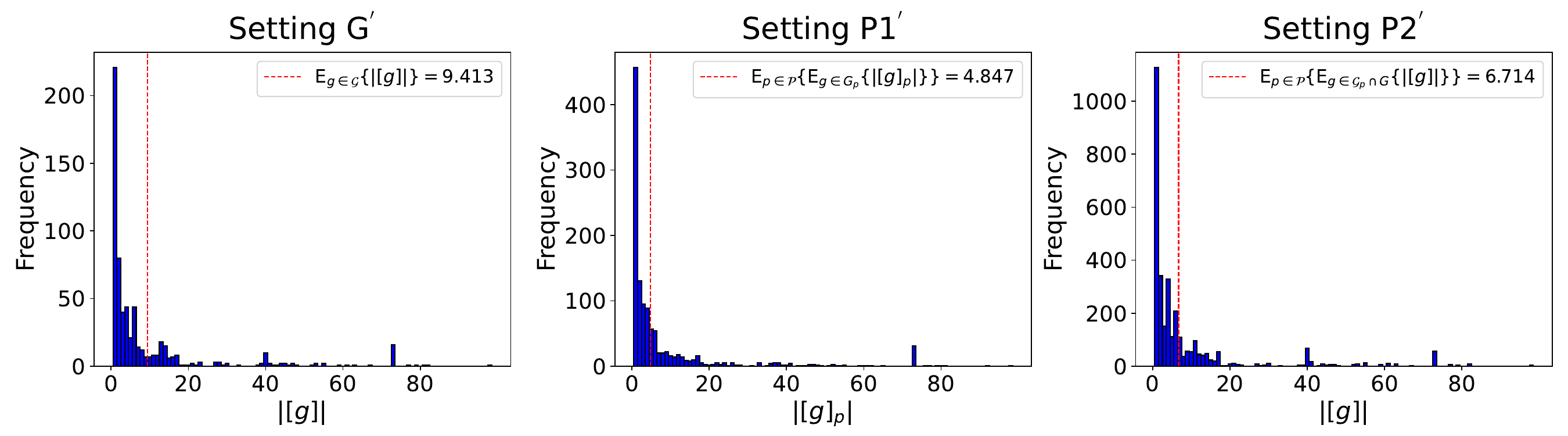}
    \caption{Our embeddings}
    \label{fig:ansdist_gene2drug_homemade}
\end{subfigure}
\caption{
Distribution of the sizes of the answer sets for each gene $g$ in the settings for predicting drugs from genes.
The mean values are also shown: for setting G$^{'}$, the value of $\mathrm{E}_{g\in G}\{|[g]|\}$ from Table~\ref{tab:stats_analogy}; for setting P1$^{'}$, the value of $\mathrm{E}_{p\in \mathcal{P}}\{\mathrm{E}_{g\in G_p}\{|[g]_p|\}\}$ from Table~\ref{tab:stats_pathway}; and for setting P2$^{'}$, the value of $\mathrm{E}_{p\in \mathcal{P}}\{\mathrm{E}_{g\in \mathcal{G}_p\cap G}\{|[g]|\}\}$ from Table~\ref{tab:stats_pathway}.
}
\label{fig:ansdist_gene2drug}
\end{figure}

Figure~\ref{fig:ansdist_drug2gene} shows the distribution of the sizes of the answer sets for each drug $d$ in settings G, P1, and P2 for BioConceptVec and our skip-gram embeddings. 
Similarly, Fig.~\ref{fig:ansdist_gene2drug} shows the distribution of the sizes of the answer sets for each gene $g$ in settings G$^{'}$, P1$^{'}$, and P2$^{'}$ for BioConceptVec and our skip-gram embeddings.
The hyperparameters used to train our skip-gram are shown in Table~\ref{tab:param}.

\subsection{Datasets}\label{supp:datasets}
In the BioConceptVec vocabulary, genes are represented by gene IDs\cite{maglott2005entrez} and drugs are represented by MeSH (Medical Subject Headings, \url{https://www.nlm.nih.gov/mesh/meshhome.html}) IDs. Therefore, a conversion from IDs to names is necessary when performing experiments.
In addition, in the data obtained from AsuratDB and the KEGG API, genes are represented by gene IDs and drugs are represented by KEGG IDs. Therefore, a conversion from KEGG IDs to MeSH IDs is also required for use with BioConceptVec. The procedures for converting these IDs are described in the following sections.

\subsubsection{Conversion from MeSH ID to drug name}\label{supp:meshid_to_drugname}
The BioConceptVec vocabulary registers drugs using MeSH (Medical Subject Headings, \url{https://www.nlm.nih.gov/mesh/meshhome.html}) IDs. 
This is due to the normalization of drug names using MeSH IDs in PubTator. 
Therefore, we explain the procedure for converting MeSH IDs to drug names.
We used MeSH SPARQL (\url{https://hhs.github.io/meshrdf/sparql-and-uri-requests}) to obtain the MeSH headings corresponding to the MeSH IDs in the BioConceptVec vocabulary. 
We used these headings to convert MeSH IDs to drug names.
We also applied this conversion to drugs in the vocabulary of our trained skip-gram model.

\begin{figure}[!t]
\centering
\begin{subfigure}{\textwidth}
    \includegraphics[width=\textwidth]{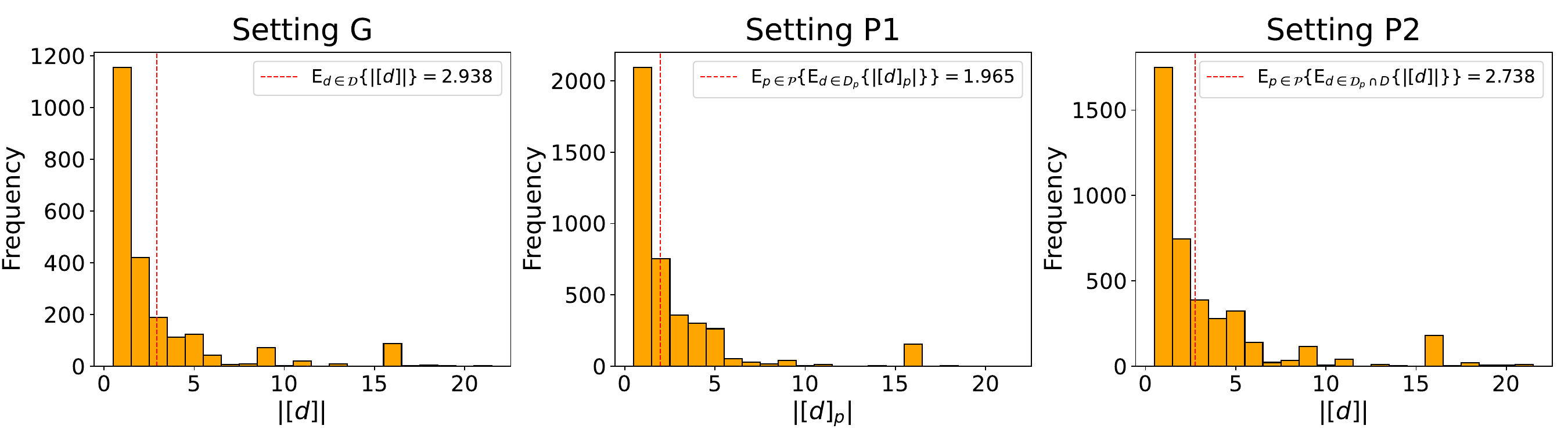}
    \caption{BioConceptVec}
    \label{fig:ansdist_drug2gene_bcv}
\end{subfigure}
\begin{subfigure}{\textwidth}
    \includegraphics[width=\textwidth]{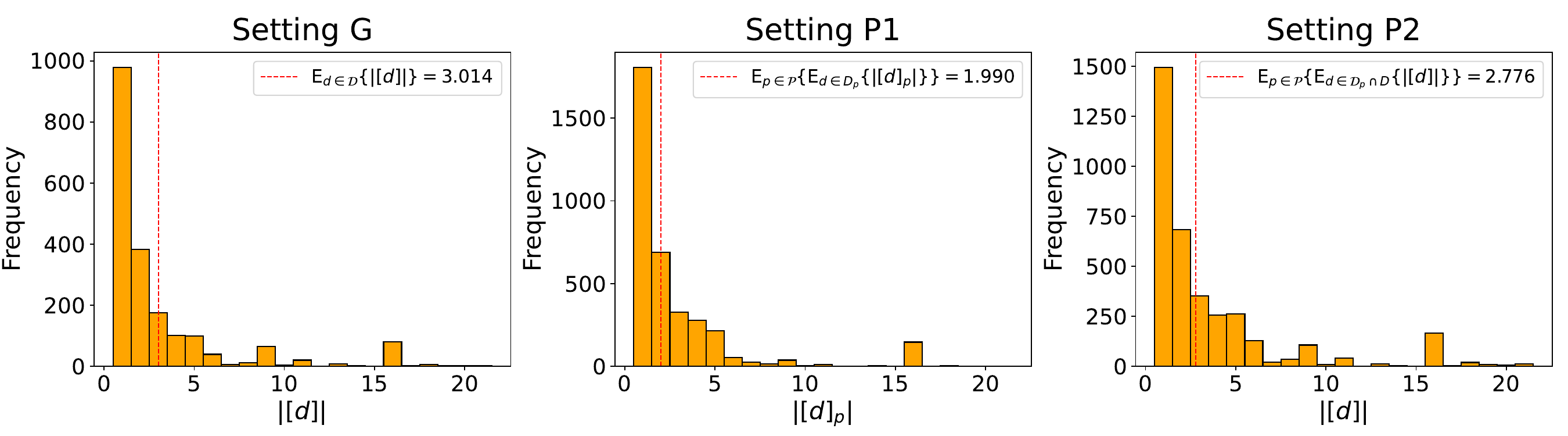}
    \caption{Our embeddings}
    \label{fig:ansdist_drug2gene_homemade}
\end{subfigure}
\caption{
Distribution of the sizes of the answer sets for each drug $d$ in the settings for predicting genes from drugs.
The mean values are also shown: for setting G, the value of $\mathrm{E}_{d\in D}\{|[d]|\}$ from Table~\ref{tab:stats_analogy}; for setting P1, the value of $\mathrm{E}_{p\in \mathcal{P}}\{\mathrm{E}_{d\in D_p}\{|[d]_p|\}\}$ from Table~\ref{tab:stats_pathway}; and for setting P2, the value of $\mathrm{E}_{p\in \mathcal{P}}\{\mathrm{E}_{d\in \mathcal{D}_p\cap D}\{|[d]|\}\}$ from Table~\ref{tab:stats_pathway}.
}
\label{fig:ansdist_drug2gene}
\end{figure}

\begin{table}[t]
\centering
\begin{tabular}{ll}
\toprule
Model & Version\\
\midrule
GPT-3.5 & gpt-3.5-turbo-0125\\
GPT-4 & gpt-4-turbo-2024-04-09\\
GPT-4o & gpt-4o-2024-05-13 \\
\bottomrule
\end{tabular}
\caption{Version of each GPT model.}
\label{tab:GPT-models}
\end{table}

\begin{table*}[t]
\scriptsize
\centering
\begin{tabular}{ll}
\toprule
Role & Prompt \\
\midrule
system & You are a biologist specializing in drug-target interactions. Your responses should be formatted as JSON, with data listed under the key `target\_genes'.\\
user & List the top $10$ potential target genes for the drug \textit{drug name}. \\
\bottomrule
\end{tabular}
\caption{
Prompt template used for the GPT series to predict the top 10 target genes for each query drug. Here, \textit{drug name} is a specific name such as Bosutinib.
}
\label{tab:GPT-prompts}
\end{table*}

\begin{table}[t]
\centering
\begin{tabular}{ccccccc}
\toprule
Model &  Setting & Method & Centering & Top1 & Top10 & MRR \\
\midrule
\multirow{15}{*}{Our embeddings} & \multirow{5}{*}{G} & Random &  & 0.020  & 0.140 & 0.065\\
 &  & $\hat{\mathbf{v}}_\text{naive}$ &  & 0.067 & 0.197 & 0.111\\
 &  &  & $\checkmark$ & 0.125 & 0.381 & 0.209\\
 &  & $\hat{\mathbf{v}}$ &  & 0.206 & 0.455 & 0.289\\
 &  &  & $\checkmark$ & \textbf{0.300} & \textbf{0.686} & \textbf{0.426}\\
\cmidrule{2-7}
 & \multirow{5}{*}{P1} & Random &  & 0.020 & 0.142 & 0.066\\
 &  & $\hat{\mathbf{v}}_{p,\text{naive}}$ &  & 0.364 & 0.661 & 0.470\\
 &  &  & $\checkmark$ & 0.455 & 0.810 & 0.581\\
 &  & $\hat{\mathbf{v}}_p$ &  & 0.521 & 0.788 & 0.615\\
 &  &  & $\checkmark$ & \textbf{0.589} & \textbf{0.862} & \textbf{0.685}\\
\cmidrule{2-7}
 & \multirow{5}{*}{P2} & Random &  & 0.023 & 0.164 & 0.074\\
 &  & $\hat{\mathbf{v}}_{p,\text{naive}}$ &  & 0.376 & 0.682 & 0.484\\
 &  &  & $\checkmark$ & 0.473 & 0.837 & 0.601\\
 &  & $\hat{\mathbf{v}}_p$ &  & 0.530 & 0.803 & 0.626\\
 &  &  & $\checkmark$ & \textbf{0.600} & \textbf{0.880} & \textbf{0.700}\\
\bottomrule
\end{tabular}
\caption{Results of the naive estimators and the centering ablation study for settings G, P1, and P2.}
\label{tab:drug2gene_naive_centering_diff}
\end{table}

\begin{table}[t]
\centering
\begin{minipage}[b]{0.6\linewidth}
\centering
\begin{tabular}{ccccccc}
\toprule
& & &\multicolumn{3}{c}{Metric}\\
\cmidrule{4-6}
Model &  Setting & Method & Top1 & Top10 & MRR \\
\midrule
\multirow{6}{*}{BioConceptVec} & \multirow{2}{*}{G$^{'}$} & Random & 0.011  & 0.075  & 0.036\\
 &  & $\hat{\mathbf{v}}'$ & 0.233 & 0.560 & 0.345\\
 & \multirow{2}{*}{P1$^{'}$} & Random & 0.010  & 0.065  & 0.032\\
 &  & $\hat{\mathbf{v}}_p'$ & 0.426 & 0.679 & 0.515\\
 & \multirow{2}{*}{P2$^{'}$} & Random & 0.009  & 0.066  & 0.031\\
 &  & $\hat{\mathbf{v}}_p'$ & 0.248 & 0.507 & 0.337\\
\midrule
\multirow{6}{*}{Our embeddings} & \multirow{2}{*}{G$^{'}$} & Random &  0.010  & 0.078  & 0.037\\
 &  & $\hat{\mathbf{v}}'$ & 0.240 & 0.577 & 0.351\\
 & \multirow{2}{*}{P1$^{'}$} & Random & 0.012  & 0.070  & 0.034\\
 &  & $\hat{\mathbf{v}}_p'$ & 0.478 & 0.751 & 0.571\\
 & \multirow{2}{*}{P2$^{'}$} & Random & 0.009  & 0.070  & 0.033\\
 &  & $\hat{\mathbf{v}}_p'$ & 0.279 & 0.516 & 0.359\\
\bottomrule
\end{tabular}
\caption{Gene prediction performance in settings G${}'$, P1${}'$, and P2${}'$.}
\label{tab:gene2drug_result}
\end{minipage}
\end{table}

\begin{table}[t]
\centering
\begin{tabular}{lllrr}
\toprule
& & &\multicolumn{2}{c}{Metric}\\
\cmidrule{4-5}
Vocabulary &  Setting & LLM & Top1 & Top10 \\
\midrule
\multirow{9}{*}{BioConceptVec} & \multirow{3}{*}{G} & GPT-3.5 & 0.611 &  0.785 \\
 & & GPT-4 & 0.665 &  0.795 \\
 & & GPT-4o & 0.718 &  0.848 \\
\cmidrule{2-5}
& \multirow{3}{*}{P1} & GPT-3.5 &  0.581 &  0.780 \\
 & & GPT-4 &  0.633 &  0.792 \\
 & & GPT-4o &  0.689 &  0.855 \\
\cmidrule{2-5}
& \multirow{3}{*}{P2} & GPT-3.5 &  0.627 &  0.803 \\
 & & GPT-4 &  0.685 &  0.813 \\
 & & GPT-4o &  0.742 &  0.873 \\
\midrule
\multirow{9}{*}{Our embeddings} & \multirow{3}{*}{G} & GPT-3.5 &  0.637 &  0.804 \\
 & & GPT-4 &  0.693 &  0.818 \\
 & & GPT-4o &  0.741 &  0.867 \\
\cmidrule{2-5}
& \multirow{3}{*}{P1} & GPT-3.5 &  0.604 &  0.796 \\
 & & GPT-4 &  0.654 &  0.808 \\
 & & GPT-4o &  0.703 &  0.868 \\
\cmidrule{2-5}
& \multirow{3}{*}{P2} & GPT-3.5 &  0.653 &  0.821 \\
 & & GPT-4 &  0.710 &  0.832\\
 & & GPT-4o & 0.760 &  0.887 \\
\bottomrule
\end{tabular}
\caption{Gene prediction performance for GPT models.}
\label{tab:GPT-result}
\end{table}

\begin{table*}[t]
\centering
\begin{tabular}{lrrrrrrrrrrr}
\toprule
 & \multicolumn{11}{c}{Year} \\
  & 1975 & 1980 & 1985 & 1990 & 1995 & 2000 & 2005 & 2010 & 2015 & 2020 & 2023\\
\midrule
$|\mathcal{D}^y|$                &   3275 &   5059 &    7125 &    9387 &   11748 &  14044 &  16827 &  19758 &  22717 &  25431 &  28284 \\
$|\mathcal{G}^y|$                &    725 &   1782 &    3413 &    5976 &   10533 &  16949 &  24784 &  32264 &  39961 &  47509 &  51057 \\
$|\mathcal{R}^y|$                &    128 &    429 &     807 &    1251 &    2337 &   3282 &   4054 &   4849 &   5551 &   5909 &   5968 \\
\midrule
$|D^y|$                          &    104 &    343 &     666 &     904 &    1253 &   1436 &   1566 &   1708 &   1875 &   1961 &   1980 \\
$|G^y|$                          &     21 &     47 &      75 &     125 &     228 &    348 &    430 &    521 &    590 &    630 &    634 \\
\midrule
$\mathrm{E}_{d\in D^y}\{|[d]|\}$ &  1.231 &  1.251 &   1.212 &   1.384 &   1.865 &  2.286 &  2.589 &  2.839 &  2.961 &  3.013 &  3.014 \\
$\mathrm{E}_{g\in G^y}\{|[g]|\}$ &  6.095 &  9.128 &  10.760 &  10.008 &  10.250 &  9.431 &  9.428 &  9.307 &  9.408 &  9.379 &  9.413 \\
\bottomrule
\end{tabular}
\caption{Statistics for setting Y1.}
\label{tab:stats_Y1}
\end{table*}

\begin{table*}[t]
\centering
\begin{tabular}{lrrrrrrrrrr}
\toprule
 & \multicolumn{10}{c}{Year} \\
&   1975 &   1980 &   1985 &   1990 &   1995 &   2000 &   2005 &   2010 &   2015 &   2020 \\
\midrule
$\left|\mathcal{R}^{y \mid L_y}\right|$                                      &     37 &    117 &    256 &    499 &    859 &   1269 &   1637 &   1978 &   2354 &   2603 \\
$\left|\mathcal{R}^{y \mid U_y}\right|$                                      &     53 &    183 &    296 &    344 &    515 &    473 &    379 &    281 &    189 &     61 \\
\midrule
$\left|D^{y\mid U_y}\right|$                                                 &     48 &    154 &    272 &    288 &    369 &    325 &    256 &    189 &    131 &     48 \\
$\left|G^{y\mid U_y}\right|$                                                 &     14 &     34 &     55 &     89 &    137 &    173 &    153 &    140 &    115 &     44 \\
\midrule
$\mathrm{E}_{d\in D^{y\mid L_y}}\left\{\left|[d]^{y\mid L_y}\right|\right\}$ &  1.028 &  1.104 &  1.133 &  1.188 &  1.372 &  1.500 &  1.613 &  1.665 &  1.722 &  1.760 \\
$\mathrm{E}_{d\in D^{y\mid U_y}}\left\{\left|[d]^{y\mid U_y}\right|\right\}$ &  1.104 &  1.188 &  1.088 &  1.194 &  1.396 &  1.455 &  1.480 &  1.487 &  1.443 &  1.271 \\
$\mathrm{E}_{g\in G^{y\mid L_y}}\left\{\left|[g]^{y\mid L_y}\right|\right\}$ &  2.467 &  4.179 &  5.020 &  5.940 &  5.335 &  4.789 &  4.547 &  4.425 &  4.467 &  4.591 \\
$\mathrm{E}_{g\in G^{y\mid U_y}}\left\{\left|[g]^{y\mid U_y}\right|\right\}$ &  3.786 &  5.382 &  5.382 &  3.865 &  3.759 &  2.734 &  2.477 &  2.007 &  1.643 &  1.386 \\
\bottomrule
\end{tabular}
\caption{Statistics for setting Y2.}
\label{tab:stats_Y2}
\end{table*}

\subsubsection{Conversion from gene ID to gene name}
The BioConceptVec vocabulary registers genes using gene IDs. 
Therefore, we explain the procedure for converting gene IDs to gene names.
We used the KEGG API to obtain the names corresponding to the gene IDs in the BioConceptVec vocabulary. 
This API allows for batch requests; for example, to process gene IDs 20, 21, 22, 23, 24, the data is available at \url{https://rest.kegg.jp/list/hsa:20+hsa:21+hsa:22+hsa:23+hsa:24}. 
If multiple names corresponded to a single gene ID, we chose the first name. 
We also applied this conversion to genes in the vocabulary of our trained skip-gram model.

\subsubsection{Conversion from KEGG ID to MeSH ID}
In AsuratDB and the KEGG API, which is used to obtain drug-gene relations, genes are represented by gene IDs, while drugs are represented by KEGG IDs. 
Therefore, to use the data in BioConceptVec, we need to convert the KEGG IDs to MeSH IDs. 
We followed a four-step procedure (I), (II), (III), and (IV) to convert KEGG IDs to MeSH IDs.

\paragraph{(I) KEGG ID to PubChem SID or ChEBI ID}
Some drugs in KEGG are manually linked to external databases such as PubChem (\url{https://pubchem.ncbi.nlm.nih.gov/}) and ChEBI (\url{https://www.ebi.ac.uk/chebi/}), which are larger databases than KEGG.
Therefore, we used these databases to link their IDs to MeSH IDs. 

Since PubChem is maintained by the NCBI (National Center for Biotechnology Information, \url{https://www.ncbi.nlm.nih.gov/}) that also maintains the MeSH database, we prioritized the conversion of KEGG IDs to PubChem Substance IDs (SIDs). If we could not convert KEGG IDs directly to PubChem SIDs, we converted them to ChEBI IDs. 
The conversion from KEGG IDs to PubChem SIDs is available at \url{https://rest.kegg.jp/conv/pubchem/drug}, and the conversion to ChEBI IDs is available at \url{https://rest.kegg.jp/conv/chebi/drug}.

\paragraph{(II) PubChem SID or ChEBI ID to PubChem CID}
Next, we converted PubChem SIDs and ChEBI IDs to PubChem Compound IDs (CIDs), which are associated with MeSH IDs. 
The conversion from PubChem SIDs to PubChem CIDs is available at \url{https://pubchem.ncbi.nlm.nih.gov/rest/pug/substance/sourceall/KEGG/cids/json}. 
To convert ChEBI IDs to PubChem CIDs, we used the PubChem API. 
For example, for ChEBI ID 39112, the corresponding data can be obtained at \url{https://pubchem.ncbi.nlm.nih.gov/rest/pug//compound/xref/RegistryID/chebi:39112/cids/json}.

\paragraph{(III) Convert CIDs to MeSH Headings}
We used the Entrez (\url{https://www.ncbi.nlm.nih.gov/Web/Search/entrezfs.html}) System API provided by NCBI, and obtained the MeSH headings associated with each CID. 
This API allows for batch requests. 
For example, for CIDs 5328940 and 156413, the corresponding data can be obtained from the following URL: \url{https://eutils.ncbi.nlm.nih.gov/entrez/eutils/esummary.fcgi?db=pccompound&id=5328940,156413&retmode=json}.

\paragraph{(IV) MeSH Headings to MeSH ID}
Since we had already obtained the correspondence between MeSH IDs and MeSH headings in Supplementary Information~\ref{supp:meshid_to_drugname}, we could match MeSH IDs with the MeSH headings obtained through the Entrez system API. 
In this way, we converted the original KEGG IDs to MeSH IDs.
Note that multiple KEGG IDs may correspond to the same MeSH ID. In such cases, we chose the smallest KEGG ID to associate with the MeSH ID. 

\begin{table*}[t]
\small
\centering
\begin{tabular}{lll|rrrrrrrrrrr}
\toprule
& & & \multicolumn{11}{c}{Year}\\
Metric & Setting & Method & 1975 & 1980 & 1985 & 1990 & 1995 & 2000 & 2005 & 2010 & 2015 & 2020 & 2023\\
\midrule
\multirow{4}{*}{Top1} & \multirow{2}{*}{Y1} & Random & 0.143 &  0.066 &  0.054 &  0.036 &  0.027 &  0.022 &  0.024 &  0.022 &  0.022 &  0.019 &  0.019\\
 &  & $\hat{\mathbf{v}}^y$ & 0.365 &  0.262 &  0.249 &  0.236 &  0.247 &  0.272 &  0.275 &  0.287 &  0.323 &  0.301 &  0.300\\
 & \multirow{2}{*}{Y2} & Random & 0.104 &  0.081 &  0.043 &  0.028 &  0.016 &  0.012 &  0.013 &  0.006 &  0.009 &  0.004 & --\\
 &  & $\hat{\mathbf{v}}^{y \mid L_y}$ &  0.333 &  0.208 &  0.165 &  0.132 &  0.076 &  0.092 &  0.062 &  0.090 &  0.107 &  0.021 & --\\
\midrule
\multirow{4}{*}{Top10} & \multirow{2}{*}{Y1} & Random & 0.830 &  0.554 &  0.426 &  0.317 &  0.236 &  0.196 &  0.185 &  0.165 &  0.153 &  0.137 &  0.140 \\
 &  & $\hat{\mathbf{v}}^y$ & 0.837 &  0.577 &  0.563 &  0.562 &  0.578 &  0.652 &  0.691 &  0.651 &  0.714 &  0.693 &  0.686\\
 & \multirow{2}{*}{Y2} & Random & 0.835 &  0.538 &  0.382 &  0.253 &  0.157 &  0.123 &  0.098 &  0.064 &  0.062 &  0.081 & --\\
 &  & $\hat{\mathbf{v}}^{y \mid L_y}$ & 0.625 &  0.487 &  0.438 &  0.410 &  0.347 &  0.397 &  0.352 &  0.344 &  0.328 &  0.271 & --\\
\midrule
\multirow{4}{*}{MRR} & \multirow{2}{*}{Y1} & Random & 0.340 &  0.204 &  0.167 &  0.127 &  0.100 &  0.084 &  0.081 &  0.073 &  0.069 &  0.064 &  0.064  \\
 &  & $\hat{\mathbf{v}}^y$ & 0.534 &  0.370 &  0.353 &  0.346 &  0.363 &  0.399 &  0.410 &  0.410 &  0.455 &  0.430 &  0.426\\
 & \multirow{2}{*}{Y2} & Random & 0.305 &  0.211 &  0.150 &  0.104 &  0.070 &  0.055 &  0.049 &  0.034 &  0.034 &  0.032 & --\\
 &  & $\hat{\mathbf{v}}^{y \mid L_y}$ &  0.438 &  0.304 &  0.253 &  0.214 &  0.163 &  0.182 &  0.147 &  0.173 &  0.188 &  0.103 & --\\
\bottomrule
\end{tabular}
\caption{Gene prediction performance in settings Y1 and Y2.}
\label{tab:drug2gene_year_result}
\end{table*}

\begin{table*}[t]
\centering
\begin{tabular}{l@{\hspace{0.5em}}r@{\hspace{0.5em}}r@{\hspace{0.5em}}r@{\hspace{0.5em}}r@{\hspace{0.5em}}r@{\hspace{0.5em}}r@{\hspace{0.5em}}r@{\hspace{0.5em}}r@{\hspace{0.5em}}r@{\hspace{0.5em}}r@{\hspace{0.5em}}r}
\toprule
 & \multicolumn{11}{c}{Year} \\
  & 1975 &    1980 &    1985 &    1990 &   1995 &   2000 &   2005 &   2010 &   2015 &   2020 &   2023 \\
\midrule
$\sum_{p\in\mathcal{P}}|D_p^y|$                                                        &      85 &     536 &    1092 &    1668 &   2271 &   2609 &   2897 &   3122 &   3447 &   3578 &   3612 \\
$\sum_{p\in\mathcal{P}}|G_p^y|$                                                        &      15 &      76 &     127 &     234 &    444 &    634 &    765 &    916 &   1088 &   1167 &   1178 \\
$\sum_{p\in\mathcal{P}}|\mathcal{D}_p^y\cap D^y|$                                      &     397 &    1015 &    1559 &    2055 &   2394 &   2690 &   2909 &   3132 &   3458 &   3581 &   3614 \\
$\sum_{p\in\mathcal{P}}|\mathcal{G}_p^y\cap G^y|$                                      &      25 &     136 &     271 &     553 &   1242 &   1896 &   2325 &   2570 &   2952 &   3174 &   3186 \\
\midrule
$\mathrm{E}_{p\in\mathcal{P}}\{\mathrm{E}_{d\in D_p^y}\{|[d]_p^y|\}\}$                 &   1.078 &   1.446 &   1.289 &   1.341 &  1.603 &  1.803 &  1.841 &  2.027 &  1.983 &  1.979 &  1.990 \\
$\mathrm{E}_{p\in\mathcal{P}}\{\mathrm{E}_{g\in G_p^y}\{|[g]_p^y|\}\}$                 &   6.143 &   8.485 &   8.084 &   7.485 &  7.027 &  6.032 &  5.607 &  5.529 &  4.973 &  4.830 &  4.847 \\
$\mathrm{E}_{p\in\mathcal{P}}\{\mathrm{E}_{d\in \mathcal{D}_p^y\cap D^y}\{|[d]^y|\}\}$ &   1.078 &   1.430 &   1.356 &   1.616 &  2.211 &  2.532 &  2.724 &  2.902 &  2.749 &  2.759 &  2.776 \\
$\mathrm{E}_{p\in\mathcal{P}}\{\mathrm{E}_{g\in \mathcal{G}_p^y\cap G^y}\{|[g]^y|\}\}$ &  10.000 &  13.496 &  12.787 &  11.633 &  8.815 &  7.163 &  6.968 &  6.505 &  6.525 &  6.692 &  6.714 \\
\bottomrule
\end{tabular}
\caption{Statistics for settings P1Y1 and P2Y1.}
\label{tab:stats_P1Y1_P2Y1}
\end{table*}

\begin{table*}[t]
\centering
\begin{tabular}{l@{\hspace{0.5em}}r@{\hspace{0.5em}}r@{\hspace{0.5em}}r@{\hspace{0.5em}}r@{\hspace{0.5em}}r@{\hspace{0.5em}}r@{\hspace{0.5em}}r@{\hspace{0.5em}}r@{\hspace{0.5em}}r@{\hspace{0.5em}}r}
\toprule
 & \multicolumn{10}{c}{Year} \\
  & 1975 &    1980 &   1985 &   1990 &   1995 &   2000 &   2005 &   2010 &   2015 &   2020 \\
\midrule
$\sum_{p\in\mathcal{P}}\left|D_p^{y \mid U_y}\right|$                                                                                       &     40 &     253 &    419 &    448 &    565 &    500 &    426 &    303 &    195 &     66 \\
$\sum_{p\in\mathcal{P}}\left|G_p^{y \mid U_y}\right|$                                                                                       &     10 &      58 &     90 &    154 &    272 &    295 &    250 &    218 &    163 &     61 \\
$\sum_{p\in\mathcal{P}}|\mathcal{D}_p^y\cap D^{y\mid U_y}|$                                                                                 &     42 &     275 &    464 &    492 &    681 &    612 &    511 &    387 &    257 &     96 \\
$\sum_{p\in\mathcal{P}}|\mathcal{G}_p^y\cap G^{y\mid U_y}|$                                                                                 &     14 &      68 &    121 &    250 &    531 &    607 &    589 &    549 &    496 &    206 \\
\midrule
$\mathrm{E}_{p\in\mathcal{P}}\left\{\mathrm{E}_{d\in D_p^{y \mid U_y}}\left\{\left|[d]_p^{y\mid U_y}\right|\right\}\right\}$                &  1.034 &   1.445 &  1.247 &  1.285 &  1.477 &  1.571 &  1.596 &  1.724 &  1.648 &  1.614 \\
$\mathrm{E}_{p\in\mathcal{P}}\left\{\mathrm{E}_{g\in G_p^{y \mid U_y}}\left\{\left|[g]_p^{y\mid U_y}\right|\right\}\right\}$                &  4.194 &   7.157 &  6.560 &  5.761 &  5.107 &  4.329 &  3.924 &  3.884 &  3.538 &  3.443 \\
$\mathrm{E}_{p\in\mathcal{P}}\left\{\mathrm{E}_{d\in \mathcal{D}_p^y\cap D^{y\mid U_y}}\left\{\left|[d]^{y\mid U_y}\right|\right\}\right\}$ &  1.034 &   1.427 &  1.301 &  1.497 &  1.925 &  2.126 &  2.288 &  2.376 &  2.176 &  2.113 \\
$\mathrm{E}_{p\in\mathcal{P}}\left\{\mathrm{E}_{g\in \mathcal{G}_p^y\cap G^{y\mid U_y}}\left\{\left|[g]^{y\mid U_y}\right|\right\}\right\}$ &  7.000 &  10.525 &  9.027 &  7.890 &  5.667 &  4.729 &  4.392 &  4.155 &  4.096 &  4.249 \\
\bottomrule
\end{tabular}
\caption{Statistics for settings P1Y2 and P2Y2.}
\label{tab:stats_P1Y2_P2Y2}
\end{table*}

\subsection{Details of predictions using random baseline}\label{supp:random_baseline}
This section describes the sets used for sampling in the random baseline across different settings. 
For settings G, P1, and P2, the set used was $G$. Similarly, $D$ was used for G$^{'}$, P1$^{'}$, and P2$^{'}$, while $G^y$ was used for settings Y1, P1Y1, and P2Y1. 
For settings Y2, P1Y2, and P2Y2, $G^{y\mid L_y}\cup G^{y\mid U_y}$ was used instead of $G^{y\mid U_y}$; 
This was intended to prevent extremely high random baseline performance when the size of $G^{y\mid U_y}$ is very small.

\subsection{Details of predictions using the GPT series}\label{supp:GPT-setting}
In this study, we used GPT-3.5, GPT-4, and GPT-4o as baselines for generative models. The details of the models are shown in Table~\ref{tab:GPT-models}. The prompts are shown in Table~\ref{tab:GPT-prompts}.

\subsection{Details of predictions by TransE}\label{supp:KGE-setting}
This section describes the details of the experimental settings for TransE.  
Embeddings are learned for each of settings G, P1, and P2.
In our experiments, we utilized TransE with the following hyperparameters: a batch size of 1024, embedding dimensions set to 500, a learning rate of 0.0001, and a total of 800 iteration steps. Additionally, we adopted the self-adversarial negative sampling method from RotatE \cite{sun2019rotate}, configuring the negative sampling size to 512 and the sampling temperature to 0.5.
For each setting, the triplets of drug $d$, gene $g$, and drug-gene relation $(d,g)$ are used to evaluate performance in analogy tasks and are then split into 60\% training, 20\% validation, and 20\% test data.  
In setting G, only a single type of relation, the drug-gene relation, is considered.  
In settings P1 and P2, the drug-gene relation $(d,g)$ is distinguished based on each pathway $p\in\mathcal{P}$, resulting in $|\mathcal{P}|$ different relations.  
For entities in the validation and test data that do not appear in the training data, random embeddings are simply used.  
Since KGE typically uses the entire set of head and tail entities as its search space, we similarly define the search space for TransE as the entire set of drugs and genes that have drug-gene relations for each setting.
Additionally, note that in Table~\ref{tab:drug2gene_result}, while random baseline, GPT series, and our method evaluate performance on the entire dataset, TransE evaluates performance on the 20\% test data.

\section{
Details of experimental results
}~\label{supp:results}
\subsection{Analogy tasks for drug-gene pairs}

\subsubsection{Comparison of estimators and centering ablation study}~\label{supp:naive-centering}
For settings G, P1, and P2, Table~\ref{tab:drug2gene_naive_centering_diff} shows the results of the naive estimators defined in equations (\ref{eq:v_def1}) and (\ref{eq:vp_naive_def1}) and the centering ablation study, using our skip-gram.
In all settings, the results of the estimators $\hat{\mathbf{v}}$ and $\hat{\mathbf{v}}_p$ calculated from drug-gene relations outperformed those of the naive estimators $\hat{\mathbf{v}}_\text{naive}$ and $\hat{\mathbf{v}}_{p,\text{naive}}$. We also observed performance improvements due to centering.
Based on this, we showed the results using the estimators calculated from the drug-gene relations with centering applied in Table~\ref{tab:drug2gene_result}.

\subsubsection{Prediction of drugs from genes}~\label{supp:gene2drug}
As seen in Supplementary Information~\ref{supp:all-setting}, Table~\ref{tab:gene2drug_setting} defines settings G$^{'}$, P1$^{'}$, and P2$^{'}$ for the analogy tasks for predicting drugs from genes.
We then define estimators to perform the analogy tasks in these settings.
Similar to the estimator $\hat{\mathbf{v}}$ in Eq.~(\ref{eq:v_def2}) for setting G, we define the estimator $\hat{\mathbf{v}}'$ for setting G$^{'}$:
\begin{align}
    \hat{\mathbf{v}}' := -\hat{\mathbf{v}} = \mathrm{E}_\mathcal{\mathcal{R}}\{\mathbf{u}_d-\mathbf{u}_g\} =
    \frac{1}{|\mathcal{R}|}\sum_{(d,g)\in \mathcal{R}}(\mathbf{u}_d-\mathbf{u}_g).\label{eq:v_prime_def}
\end{align}
Similar to the estimator $\hat{\mathbf{v}}_p$ in Eq.~(\ref{eq:vp_def}) for settings P1 and P2, we also define the estimator $\hat{\mathbf{v}}_p'$ for settings P1$^{'}$ and P2$^{'}$:
\begin{align}
    &\hat{\mathbf{v}}_p' := -\hat{\mathbf{v}}_p=
\mathrm{E}_{\mathcal{R}_p}\{\mathbf{u}_d - \mathbf{u}_g\}
= \frac{1}{|\mathcal{R}_p|}\sum_{(d,g)\in \mathcal{R}_p}(\mathbf{u}_d - \mathbf{u}_g)\label{eq:vp_prime_def}
.\end{align}

For settings G$^{'}$, P1$^{'}$, and P2$^{'}$, we performed the analogy tasks using $\hat{\mathbf{v}}'$ and $\hat{\mathbf{v}}_p'$, and Table~\ref{tab:gene2drug_result} shows the results.
Similar to Table~\ref{tab:drug2gene_result}, the estimators calculated from the drug-gene relations outperformed the random baseline.

While Table~\ref{tab:drug2gene_result} shows higher scores for setting P2 compared to setting P1, Table~\ref{tab:gene2drug_result} shows higher scores for setting P1${}'$ compared to setting P2${}'$. 
To explain these results, we focus on the sizes of the query and answer sets, using actual values from BioConceptVec.
As shown in Table~\ref{tab:stats_pathway}, the sizes of the query sets for settings P1 and P2 are $\sum_{p\in\mathcal{P}}|D_p|=4087$ and $\sum_{p\in\mathcal{P}}|\mathcal{D}_p\cap D|=4091$, respectively. The ratio is $\sum_{p\in\mathcal{P}}|\mathcal{D}_p\cap D|/\sum_{p\in\mathcal{P}}|D_p| \approx1.001$.
On the other hand, for settings P1$^{'}$ and P2$^{'}$, the sizes of the query sets are $\sum_{p\in\mathcal{P}}|G_p|=1251$ and $\sum_{p\in\mathcal{P}}|\mathcal{G}_p\cap G|=3463$, respectively, with a ratio of $\sum_{p\in\mathcal{P}}|\mathcal{G}_p\cap G|/\sum_{p\in\mathcal{P}}|G_p|\approx2.768$. 
For settings P1, P2, P1${}'$, and P2${}'$, the answer sets defined in Tables~\ref{tab:drug2gene_setting} and~\ref{tab:gene2drug_setting} are $[d]_p$, $[d]$, and $[g]_p$, $[g]$, respectively. 
By definition, the sizes of the answer sets satisfy $|[d]_p|\leq |[d]|$ and $|[g]_p|\leq |[g]|$. 
In fact, the expected values of them are $\mathrm{E}_{p\in\mathcal{P}}\{\mathrm{E}_{d\in D_p}\{|[d]_p|\}\}=1.965$, $\mathrm{E}_{d\in D}\{|[d]|\}=2.938$, $\mathrm{E}_{p\in\mathcal{P}}\{\mathrm{E}_{g\in G_p}\{|[g]_p|\}\}=5.050$, and $\mathrm{E}_{g\in G}\{|[g]|\}=10.008$, respectively.
In settings P1 and P2, the sizes of the query sets are nearly identical, but the larger answer sets in setting P2 probably make the task easier. 
On the other hand, the sizes of the query sets in setting P2$^{'}$ are more than twice as large as those in P1$^{'}$, which likely contributes to the lower evaluation metric scores in setting P2$^{'}$.

\subsubsection{Predictions by GPT models}\label{supp:GPT-result}
The prediction results from the GPT models are shown in Table~\ref{tab:GPT-result}.
The set of query drugs and the set of correct target genes depend not only on settings G, P1, and P2 but also on the vocabulary set. Here, we conducted experiments using vocabularies obtained from BioConceptVec or our skip-gram embeddings. The best accuracies for all combinations of vocabularies and settings are summarized in the Table~\ref{tab:drug2gene_result} in the main text.

\subsection{Analogy tasks for drug-gene pairs by year}

\begin{figure*}[t!]
  \begin{subfigure}[b]{0.48\textwidth}
    \includegraphics[keepaspectratio, width=\textwidth]{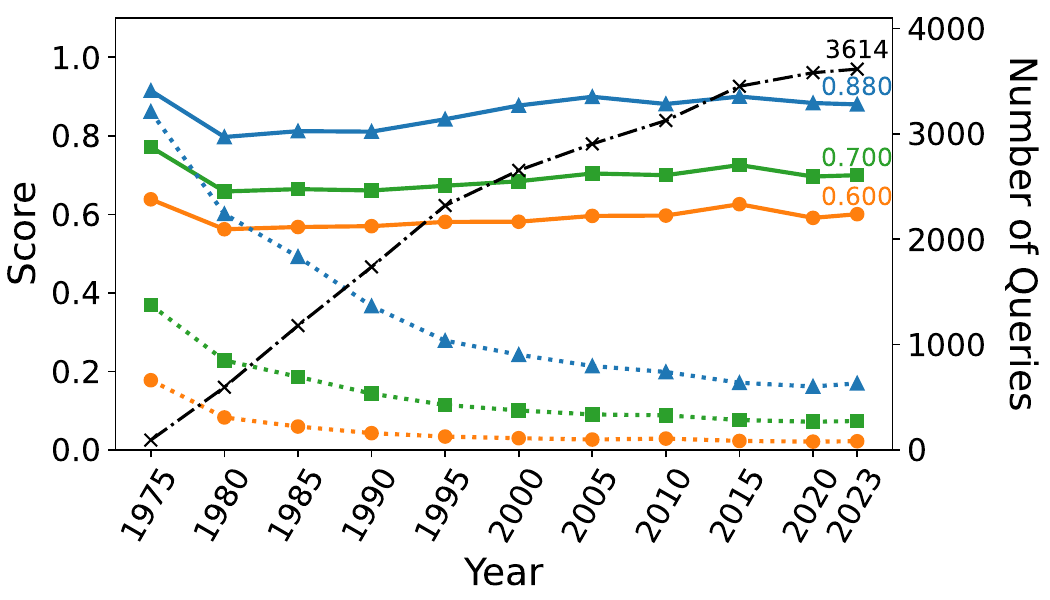}
    \caption{P2Y1}
    \label{fig:py2}
  \end{subfigure}
  \begin{subfigure}[b]{0.48\textwidth}
    \includegraphics[keepaspectratio, width=\textwidth]{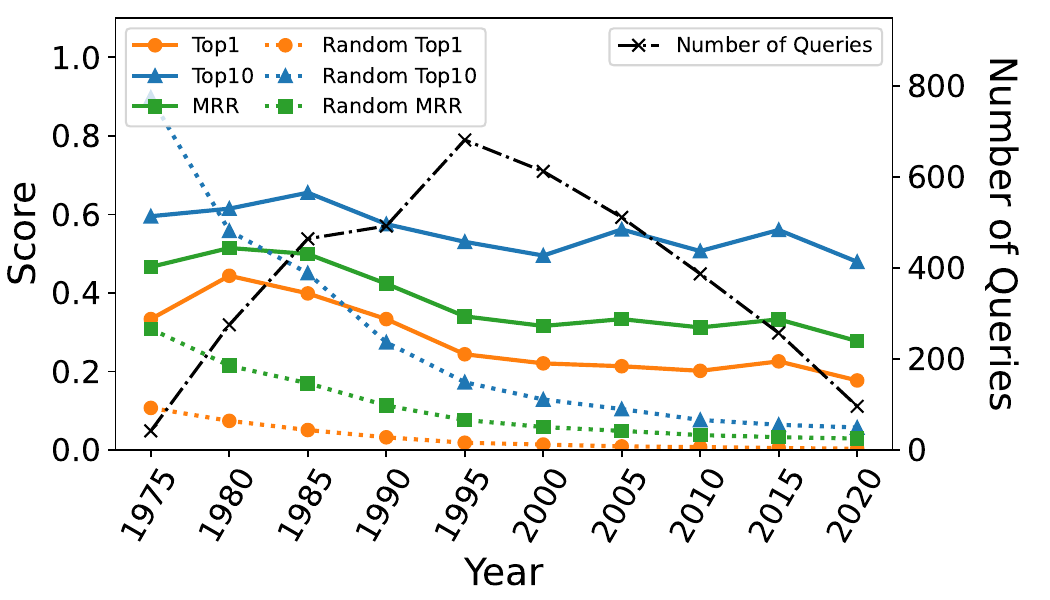}
    \caption{P2Y2}
    \label{fig:py4}
  \end{subfigure}
  \caption{Gene prediction performance and the number of queries for each year in settings P2Y1 and P2Y2.}
  \label{fig:P2Y1-P2Y2}
\end{figure*}

\begin{table*}[t]
\small
\centering
\begin{tabular}{lll|rrrrrrrrrrr}
\toprule
& & & \multicolumn{11}{c}{Year}\\
Metric & Setting & Method & 1975 &   1980 &   1985 &   1990 &   1995 &   2000 &   2005 &   2010 &   2015 &   2020 &   2023 \\
\midrule
\multirow{8}{*}{Top1} & \multirow{2}{*}{P1Y1} & Random & 0.151 &  0.086 &  0.062 &  0.044 &  0.028 &  0.025 &  0.022 &  0.021 &  0.021 &  0.019 &  0.020 \\
 &  & $\hat{\mathbf{v}}_p^y$ & 0.706 &  0.621 &  0.613 &  0.592 &  0.589 &  0.588 &  0.588 &  0.587 &  0.614 &  0.582 &  0.589 \\
 & \multirow{2}{*}{P2Y1} & Random & 0.178 &  0.083 &  0.060 &  0.043 &  0.034 &  0.030 &  0.027 &  0.029 &  0.023 &  0.021 &  0.022  \\
 &  & $\hat{\mathbf{v}}_p^y$ & 0.638 &  0.562 &  0.568 &  0.570 &  0.581 &  0.581 &  0.596 &  0.597 &  0.626 &  0.591 &  0.600 \\
 & \multirow{2}{*}{P1Y2} & Random & 0.147 &  0.079 &  0.054 &  0.033 &  0.017 &  0.012 &  0.013 &  0.007 &  0.004 &  0.006 & --\\
 &  & $\hat{\mathbf{v}}_p^{y \mid L_y}$ & 0.350 &  0.482 &  0.442 &  0.366 &  0.290 &  0.266 &  0.254 &  0.244 &  0.272 &  0.258 & --\\
 & \multirow{2}{*}{P2Y2} & Random & 0.107 &  0.074 &  0.051 &  0.032 &  0.019 &  0.014 &  0.009 &  0.007 &  0.005 &  0.003 & --\\
 &  & $\hat{\mathbf{v}}_p^{y \mid L_y}$ & 0.333 &  0.444 &  0.399 &  0.333 &  0.244 &  0.221 &  0.213 &  0.202 &  0.226 &  0.177 & --\\
\midrule
\multirow{8}{*}{Top10} & \multirow{2}{*}{P1Y1} & Random & 0.906 &  0.602 &  0.486 &  0.351 &  0.245 &  0.214 &  0.186 &  0.165 &  0.146 &  0.140 &  0.142  \\
 &  & $\hat{\mathbf{v}}_p^y$ & 1.000 &  0.871 &  0.862 &  0.835 &  0.847 &  0.881 &  0.889 &  0.867 &  0.886 &  0.869 &  0.862 \\
 & \multirow{2}{*}{P2Y1} & Random & 0.862 &  0.601 &  0.492 &  0.366 &  0.278 &  0.243 &  0.214 &  0.199 &  0.171 &  0.162 &  0.170  \\
 &  & $\hat{\mathbf{v}}_p^y$ & 0.915 &  0.797 &  0.812 &  0.811 &  0.842 &  0.877 &  0.899 &  0.881 &  0.900 &  0.884 &  0.880 \\
 & \multirow{2}{*}{P1Y2} & Random & 0.922 &  0.583 &  0.448 &  0.269 &  0.156 &  0.118 &  0.098 &  0.068 &  0.061 &  0.076 & --\\
 &  & $\hat{\mathbf{v}}_p^{y \mid L_y}$ & 0.625 &  0.648 &  0.704 &  0.621 &  0.586 &  0.564 &  0.620 &  0.558 &  0.662 &  0.606 & --\\
 & \multirow{2}{*}{P2Y2} & Random & 0.898 &  0.558 &  0.450 &  0.274 &  0.173 &  0.129 &  0.104 &  0.076 &  0.064 &  0.057 & --\\
 &  & $\hat{\mathbf{v}}_p^{y \mid L_y}$ & 0.595 &  0.615 &  0.655 &  0.575 &  0.530 &  0.495 &  0.562 &  0.506 &  0.560 &  0.479 & --\\
\midrule
\multirow{8}{*}{MRR} & \multirow{2}{*}{P1Y1} & Random & 0.366 &  0.231 &  0.186 &  0.140 &  0.102 &  0.090 &  0.080 &  0.073 &  0.068 &  0.064 &  0.065 \\
 &  & $\hat{\mathbf{v}}_p^y$ &  0.849 &  0.724 &  0.712 &  0.684 &  0.680 &  0.690 &  0.694 &  0.688 &  0.713 &  0.684 &  0.685 \\
 & \multirow{2}{*}{P2Y1} & Random & 0.369 &  0.228 &  0.186 &  0.144 &  0.115 &  0.101 &  0.091 &  0.088 &  0.076 &  0.072 &  0.074 \\
 &  & $\hat{\mathbf{v}}_p^y$ & 0.772 &  0.658 &  0.664 &  0.661 &  0.673 &  0.684 &  0.704 &  0.700 &  0.726 &  0.696 &  0.700 \\
 & \multirow{2}{*}{P1Y2} & Random & 0.358 &  0.224 &  0.169 &  0.113 &  0.071 &  0.054 &  0.049 &  0.034 &  0.031 &  0.034 & --\\
 &  & $\hat{\mathbf{v}}_p^{y \mid L_y}$ & 0.488 &  0.555 &  0.547 &  0.459 &  0.392 &  0.370 &  0.384 &  0.362 &  0.400 &  0.382 & --\\
 & \multirow{2}{*}{P2Y2} & Random & 0.308 &  0.215 &  0.170 &  0.113 &  0.076 &  0.058 &  0.049 &  0.038 &  0.033 &  0.029 & --\\
 &  & $\hat{\mathbf{v}}_p^{y \mid L_y}$ & 0.466 &  0.514 &  0.499 &  0.423 &  0.340 &  0.316 &  0.333 &  0.312 &  0.333 &  0.277 & --\\
\bottomrule
\end{tabular}
\caption{Gene prediction performance in settings P1Y1, P2Y1, P1Y2, and P2Y2.}
\label{tab:drug2gene_year_pathway_result}
\end{table*}

\subsubsection{Global setting by year}~\label{supp:year-results}
Tables~\ref{tab:stats_Y1} and~\ref{tab:stats_Y2} show the statistics for settings Y1 and Y2. 
As seen in Table~\ref{tab:drug2gene_setting}, in setting Y1, the query is $d \in D^y$, and the search space is $\mathcal{G}^y$. 
In setting Y2, the query is $d \in D^{y \mid U_y}$, and the search space is $\mathcal{G}^y \setminus [d]^{y \mid L_y}$. 
In Table~\ref{tab:stats_Y1}, $D^y$ and $\mathcal{G}^y$ show a monotonic increase over the years.
On the other hand, Table~\ref{tab:stats_Y2} shows that $D^{y \mid U_y}$ has increased since 1975 and started to decrease after 1995. 

We showed the plots of the results for settings Y1 and Y2 in Fig.~\ref{fig:Y1-Y2-P1Y1-P1Y2}.
For settings Y1 and Y2, Table~\ref{tab:drug2gene_year_result} shows detailed results used in Fig.~\ref{fig:Y1-Y2-P1Y1-P1Y2}.

\subsubsection{Pahtway-wise setting by year}~\label{supp:year-pathway-results}
Tables~\ref{tab:stats_P1Y1_P2Y1} and~\ref{tab:stats_P1Y2_P2Y2} show the statistics for settings P1Y1, P2Y1, P1Y2, and P2Y2.
As seen in Table~\ref{tab:drug2gene_setting}, the query is $d\in D_p^{y}$ in setting P1Y1, $d\in \mathcal{D}_p^y\cap D^{y}$ in setting P2Y1, $d\in D_p^{y\mid U_y}$ in setting P1Y2, and $d\in \mathcal{D}_p^y\cap D^{y\mid U_{y}}$ in P2Y2.
In Table~\ref{tab:stats_P1Y1_P2Y1}, $D_p^{y}$ and $\mathcal{D}_p^y\cap D^{y}$ show a monotonic increase over the years.
On the other hand, Table~\ref{tab:stats_P1Y2_P2Y2} shows that $D_p^{y\mid U_y}$ and $\mathcal{D}_p^y\cap D^{y\mid U_{y}}$ have increased since 1975 and started to decrease after 1995. 

We showed the plots of the results for settings P1Y1 and P1Y2 in Fig.~\ref{fig:Y1-Y2-P1Y1-P1Y2}.
Fig.~\ref{fig:P2Y1-P2Y2} shows the plots of the results for settings P2Y1 and P2Y2. 
For settings P1Y1, P2Y1, P1Y2, and P2Y2, Table~\ref{tab:drug2gene_year_pathway_result} shows detailed results used in Fig.~\ref{fig:Y1-Y2-P1Y1-P1Y2} and Fig.~\ref{fig:P2Y1-P2Y2}.
As shown in Table~\ref{tab:drug2gene_result}, settings P1 and P2 show roughly the same trends. Similarly, in Table~\ref{tab:drug2gene_year_pathway_result}, settings P1Y1 and P2Y1, as well as P1Y2 and P2Y2, each show similar trends.

\subsection{Correlation between answer set sizes and predicted ranks}\label{supp:ansfreq_rank}
\begin{figure}[!t]
\centering
\begin{subfigure}{\textwidth}
    \includegraphics[width=\textwidth]{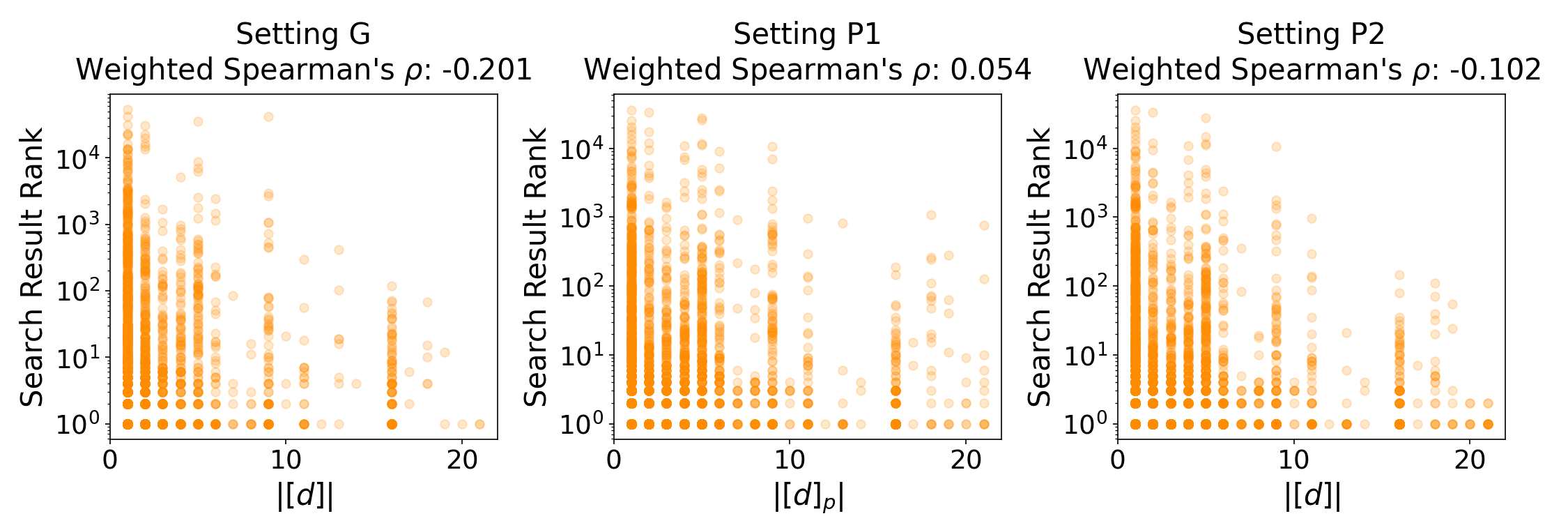}
    \caption{BioConceptVec}
    \label{fig:ansfreq_rank_bcv}
\end{subfigure}
\begin{subfigure}{\textwidth}
    \includegraphics[width=\textwidth]{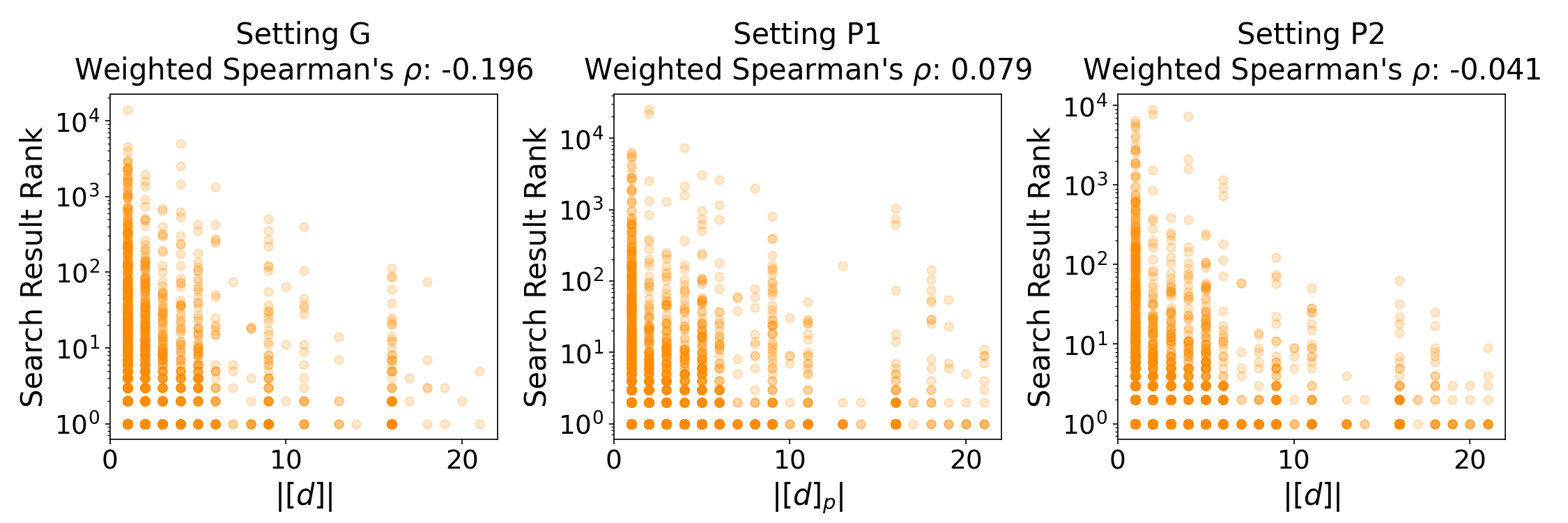}
    \caption{Our embeddings}
    \label{fig:ansfreq_rank_homemade}
\end{subfigure}
\caption{
Scatter plots of the answer set sizes for each drug $d$ and the ranks of the predicted genes.
}
\label{fig:ansfreq_rank}
\end{figure}
Fig.~\ref{fig:ansfreq_rank} shows scatter plots and correlation coefficients between the answer set sizes for each drug $d$ and the ranks of the predicted genes obtained by adding relation vectors in settings G, P1, and P2.
Since smaller answer set sizes are more frequent (see Fig.~\ref{fig:ansdist_drug2gene} in Supplementary Information~\ref{supp:embeddings} for the distribution of answer set sizes), we calculated a weighted Pearson correlation coefficient, where the weight is the reciprocal of the frequency of each answer set size.

\subsection{Details of comparison of analogy computation and TransE}\label{supp:analogy_vs_transe}
This section provides details regarding the experiments presented in Fig.~\ref{fig:analogy_vs_transe}, comparing our analogy computation with TransE. 
We conducted experiments under setting P1, which is a simple setting and involves multiple drug-gene relations.
For drug-gene relations in Setting P1, we fixed 20\% of the data as validation data and another 20\% as test data, following the same setting for TransE in Table~\ref{tab:drug2gene_result}, and used them consistently across both methods.
From the remaining 60\% of the data, we created training subsets ranging from 10\% to 60\%, ensuring that each smaller subset is contained within the next larger subset. 
Formally, if we define the 10\% to 60\% training sets as $\mathcal{T}_{10}, \ldots, \mathcal{T}_{60}$, then $\mathcal{T}_{10} \subset \mathcal{T}_{20} \subset \mathcal{T}_{30} \subset \mathcal{T}_{40} \subset \mathcal{T}_{50} \subset \mathcal{T}_{60}$.

We explain the experimental settings of our method. We used our skip-gram embeddings.
At test time, for pathways present in the training data, we used the relation vector $\hat{\mathbf{v}}_p$ defined in Eq.~(\ref{eq:vp_def}).
For pathways not present in the training data, we used the relation vector $\hat{\mathbf{v}}$ computed from all drug-gene relations in the training data, as defined in Eq.~(\ref{eq:v_def2}).
The search space consisted of all genes appearing in the training, validation, and test data.

The experimental settings for TransE follow those described in Supplementary Information~\ref{supp:KGE-setting}.
\end{document}